# On the (In)Significance of Feature Selection in High-Dimensional Datasets

Bhavesh Neekhra[1*], Debayan Gupta[1*], Partha Pratim Chakrabarti[2]

**Affiliations:**

[1]Department of Computer Science, Ashoka University; Sonipat, 131029, India.

[2]Department of Computer Science and Engineering, Indian Institute of Technology; Kharagpur, 721302, India

## Abstract

Feature selection (FS) is assumed to improve predictive performance and identify meaningful features in high-dimensional datasets. Surprisingly, small random subsets of features (0.02 - 1%) match or outperform the predictive performance of both full feature sets and FS across 28 out of 30 diverse datasets (microarray, bulk and single-cell RNA-Seq, mass spectrometry, imaging, etc.). In short, any arbitrary set of features is as good as any other (with surprisingly low variance in results) - so how can a particular set of selected features be "important" if they perform no better than an arbitrary set? These results challenge the assumption that computationally selected features reliably capture meaningful signals, emphasizing the importance of rigorous validation before interpreting selected features as actionable, particularly in computational genomics.

**Keywords:** Feature selection, gene expression, null hypothesis testing, high-dimensional data, computational genomics, machine learning

## Introduction

Feature selection is a critical part of practically all research involving machine learning on high-dimensional datasets. Beyond just computational efficiency, feature selection is often used to find features that are "important" to a given outcome. Some examples of feature selection methods are *(1-6)*. As of September 2025, a search for "feature selection on high dimensional datasets" on Google Scholar lists about 2,790,000 results - it is beyond doubt that huge resources are being invested in this area.

Corresponding authors email: debayan.gupta@ashoka.edu.in, bhavesh.neekhra_phd18@ashoka.edu.in

A natural question to ask is: "How significant are the results?" - i.e., to test the null hypothesis implicit in these works. We investigate this question using multiple datasets with varying characteristics in terms of sample size, number of features, number of classes, and datatypes.

In particular, gene expression datasets are high dimensional but comparatively low in sample size. A common strategy is to apply feature selection methods to identify "important'" genes for subsequent model training. However, a critical baseline, comparing the performance of selected features against that of random feature subset (of same size as "important" features), has not been explored. In other words, the null hypothesis comparison of how much better these selected genes are vs random choice does not appear to have been investigated: we find that gene expression data exhibits peculiar statistical properties as we will see in the Results section.

We find that, in many cases, the performance of models trained on features selected by published methods can be matched by using random subsets of size $c*k$, for a small constant $c$, where $k$ is the number of selected "important'" features (using ensembles further improves performance when constituent models are trained with random subsets of features). That is, testing the null hypothesis of random feature selection – given the claim of clever feature selection – routinely fails.

In our analysis, for each dataset, we compare model's performance when trained on:

1) all features.
2) randomly selected subsets of features.
3) an ensemble trained on different random subsets of features.

For all datasets, where we could find published FS results,

4) we also compare with "important" features identified by the published studies.

We contend that most published claims of feature importance for high-dimensional datasets, especially for gene expression datasets, do not withstand even a basic null hypothesis test. Our findings suggest that high dimensional data - especially human gene expression data - have some very strange characteristics which undermines many conventional assumptions about feature selection.

We realize that this is a very strong claim to make; as such, all of our code and sample datasets are provided at the following GitHub link[2]: https://github.com/bhaveshneekhra/Feature_Selection_HD

[2] We asked multiple other teams to independently rewrite all code and re-test everything, with similar results.

All the datasets used in our experiments are publicly available.

This work makes the following contributions:
(1) We present a large-scale empirical study across 30 high-dimensional datasets to evaluate the effectiveness of randomly selected feature subsets for classification tasks.
(2) We show that randomly selected subsets, sometimes comprising as little as 0.02% - 1% of features, can match or exceed the performance of models trained on all features or on features selected by published studies.
(3) We discuss the broader implications of these findings for evaluating feature selection methods, understanding variance stability and interpreting computational results in genomics and other high-dimensional domains.

## Results

### *Datasets*

We conducted our experiment on 30 high-dimensional datasets, as presented in table 1 (sorted by sample size, with NIPS FS challenge datasets shown at the end). We intentionally chose this heterogeneous collection of datasets spanning multiple cancer types and molecular profiling platforms including microarray, single-cell RNA-Seq and bulk RNA-seq gene expression data. The datasets vary widely in tissue origin, sample size, and class distribution. In addition, we perform our experiments with five benchmark datasets from the NIPS feature selection challenge *(7)*, covering domains such as cancer prediction via mass-spectrometry data, handwritten digit recognition, text classification, and molecular activity prediction and one synthetically generated dataset. Most of the microarray dataset are sourced from *(9)* as they are curated specifically for ML model training. All the datasets are publicly available.

| S No | Name | Samples | Features | Classes | Does a random set match full feature AUC? | Data Type | Application Domain |
|---|---|---|---|---|---|---|---|
| **1** | **Colon** | 62 | 2001 | 2 | within 2% | microarray | colon cancer |
| **2** | **Leukemia (ALL/AML)** | 72 | 7129 | 2 | within 5% | microarray | leukemia |
| **3** | **Ovary (GSE6008)** | 98 | 22283 | 4 | **surpasses significantly** | microarray | ovarian cancer |
| **4** | **Lung (GSE18842)** | 90 | 54675 | 2 | within 5% - 2% | microarray | lung cancer |

| S No | Name | Samples | Features | Classes | Does a random set match full feature AUC? | Data Type | Application Domain |
|---|---|---|---|---|---|---|---|
| **5** | **Throat/Oral (GSE42743)** | 103 | 54675 | 2 | within 5% - 2% | microarray | oral cavity cancer |
| **6** | **Lung (GSE19804)** | 114 | 54675 | 2 | within 2% | microarray | lung cancer |
| **7** | **Prostate (GSE6919_U95B)** | 124 | 12620 | 2 | **matches** | microarray | prostate cancer |
| **8** | **Bowel (GSE3365)** | 127 | 22814 | 3 | within 5% | microarray | inflammatory bowel disease |
| **9** | **Brain (GSE50161)** | 130 | 54575 | 5 | within 2% | microarray | brain tumours |
| **10** | **Breast (GSE22820)** | 139 | 33579 | 2 | **matches** | microarray | breast cancer |
| **11** | **Renal (GSE53757)** | 143 | 54675 | 2 | **surpasses** | microarray | kidney cancer |
| **12** | **Lung (GSE30219)** | 146 | 54675 | 2 | **matches** | microarray | lung cancer |
| **13** | **Colorectal (GSE21510)** | 147 | 54675 | 3 | **matches** | microarray | colorectal cancer |
| **14** | **Breast (GSE45827)** | 151 | 54675 | 6 | **matches** | microarray | breast cancer |
| **15** | **Liver (GSE76427)** | 165 | 47322 | 2 | within 2% | microarray | liver cancer |
| **16** | **Lung (GSE4115)** | 187 | 22215 | 2 | within 2% | microarray | lung epithelial cells cancer |
| **17** | **Colorectal (GSE44076)** | 194 | 49386 | 2 | **matches** | microarray | colorectal cancer |
| **18** | **Colon (GSE11223)** | 202 | 40991 | 3 | **matches** | microarray | colon inflammation |
| **19** | **Leukemia (GSE28497)** | 281 | 22284 | 7 | within 2% | microarray | pediatric leukemia |
| **20** | **Breast (GSE70947)** | 289 | 35981 | 2 | within 5% | microarray | breast cancer |
| **21** | **Liver (GSE14250)** | 357 | 22277 | 2 | **surpasses** | microarray | hepatocellular carcinoma |
| **22** | **TCGA (LUAD/LUSC)** | 1016 | 20253 | 2 | **matches** | bulk RNA-Seq | lung cancer |
| **23** | **TCGA** | 10223 | 20253 | 33 | **matches** | bulk RNA-Seq | pan-cancer |

| S No | Name | Samples | Features | Classes | Does a random set match full feature AUC? | Data Type | Application Domain |
|---|---|---|---|---|---|---|---|
| **24** | **Lung (ScRNA-Seq)** | 20966 | 33514 | 9 | within 5% | scRNA-Seq | lung adenocarcinoma |
| **25** | **Lung (ScRNA-Seq)** | 24421 | 33514 | 2 | within 5% | scRNA-Seq | lung adenocarcinoma |
| | | | | | | | |
| **26** | **Arcene** | 200 | 10000 | 2 | **surpasses** | mass-spectometry | ovarian vs prostate cancer |
| **27** | **Dexter** | 600 | 20000 | 2 | No | text | Reuters text categorization |
| **28** | **Dorothea** | 1150 | 100000 | 2 | **matches** | binary fingerprint | drug activity prediction |
| **29** | **Madelon** | 2600 | 500 | 2 | No | synthetic | cluster classification |
| **30** | **Gisette** | 7000 | 5000 | 2 | **matches** | image | digit classification (4 vs 9) |

**Table 1. Datasets used in our experiments.** Comparison of random subsets with all features using Random Forest

### *Models*

To test our hypothesis, we experimented mainly with the Random Forest model.
However, our code allows to experiment with various classification models including:

1. Logistic Regression (LR)
2. Support Vector Machine (SVM)
3. Decision Tree (DT)
4. eXtreme Gradient Boosting (XGB)
5. Gradient Boosting Classifier (GBM)
6. HistGradient Boosting Classifier (HistGB)
7. Ridge Classifier
8. SGDClassifier
9. Multilayer Perceptron (MLP)

These models were selected for two main reasons. First, they represent a diverse set of learning algorithms, ranging from linear models (LR, SVM) to tree-based ensembles (RF, XGB) and neural networks (MLP), providing a comprehensive evaluation across modeling paradigms.

Second, several related studies have employed one or more of these models in their analyses. By aligning our model choices with theirs, we facilitate meaningful comparison and benchmarking of results.
In the following subsection, we present our results with Random Forest models using different datasets.

***Random Forest results with cross-dataset evaluation***

To test our null hypothesis, we use two **microarray** datasets with identical features: one for training and another for testing model performance. Here, we report the Area Under the Curve (AUC) and include accuracy results in the supplementary material. We then repeat the experiment by swapping the roles of the two datasets.

To illustrate, we first train a model using GSE18842 *(9)* (90 samples, 54,765 features) and test it using GSE19804 *(9)* (114 samples, 54,765 features). As shown in Fig. 1(A), randomly selected subsets of just 200 features (~0.4% of all features), selected without replacement, achieve AUC comparable to using all features. Fig. 1(B) shows the reverse setting, where the model is trained on GSE19804 and tested on GSE18842. Both plots show similar results. The 2D projections of these two datasets using PCA and t-SNE show that GSE18842 exhibits better class separability than GSE19804. We hypothesise that such low-dimensional separability enables models trained on small random subsets of features to perform well. Accuracy plots for this dataset pair show similar trends in the supplementary Fig. S1: even very small random subsets can match the performance of models using all features.

Results with another dataset pair, Gisette, with cross-dataset evaluation are shown in Fig. 2. The Gisette image dataset is from NIPS 2003 FS challenge *(7)*. The challenge organisers provided a separate train and validation set. We find that with <200 (<4%) randomly selected features, models could match the AUC with all features. The results are significant as the dataset is from the vision domain.

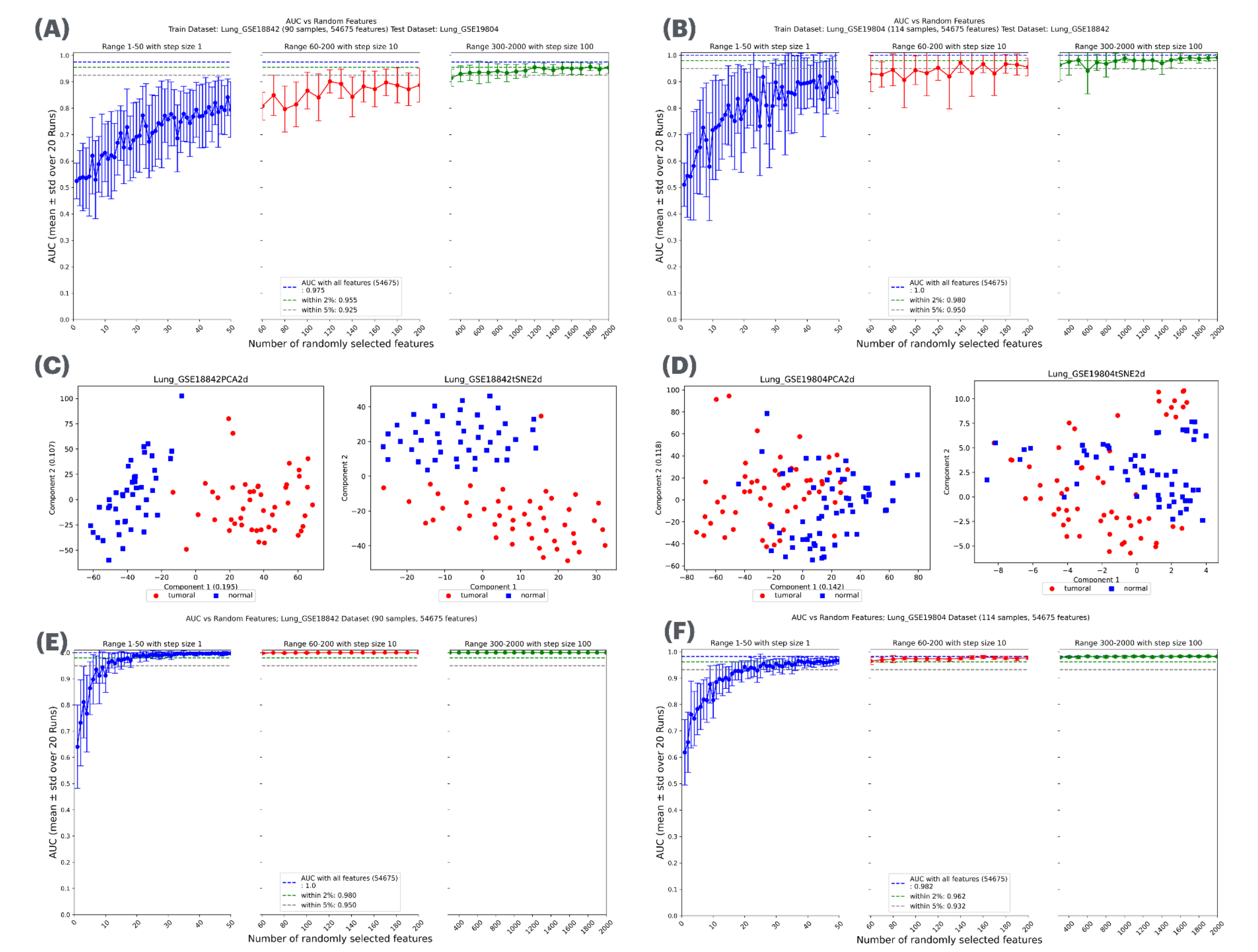

**Fig. 1. Random Forest (RF) performance on lung cancer microarray datasets** (mean and standard deviation reported over 20 runs)

(A) RF models trained on GSE18842 and tested on GSE19804 show that randomly selected subsets of just 200 features (~0.4% of all features) achieve AUC comparable to using all features.

(B) RF models trained on GSE19804 and tested on GSE18842 show that 400 randomly selected features (~0.7% of all features) perform comparable to all features.

(C) & (D) PCA and t-SNE plots showing class separation in GSE18842 and GSE19804, respectively.

(E) & (F) Model performance with an 80:20 train-test split using randomly selected feature subsets. For GSE18842, just 20 randomly selected features (~0.04% of all features) are sufficient to match the AUC comparable to all features. Similarly, for GSE19804, 100 features (~0.2% of all features) suffice.

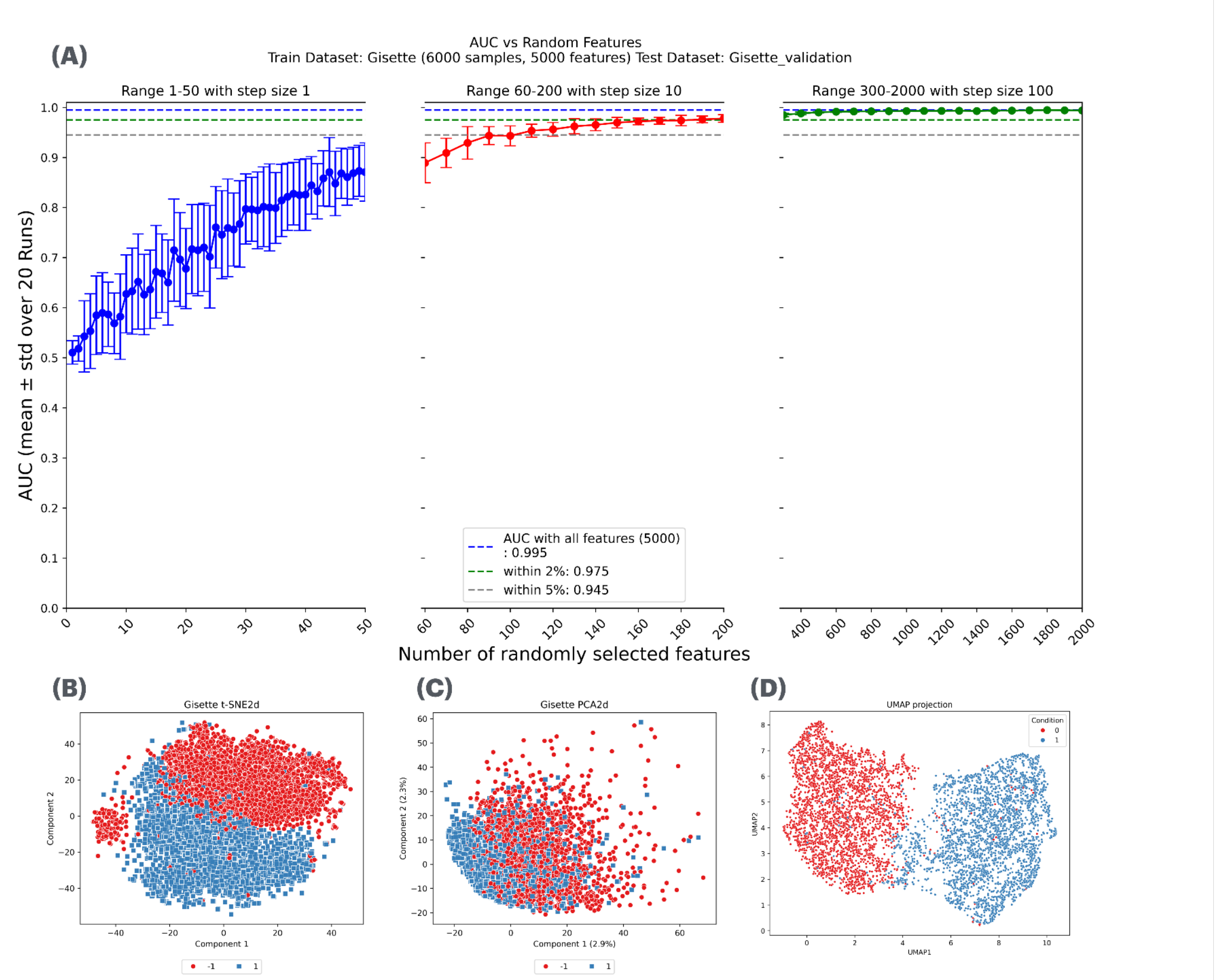

**Fig. 2. Random Forest performance with Gisette image dataset** (mean and standard deviation are reported over 20 runs)
The task of GISETTE is to discriminate between two confusable handwritten digits: the four and the nine.
(A): Model trained on Gisette train dataset and tested on Gisette validation dataset shows that randomly selected subsets of just 90 (~1.8% of all features) achieve AUC comparable to using all features.
(B), (C) & (D): PCA, t-SNE and UMAP plots showing class separation in Gisette training dataset.

In this section we have shown that for two distinct train-test dataset pairs, randomly selected subsets of features can match - and in some cases exceed - the performance of models trained on all features. As the size of the random subset increases, performance generally improves. However, a more striking observation is the consistent reduction in performance variance across 20 independent runs for each subset size. Notably, even when features are selected without replacement (a setting used when the total number of features exceeds 20,000), random subsets yield reliable performance with decreasing variance. Thus our results challenge the common assumption that higher dimensionality leads to improved classification performance. They also raise important questions about redundancy and noise in high-dimensional datasets, motivate further investigation into the structure of feature spaces and the role of feature selection - even when done randomly. In the next section, we show that similar results hold for intra-dataset 80:20 splits across all 30 datasets.

### ***Random Forest results with intra-dataset splits***

In addition to cross-dataset evaluation, we assessed model performance using standard intra-dataset train–test splits of 80:20. For each dataset, we randomly divided the samples into 80% for training and 20% for testing, while ensuring class balance was preserved. For example, figures 1(E) and 1(F) show that for GSE19804 and GSE18842, with a random subset we can match the performance of all features.

Figure 3 shows results for **bulk and Single-cell RNA-Seq dataset**. For the bulk RNA-Seq data, as shown in Fig. 3(A), Random Forest models trained with just 50 randomly selected features (~0.22% of all features) can match the performance of the full feature set. For the Single-cell RNA-Seq dataset, Fig. 3(E) shows that with ~1200 (~3.6% of all features) randomly selected features, random forest gets AUC within 5% of the full-feature one. Again, we hypothesise that better low-dimensional separability for TCGA_LUAD_LUSC *(10)* (bulk RNA-Seq dataset) enables random subsets to perform better. Even for a bulk RNA-Seq dataset like TCGA *(10)* with 33 classes, a random subset is able to perform as well as all features as shown in supplementary Fig. S24.

We observe similar patterns in three of the five benchmark datasets from the NIPS 2003 Feature Selection Challenge *(7)*. Consistent with the cross-dataset results, we found that models trained on extremely small subsets of randomly selected features often outperformed those trained on the full feature set. In several cases, using just 0.02% - 0.1% of the available features led to improved accuracy and AUC, with performance saturating well before reaching the full dimensionality. Table 1 summarises the results across all 30 datasets, showing that for 28 datasets this pattern holds consistently across diverse data types, including microarray, single-cell RNA-seq, bulk RNA-Seq and benchmarks datasets from NIPS FS challenge. The only two cases where we fail are in journalistic text data categorization and drug activity prediction; we succeed in all standard biology domains. Results for these two datasets, Dexter and Dorothea *(7)*, are shown in Fig. S27 and S28 respectively.

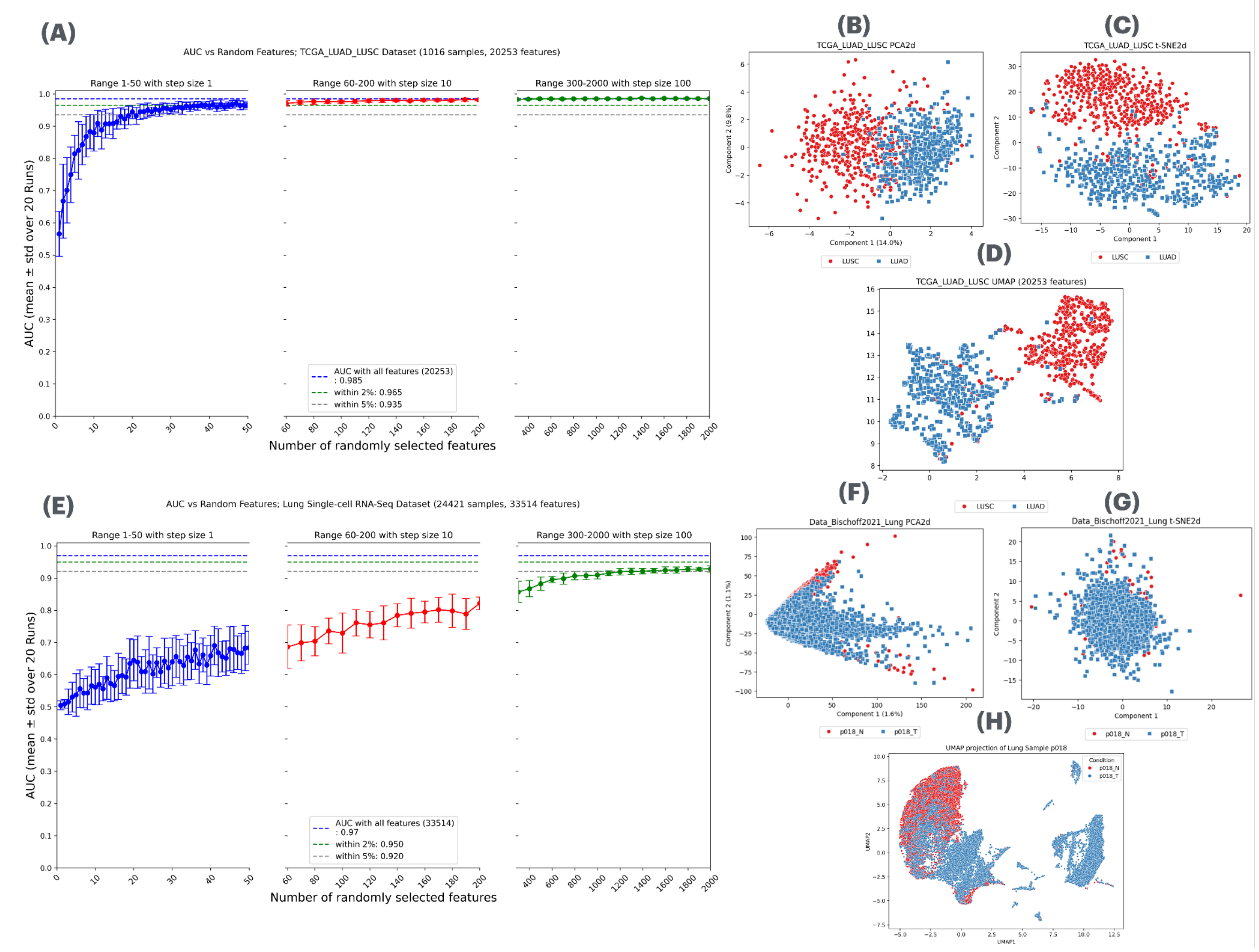

**Fig.3. Random Forest performance with bulk RNA-Seq and Single-cell RNA-Seq datasets** (mean and standard deviation are reported over 20 runs)

(A) models trained and tested on TCGA-LUAD-LUSC bulk RNA-Seq dataset (80:20 split) shows that a random subset of size 50 (<0.3%) is able to match AUC with all features.

(B), (C) & (D) PCA, t-SNE and UMAP plots showing class separation.

(E) On the lung cancer SIngle-cell RNA-Seq dataset (80:20 split), random subsets of features achieve test AUC within 5% of the full-feature model.

(F), (G) & (H) PCA, t-SNE and UMAP plots showing class separation.

In the next subsection, we show that the model's performance with "important" features from published studies is comparable to the baseline of random subsets of features- questioning the utility and statistical significance of such exercise.

### *Comparison with published studies on feature selection*

Table 2 provides a comparison of results for selected datasets and related published studies. The first column lists the published study followed by the dataset. The next 4 columns list accuracy with (1) all features, (2) "important" features from the published study, (3) randomly selected features of the same size as "important" features and, (4) ensemble with random subsets of features of same size as "important" features. The last column (5) lists the results with an ensemble when the constituent models are trained with 1% of all features. For example, as shown in Row 3, a dataset with 7129 features, and a published study that selects 35 of those using FS, the 4th column would provide the accuracy with 35 randomly selected features in an ensemble and the 5th column would do the same thing with 1%( =71) features. Figure 3 displays the bar plots for the same set of results.

| Publication | Dataset | Model Accuracy (No. of features) | | | | |
|---|---|---|---|---|---|---|
| | | All Features | "selected" features from the paper | Randomly selected features | Ensemble of (LR, RF, XGB)- 3 variation of each model | Ensemble of (LR, RF, XGB)- 3 variation of each model with **~1%** randomly selected |
| **Chen & Dhahbi 2021** | TCGA (LUAD/ LUSC) | 94.8 (20253) | 94.2 (500) | 93.5 (100) | **95.4** (500) | 95.1 (203) |
| **Li et al. 2017** | TCGA | 94.6 (20253) | **95.6** (50) | 87.2 (50) | 92.8 (50) | 95.02 (200) |
| **Lall & Bandyopadhyay 2020** | ALL/AML | **94.3** (7129) | 86 (35) | 79 (35) | 90.1 (35) | 91.4 (71) |
| **Golub et al. 1999** | ALL/AML | 94.3 (7129) | **98.6** (51) | 86 (51) | 93 (51) | 91.4 (71) |
| **Cilia et al. (2019)** | ALL/AML | 94.3 (7129) | **99.4** (51) | 86 (51) | 96 (200) | 91.4 (71) |
| **Zanella et al. 2022** | GSE4115 | 68.9 (22215) | 67.93 (5) | 58.6 (5) | 64.7 (5) | **73.3** (222) |

**Table 2. Comparison of model accuracy with published studies across multiple datasets.** The highest accuracy per dataset is shown in **bold** and the second-highest is underlined. The final two columns show performance of our ensemble models using random subsets of features, including a configuration with ~1% randomly selected features.

Row 1 shows that for the TCGA(lung) dataset, any randomly selected subset of 100 features (< 0.5% of all features) performs comparably to feature selection. When using an ensemble where constituent models are trained on different random subsets, the model's performance exceeds that of models trained on all features or on "important" features selected by published methods. In five out of six rows, the ensemble of random feature subsets ranks among the top two performers and is comparable to models using "important" features. These results suggest that randomly selected subsets can capture as much discriminatory signal as all features or those deemed "important" by established feature selection methods.

Results for other datasets with Random Forest can be accessed in the supplementary materials- showing similar patterns. Table 1 summarises the results for all 30 datasets used in our experiments.

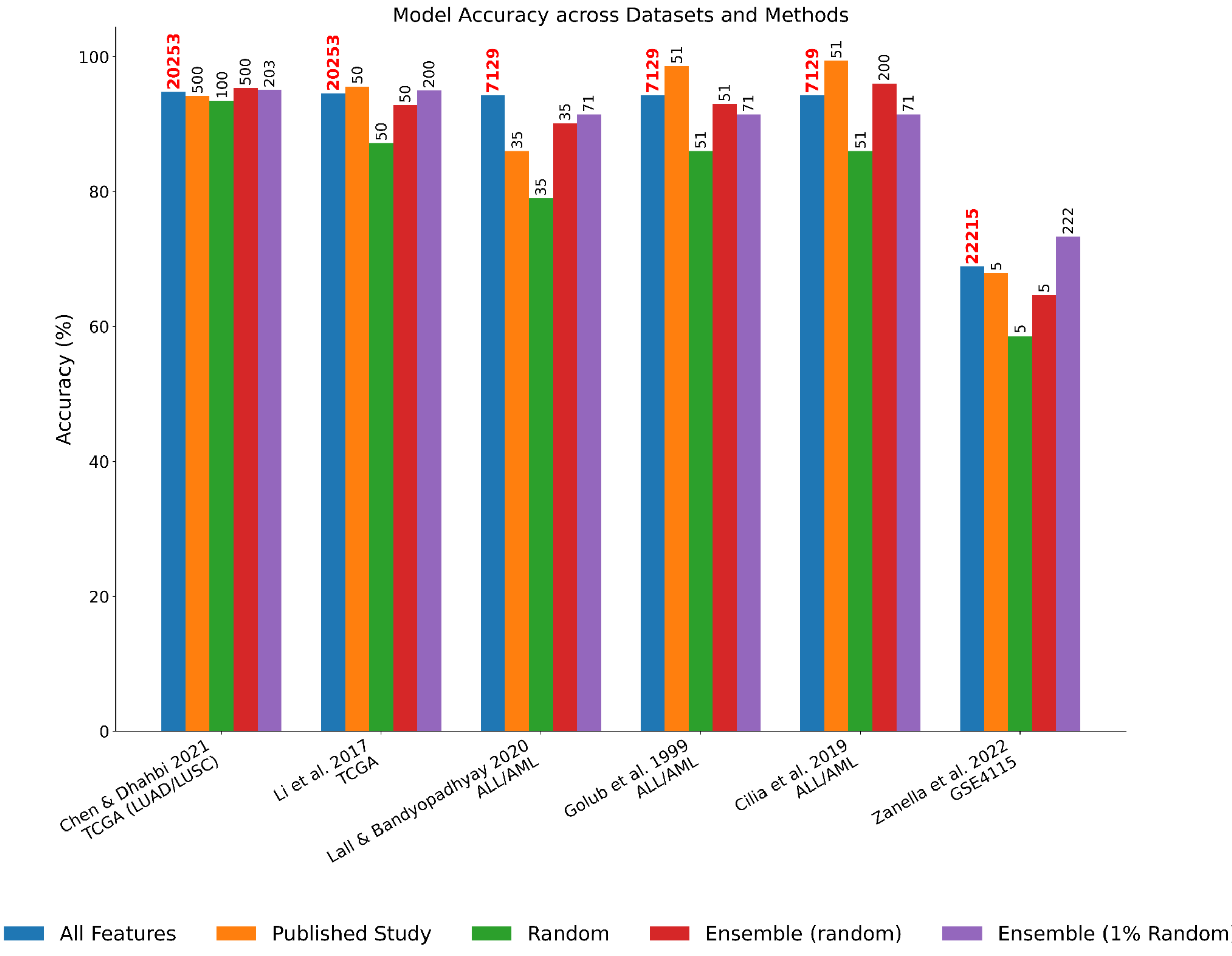

**Fig.3. Comparison of model accuracy with published studies across multiple datasets.** Feature counts are shown above bars and full feature counts are highlighted in red.

### *Impact of Dataset Characteristics on Random Subset Performance*

We observe that the success of randomly selected feature subsets in matching or surpassing the performance of selected features appears strongly dependent on the dataset's characteristics, especially the number of samples $n$, number of features $p$, and number of classes $k$. Specifically:

- Datasets with **low sample size and high dimensionality** (low $n$, high $p$) often suffer from overfitting when using complex feature selection algorithms. In such cases, random subsets may generalize better by avoiding over-specialization.
- In **high-dimensional datasets** (large $p$), many features may be redundant or weakly informative. This increases the likelihood that a random subset captures sufficient signal for classification.
- **Binary classification tasks** (low $k$) are inherently less complex, making it easier for random feature sets to achieve performance comparable to carefully selected ones.

- While we are not entirely certain, we suspect that one contributing factor to the success of random selection is: in datasets where the **first few principal components explain a large fraction of total variance**, the data likely has an intrinsically low-dimensional structure. This suggests that an important signal is diffusely spread across the feature space, making random selection surprisingly effective.

These observations underscore the need to examine dataset characteristics - such as class separability, redundancy, and feature informativeness - when evaluating feature selection methods.

***Discussion***

Our findings challenge the conventional, intuitive assumption that more features, and carefully (or cleverly) selected features lead to better classification performance in high-dimensional settings. Across 30 diverse datasets, for 28 out of 30, we observe that randomly selected feature subsets - sometimes comprising as little as 0.02% of all features - can match or even outperform models trained on the full feature set or on FS sets. This result holds across both cross-dataset and intra-dataset validation and spans multiple data types, including gene expression profiles (microarray, bulk and single-cell RNA-seq), text, and mass-spectrometry datasets.

These results reinforce earlier insights on feature redundancy in high-dimensional data and resonate with prior work showing that Random Forests are robust to noise and overfitting due to their ensemble nature and internal feature sampling mechanisms *(12, 13)*. Our findings extend these observations by demonstrating that even explicit external subsampling of features - performed entirely at random - can yield stable and often superior classification performance. This robustness is further supported by our observation that the standard deviation in performance *decreases consistently with subset size*, suggesting the presence of many equally informative, often non-overlapping, feature combinations.

This phenomenon may have conceptual ties to random subspace methods *(14)* and the theory of random projections, particularly the Johnson–Lindenstrauss lemma *(15)*, which shows that high-dimensional data can be embedded in lower dimensions while preserving pairwise distances. While our method does not perform explicit projections, the empirical success of randomly selected subspaces suggests that useful structure can be retained without sophisticated transformations. Unlike methods such as Principal Component Analysis (PCA), which apply global, often opaque transformations, random feature selection is both conceptually and computationally simple: the only surprising thing is that it works so well!

Our results also raise critical questions regarding the interpretation of feature importance in gene expression datasets. While computational feature selection methods are often used to identify genes of interest, our findings show that a "typical"-size set of randomly selected genes can achieve classification performance comparable to that of models trained on sets of "important" genes identified by published studies. This suggests that computationally derived feature importance may reflect statistical signals more than underlying biological causality. We do not argue against the search for biologically meaningful genes; on the contrary, we emphasize that identifying causal or mechanistically relevant genes requires biological validation. Features identified as important by computational models should ideally be corroborated through independent experimental methods, such as perturbation assays or wet-lab validation.

These findings open multiple directions for future research. In this study, we have already demonstrated that random feature subsets serve as a strong baseline for evaluating the added value of sophisticated feature selection algorithms. In many cases, the performance of random subsets was comparable to - or even exceeded - that of models trained on computationally selected "important" genes. Investigating the diversity and overlap among high-performing random subsets may be interesting and reveal deeper insights into the intrinsic dimensionality, redundancy, and structure of high-dimensional biological data. Finally, incorporating random subspace strategies into ensemble learning or active learning pipelines could enhance both performance and generalization, especially in domains like genomics where data is high-dimensional and sample sizes are often limited.

Overall, our study highlights the underappreciated potential of simple, randomized strategies in high-dimensional learning. By showing that stable and accurate classification can be achieved with minimal, randomly chosen features, we call for a reassessment of how feature selection and dimensionality are approached in feature-rich, sample-poor environments—especially in computational biology.

**Acknowledgements:** We sincerely thank Subhashis Banerjee, Shubhasis Haldar, Shreyansh Priyadarshi, Swapnanil Mukherjee, Mayank Garg, Naveen Kumar Bhatraju and Vikas Sahni for discussion and feedback. Special mention to Swapnanil Mukherjee, Aritra Das and Neil Chaudhary for helping with independently reproducing and verifying the results.

**Data and materials availability:** All datasets used are publicly available. The full codebase with one pre-processed dataset (Leukemia), to reproduce the results, are hosted at the GitHub repo: https://github.com/bhaveshneekhra/Feature_Selection_HD

# Supplementary Materials for

## On the (In)Significance of Feature Selection in High-Dimensional Datasets

Bhavesh Neekhra*, Debayan Gupta*, Partha Pratim Chakrabarti

*Corresponding author: debayan.gupta@ashoka.edu.in, bhavesh.neekhra_phd18@ashoka.edu.in

**Materials and Methods**

<u>Datasets</u>

We used a diverse set of 30 high-dimensional datasets across different domains and data types (see Table 1). These included 21 microarray gene expression datasets, 4 single-cell or bulk RNA-seq datasets, 1 mass spectrometry dataset, 2 image datasets, and 2 others (synthetic and binary fingerprint). All datasets contain thousands to tens of thousands of features, and span a range of sample sizes and classification complexities. The datasets represent various cancers, tissue types, and experimental conditions. Preprocessing followed dataset-specific conventions; where applicable, raw values were log-transformed and normalized using z-score scaling.

<u>Experimental Setup</u>

For each dataset, we compared classification performance using

(1) All features

(2) Random Feature Subsets (baseline)

All datasets were split into train/test subsets using an 80/20 stratified split. Features were standardized before selection or model training.

Additionally, for all datasets, where we could find published FS results, we also compare with

(3) “important” features identified by such studies.

<u>Random Feature Subsets</u>

We sampled subsets of features of various sizes (1, 2 ,... 50; 60, 70,... 200; 300, 400… 2000) at random (without replacement when total features > 20,000) resulting in 83 subsets. For each subset, we repeated the process 20 times using different random features. A Random Forest classifier was trained and evaluated for each random subset. We report average accuracy and AUC across 20 runs. (Our code allows experimenting with other classification models such as LR, SVM, DT, XGB, MLP etc. )

## Evaluation Metrics

We used Accuracy and Area Under the ROC Curve (AUC-ROC) as the primary metrics. For random subsets, we report the mean performance over 20 runs. Each plot has three horizontal reference lines, one for Accuracy (or AUC) with all features, and other two for within-2% and within-5% Accuracy (or AUC).

## Software and Environment

All experiments were conducted using Python 3.11 and the following libraries:
scikit-learn: 1.5.2
numpy: 1.26.4
pandas: 2.2.0
matplotlib: 3.8.0
seaborn: 0.13.2
Experiments were performed on the M4 macbook Pro with 16 GB RAM. Random seeds were fixed (np.random.seed(42) and random_state=42) for reproducibility.

## Reproducibility

The full codebase, dataset loaders, and experimental scripts are available at the following anonymous github link : https://github.com/bhaveshneekhra/Feature_Selection_HD

## Fig. S1.

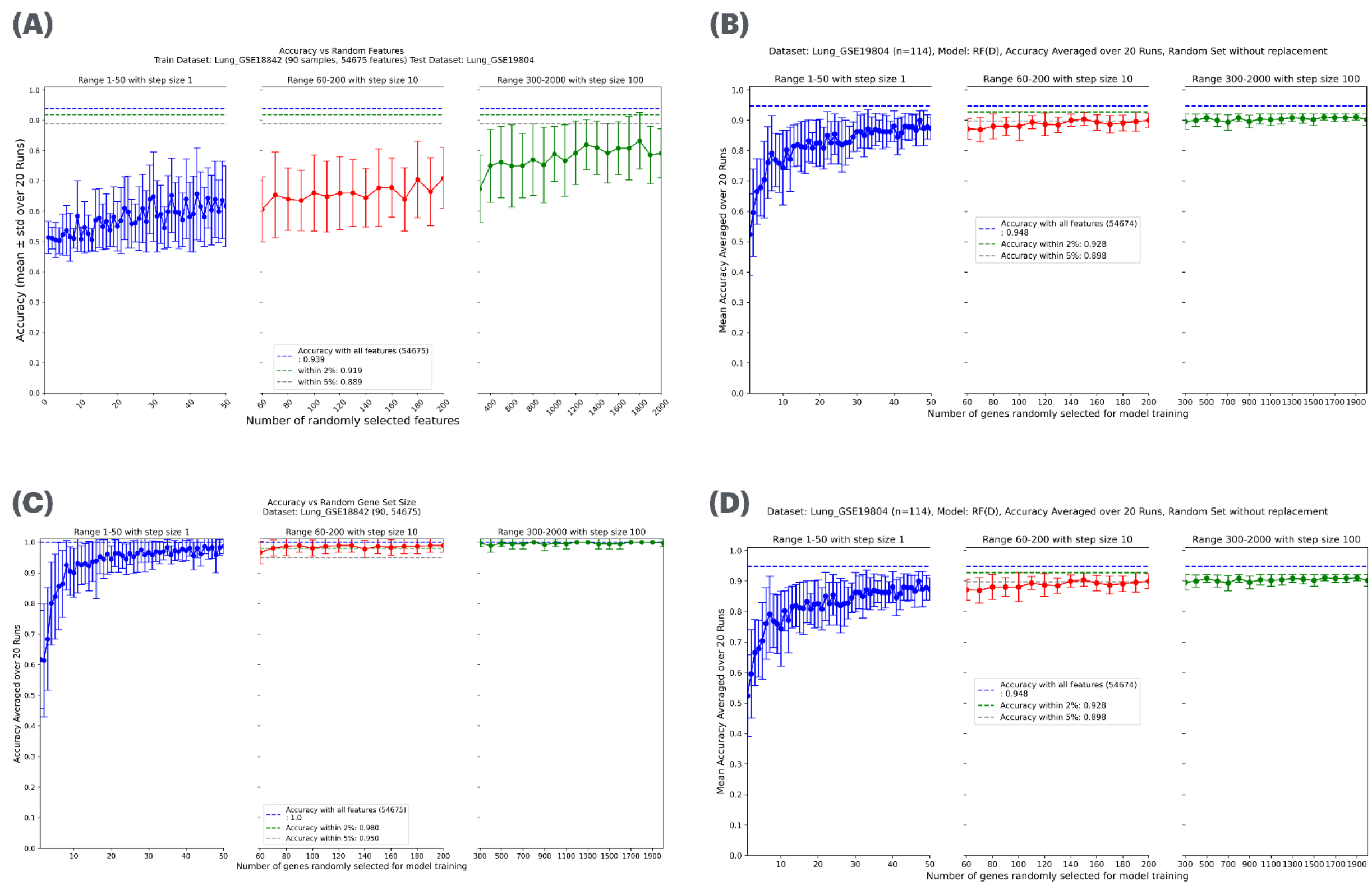

**Random Forest performance with lung microarray dataset pairs** (mean and standard deviation are reported over 20 runs) (A) RF models trained on GSE18842 and tested on GSE19804 show that randomly selected subsets never achieve accuracy comparable to using all features. (B) RF models trained on GSE19804 and tested on GSE18842 show that 200 randomly selected features (~0.4% of all features) perform comparable to all features. (C) Model performance with an 80:20 train-test split using randomly selected feature subsets. For GSE18842, just 50 randomly selected features are sufficient to match the accuracy comparable to all features. (D) Similarly, for GSE19804, 200 features suffice to match accuracy with all features.

## Fig. S2.

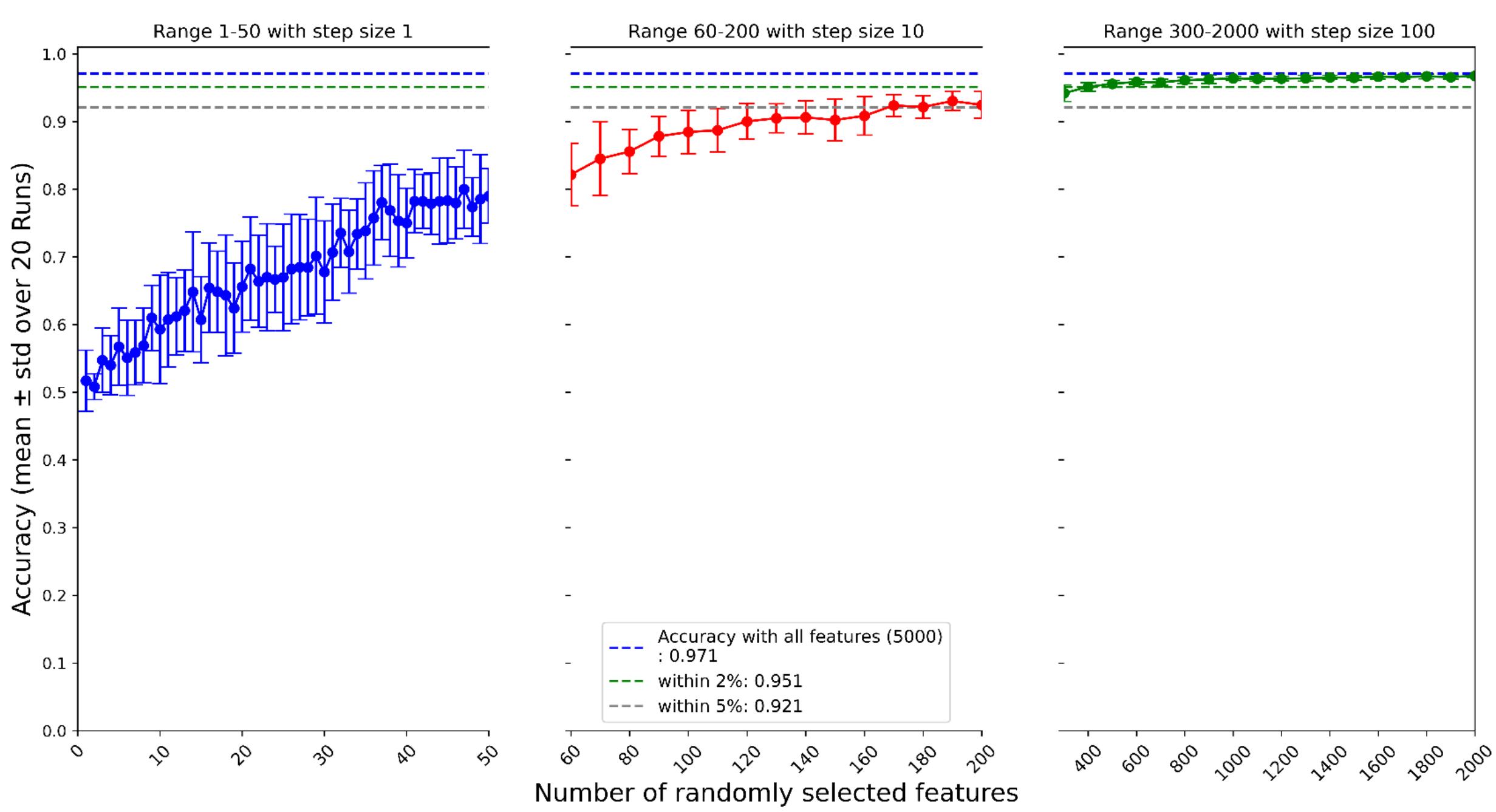

**Random Forest performance with Gisette image dataset** (mean and standard deviation are reported over 20 runs)
The task of GISETTE is to discriminate between two confusable handwritten digits: the four and the nine. The model is trained on Gisette train dataset and tested on Gisette validation dataset shows that randomly selected subsets of just 200 achieve accuracy comparable to using all features.

**Fig. S3.**

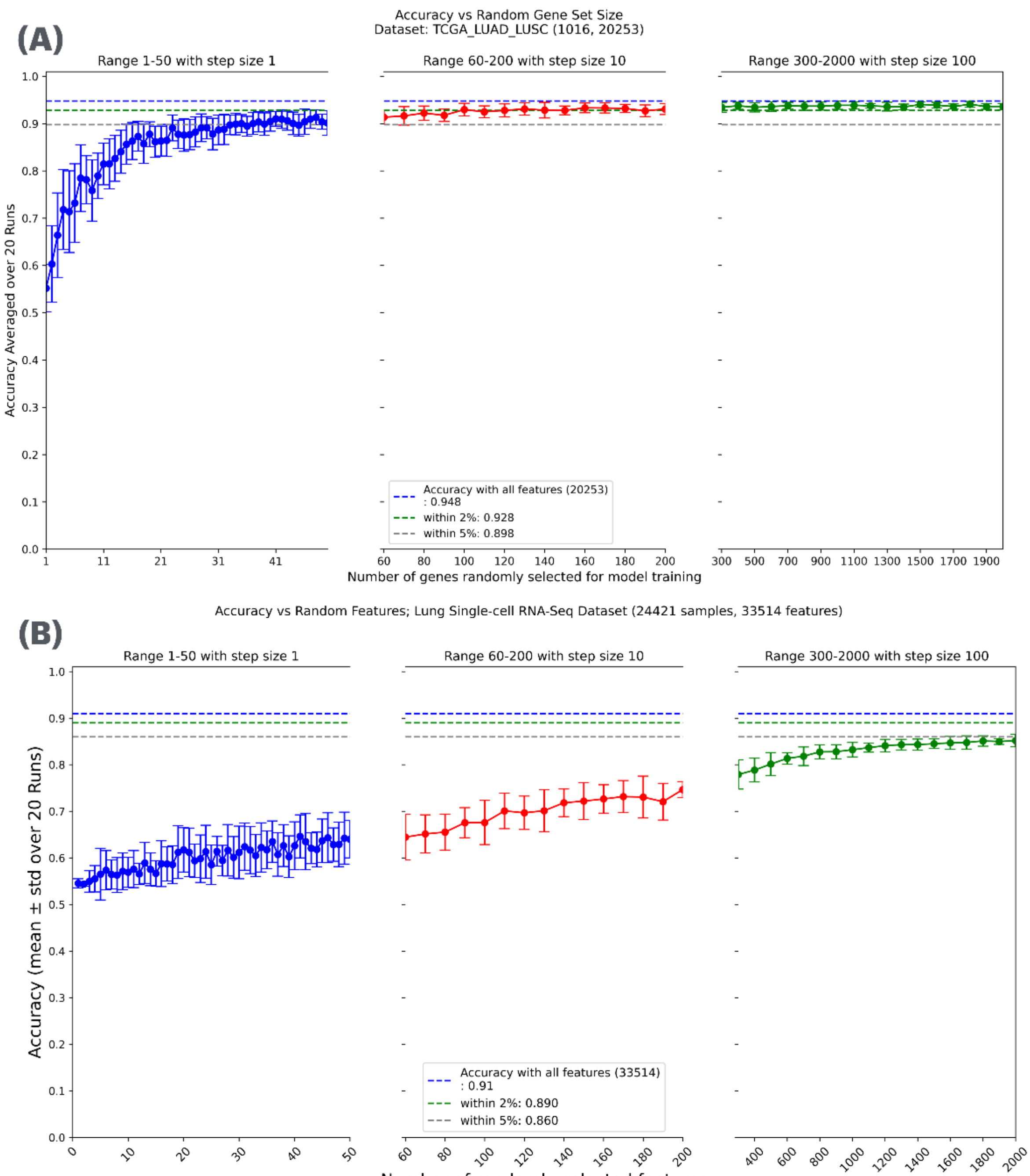

**Random Forest performance with bulk RNA-Seq and Single-cell RNA-Seq datasets** (mean and standard deviation are reported over 20 runs) (A) models trained and tested on TCGA-LUAD-LUSC bulk RNA-Seq dataset (80:20 split) shows that a random subset of size 50 (<0.3%) is able to match within-5% accuracy of all features. (B) On the lung cancer SIngle-cell RNA-Seq dataset (80:20 split), randomly selected subsets of size 2000 achieve accuracy within 5% of the full-feature model.

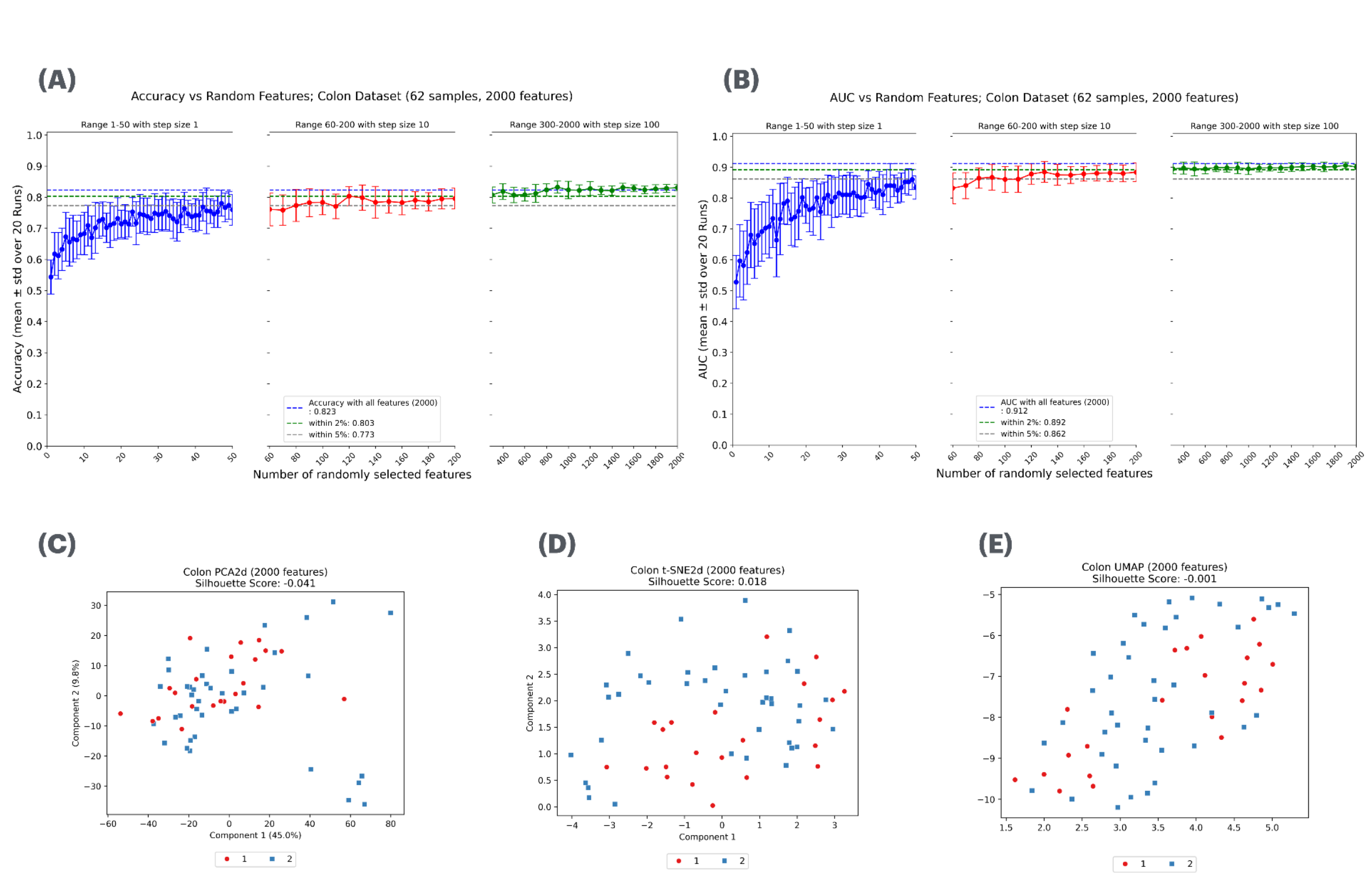

**Random Forest performance with Colon microarray dataset** (mean and standard deviation are reported over 20 runs) (A) & (B) models trained and tested on 80:20 split shows that a random subset of size ~100 is able to match accuracy and AUC with all features, respectively. (C) & (D) & (E) PCA, t-SNE and UMAP plots showing class separation.

## Fig. S5.

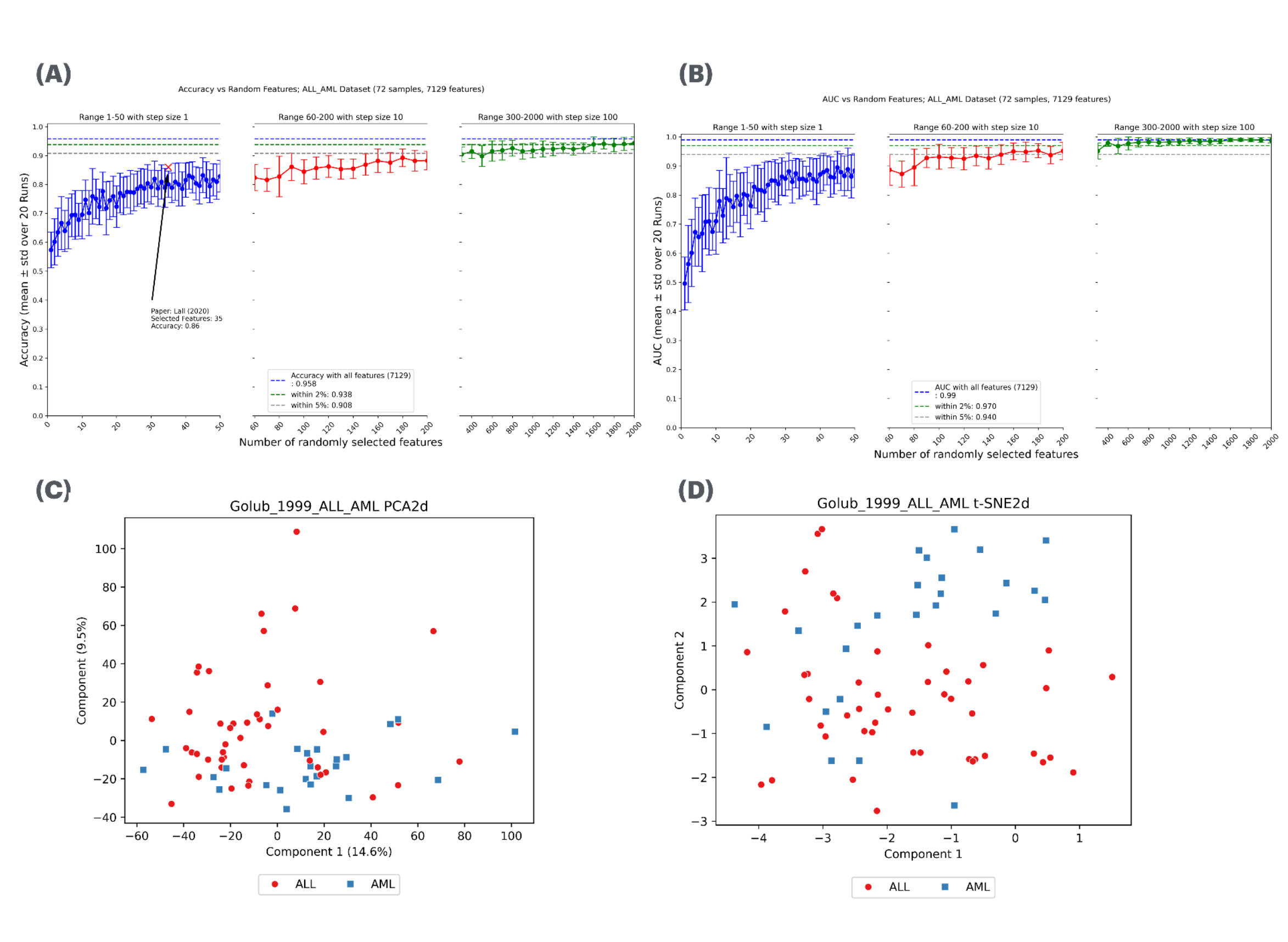

**Random Forest performance with ALL/AML Leukemia microarray dataset** (mean and standard deviation are reported over 20 runs). (A) & (B) models trained and tested on 80:20 split shows that a random subset of size ~200 is able to match accuracy and AUC with all features, respectively.  (C) & (D) PCA, t-SNE plots showing class separation.

## Fig. S6.

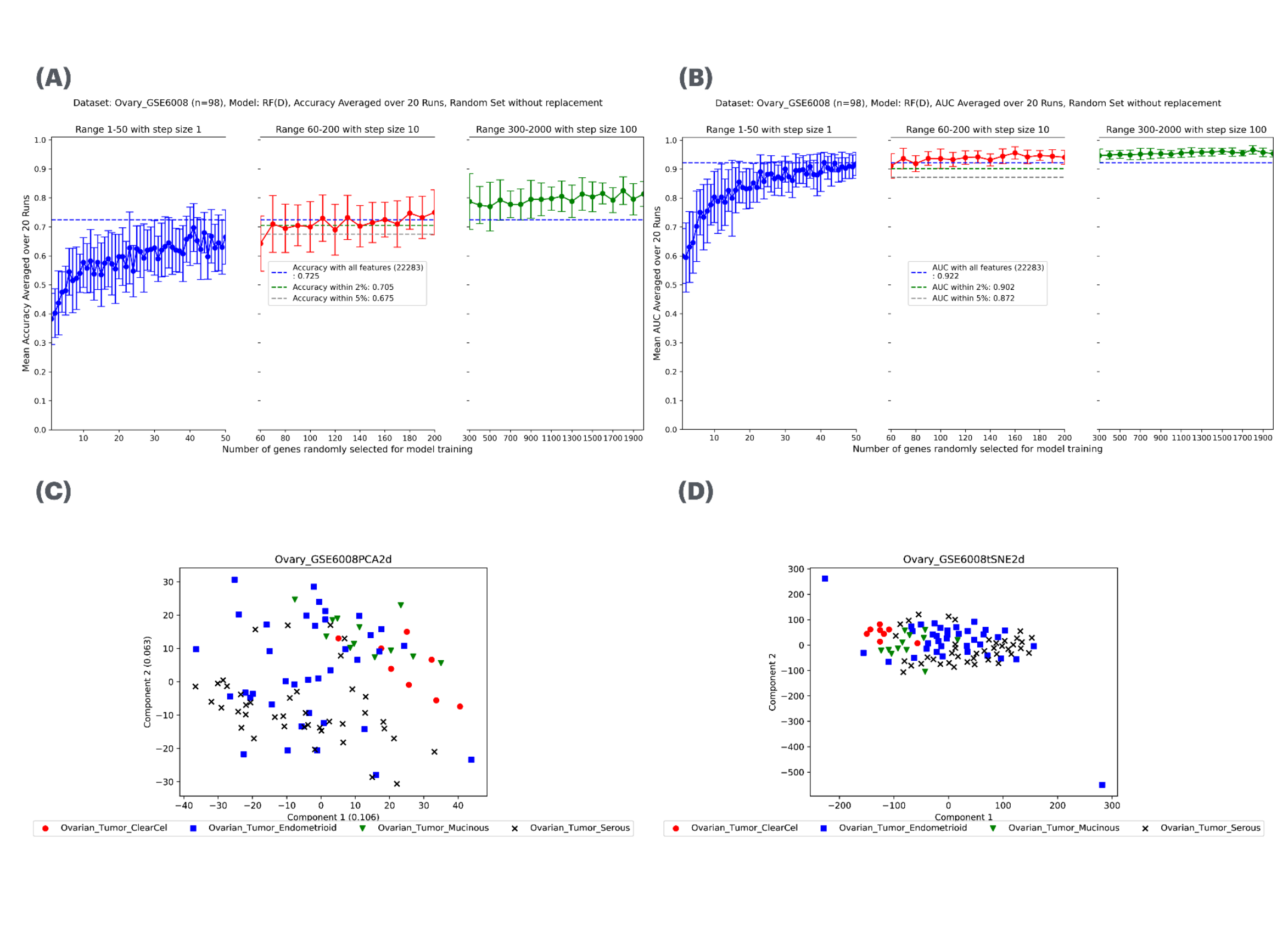

**Random Forest performance with Ovary (GSE6008) microarray dataset** (mean and standard deviation are reported over 20 runs). (A) & (B) models trained and tested on 80:20 split shows that a random subset is able to match accuracy and AUC with all features, respectively. (C) & (D) PCA, t-SNE plots showing class separation.

**Fig. S7.**

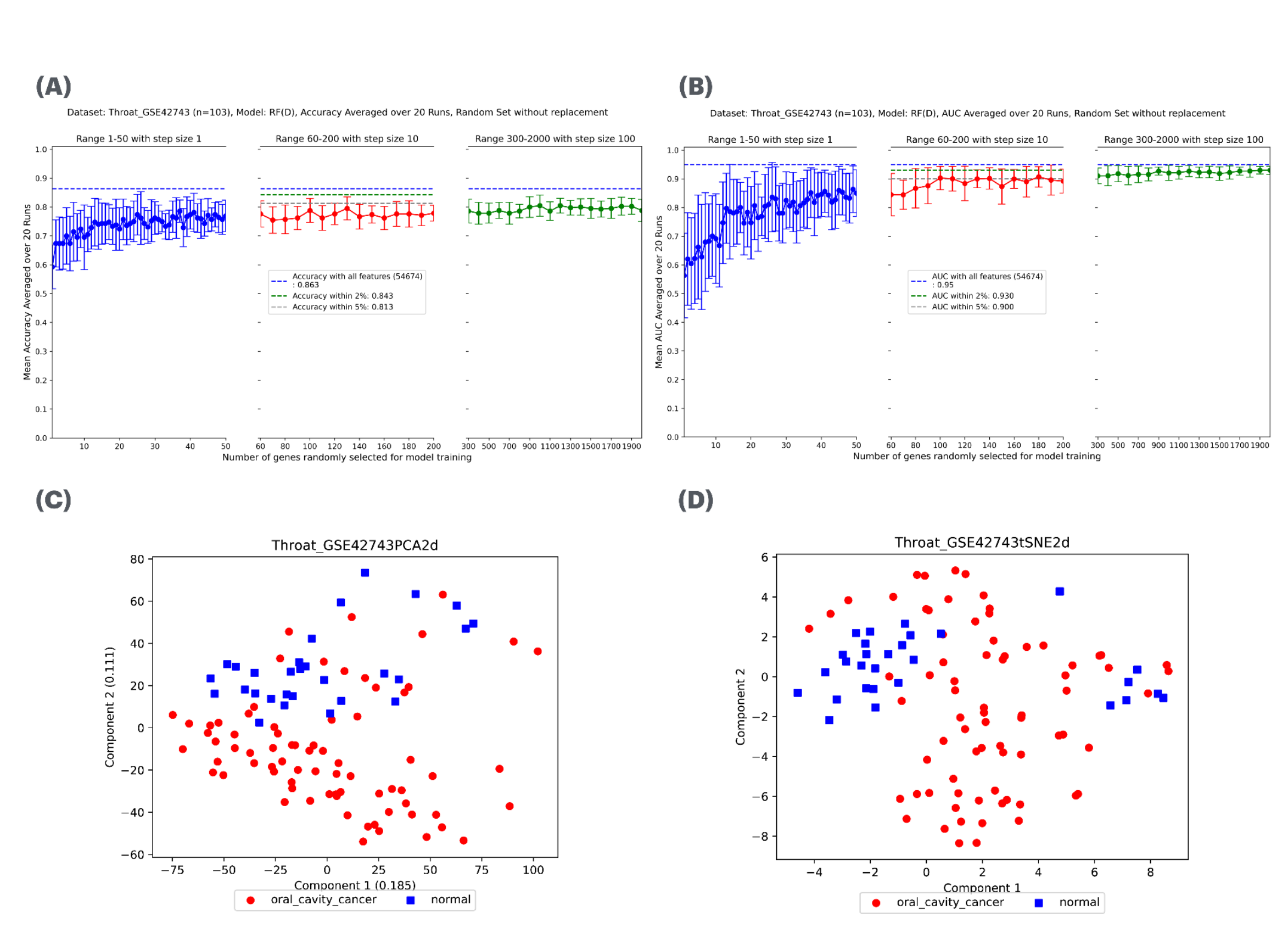

**Random Forest performance with Oral/Throat (GSE42743)) microarray dataset** (mean and standard deviation are reported over 20 runs). (A) & (B) models trained and tested on 80:20 split shows that a random subset is able to match within-5% accuracy and AUC with all features, respectively. (C) & (D) PCA, t-SNE plots showing class separation.

**Fig. S8.**
**(A)**

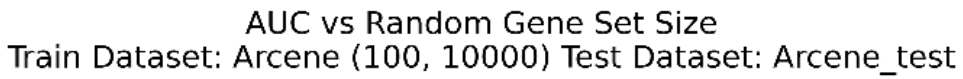

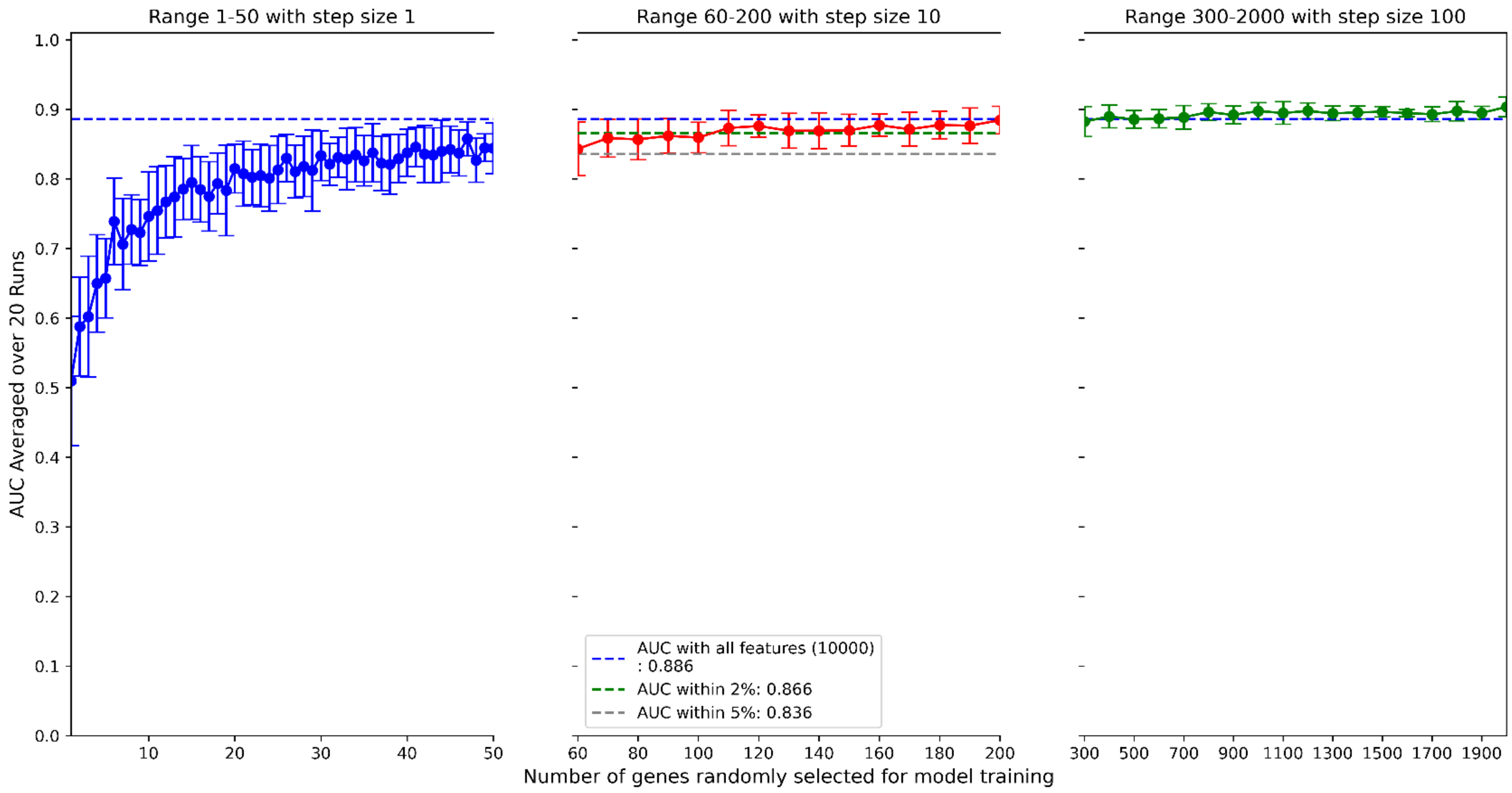

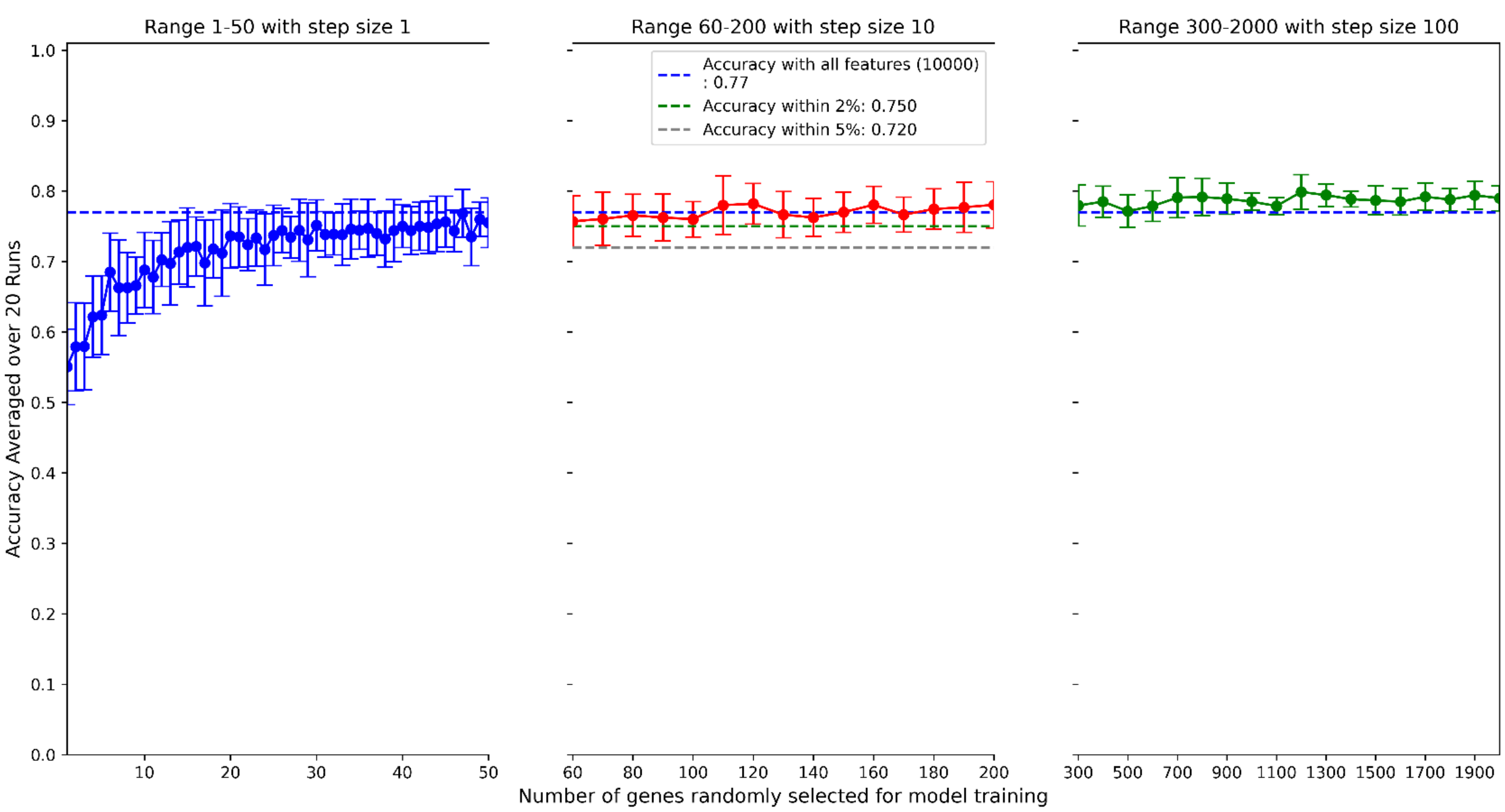

**(B)**

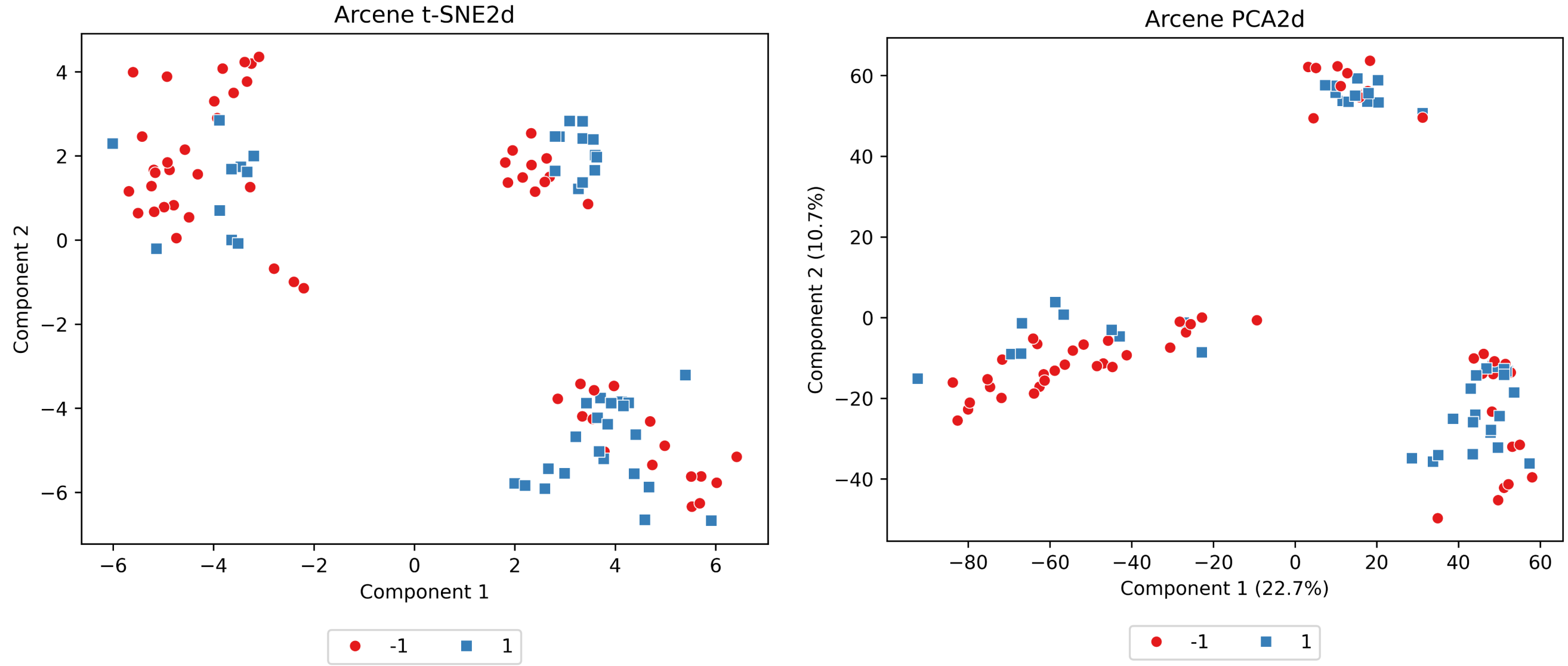

**Random Forest performance with Arcene mass-spectrometry dataset** (mean and standard deviation are reported over 20 runs). The task of ARCENE is to distinguish cancer versus normal patterns from mass-spectrometric data. (A) Models trained and tested on 80:20 split shows that a random subset of size ~50 (0.5% of all features) is able to match within-5% Accuracy and AUC of all features. (B) PCA, t-SNE plots showing class separation.

**Fig. S9.**

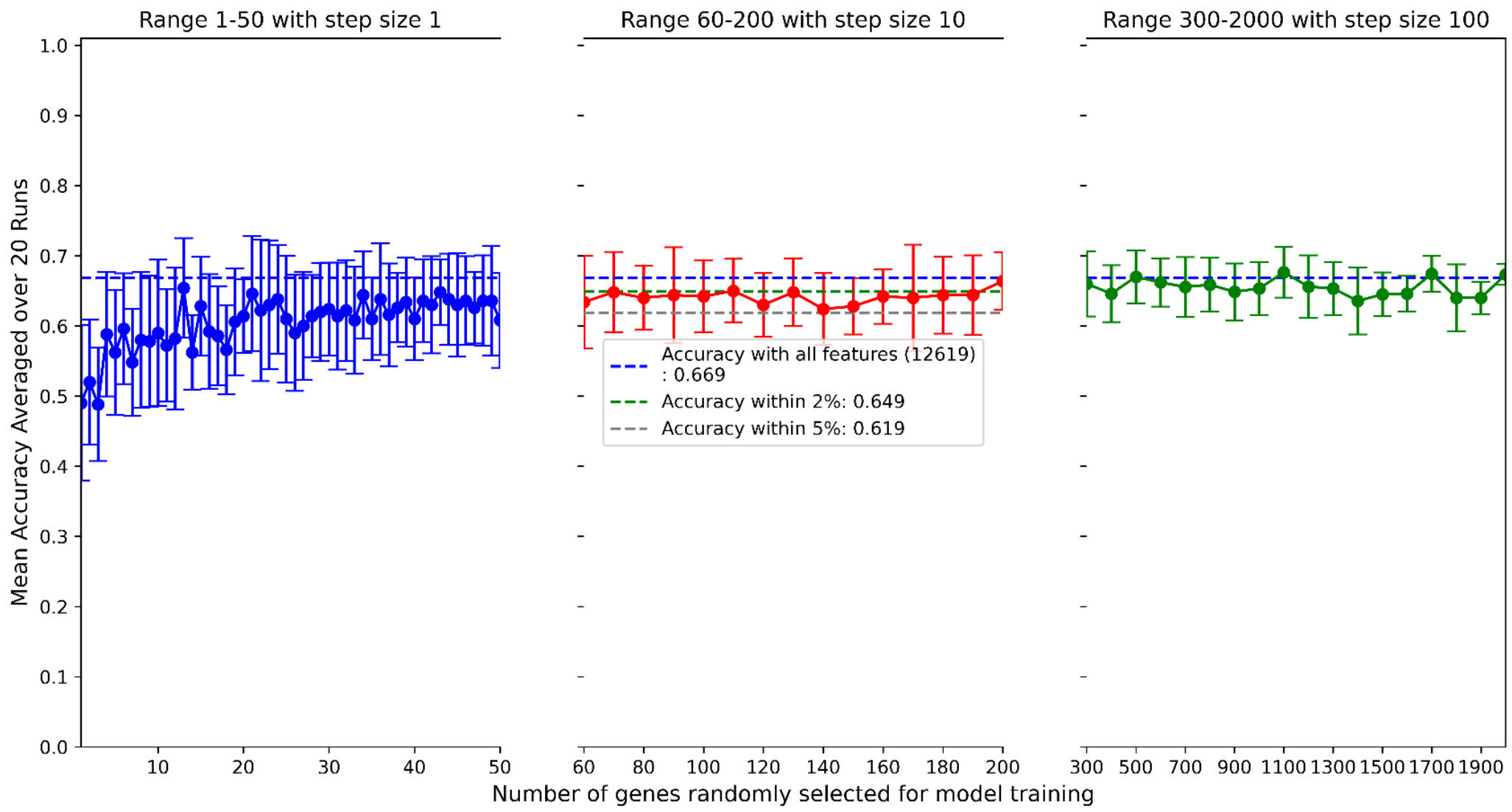

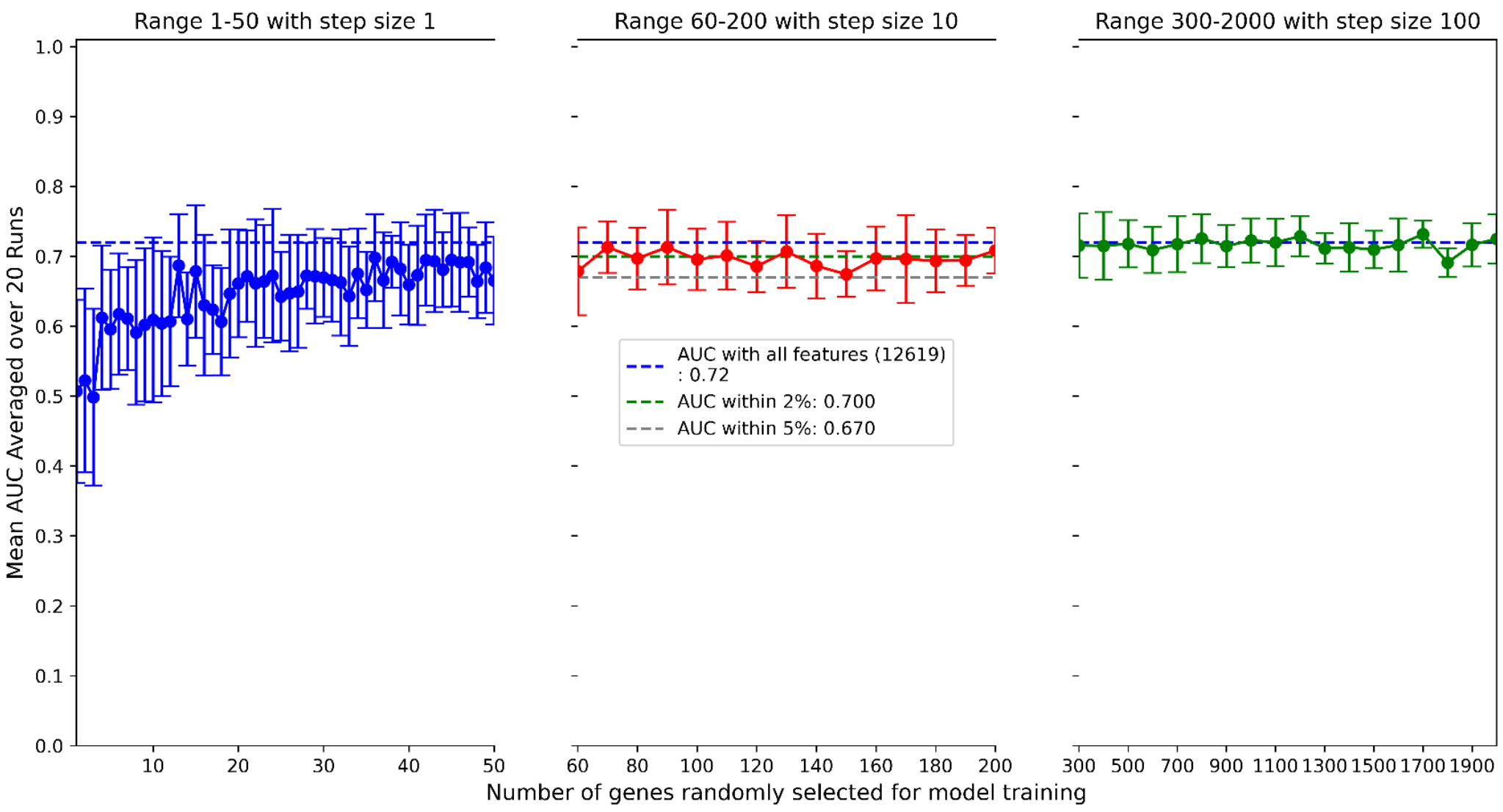

**Random Forest performance with Prostate (GSE6919_U95B) dataset** (mean and standard deviation are reported over 20 runs). Models trained and tested on 80:20 split shows that a random subset of size ~50 (0.4% of all features) is able to match within-5% Accuracy and AUC of all features.

**Fig. S10.**

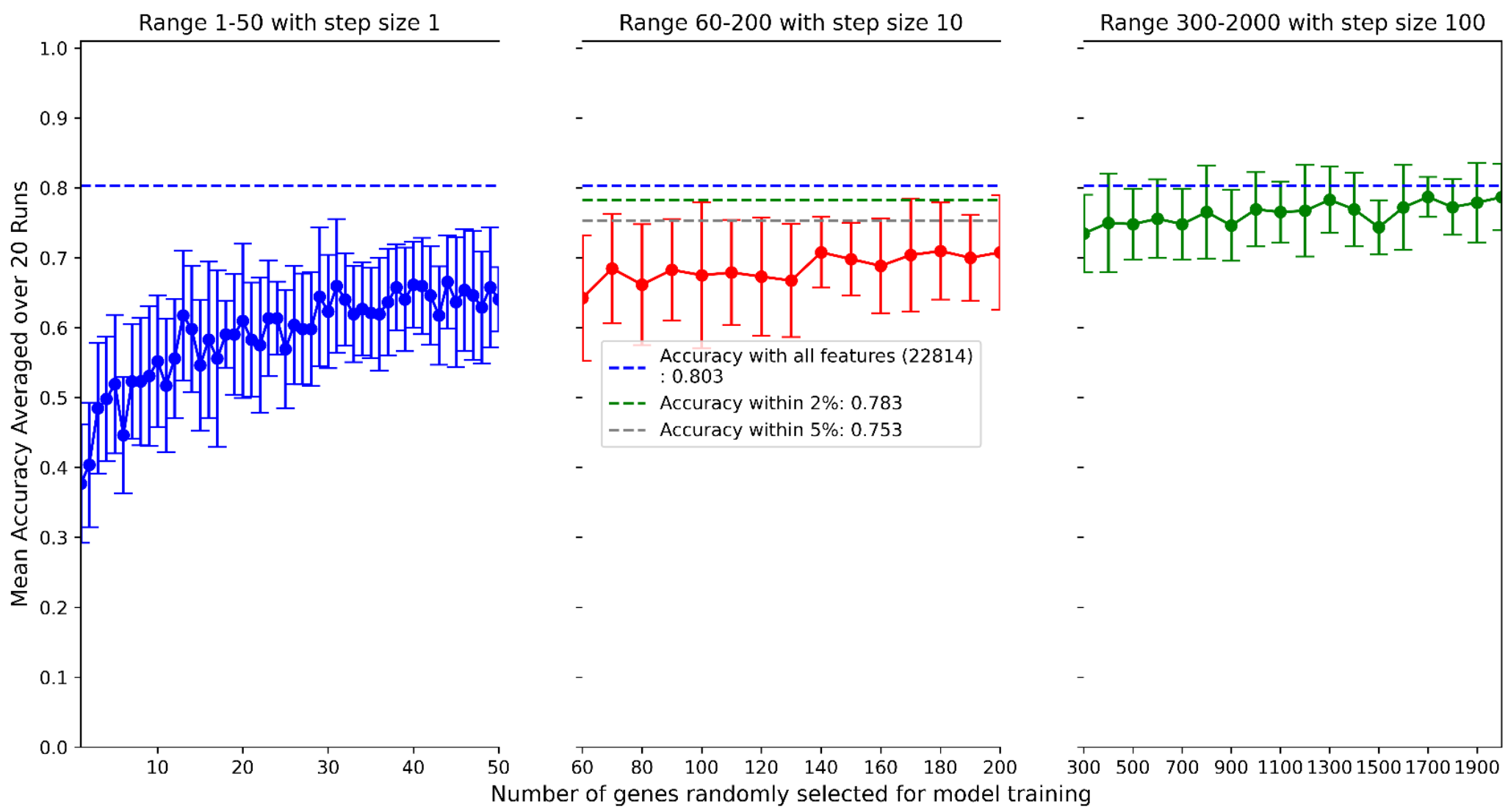

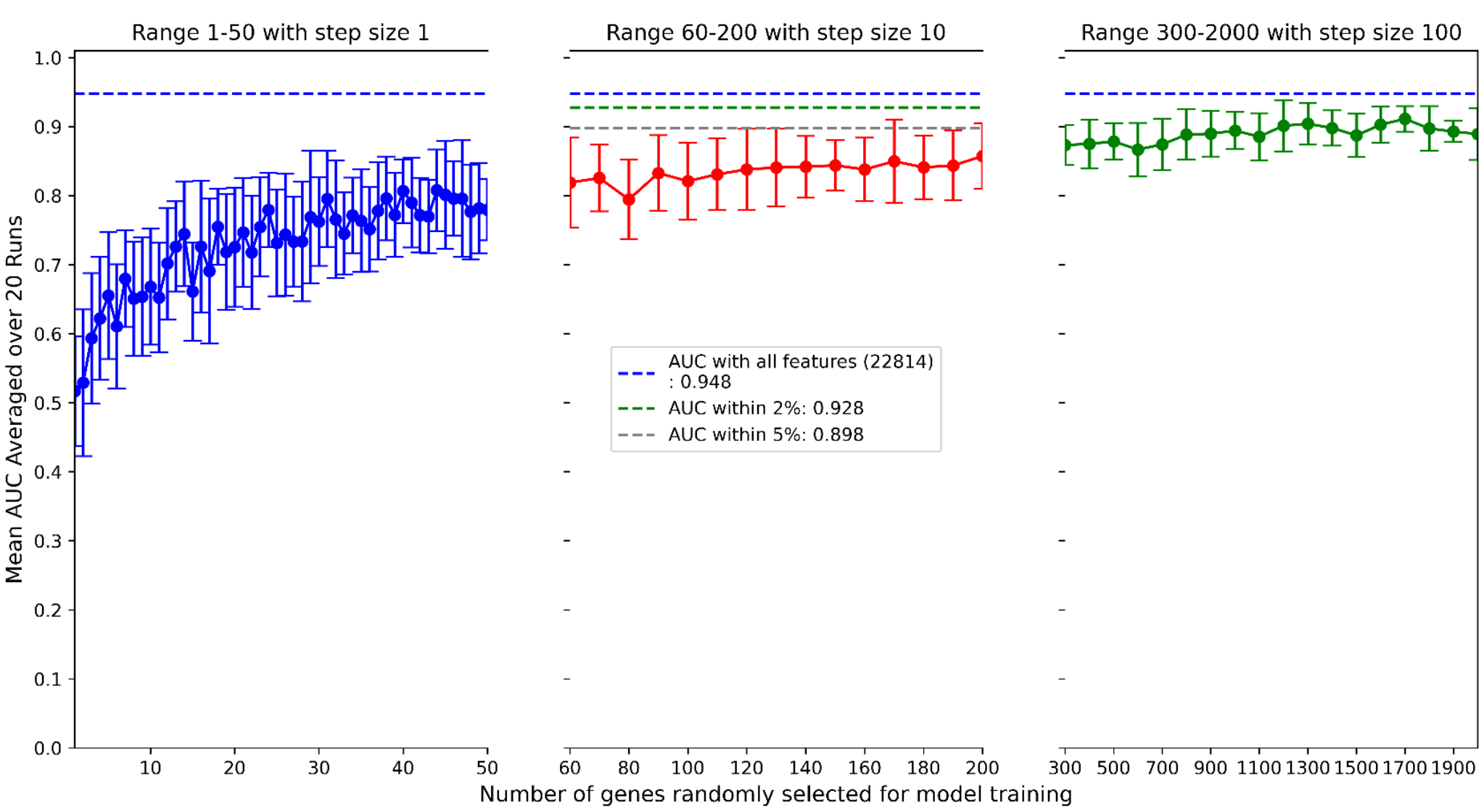

**Random Forest performance with Bowl (GSE3365) dataset** (mean and standard deviation are reported over 20 runs). Models trained and tested on 80:20 split shows that a random subset of size ~500 (~2.2% of all features) is able to match within-5% Accuracy and AUC of all features.

**Fig. S11.**

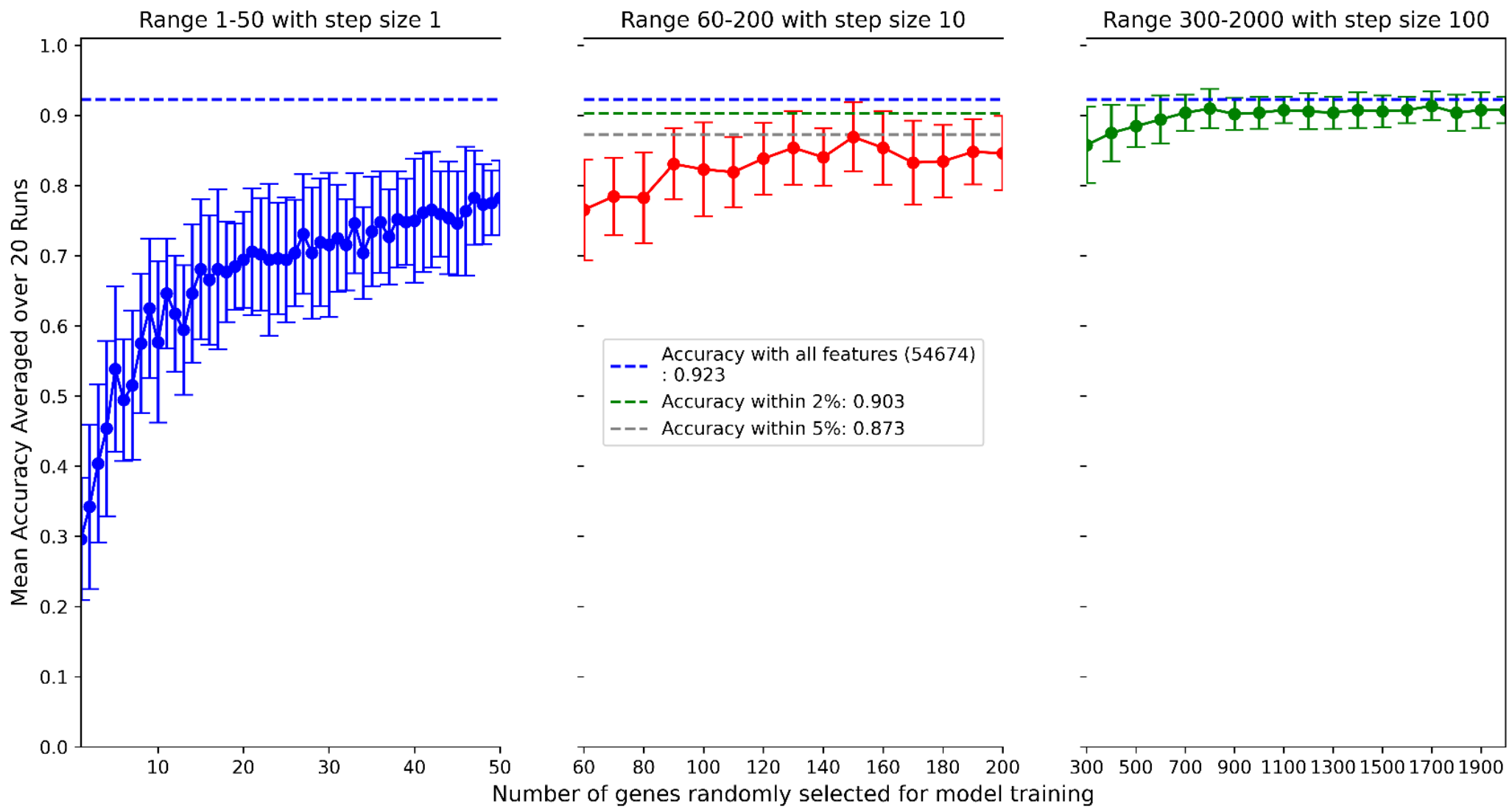

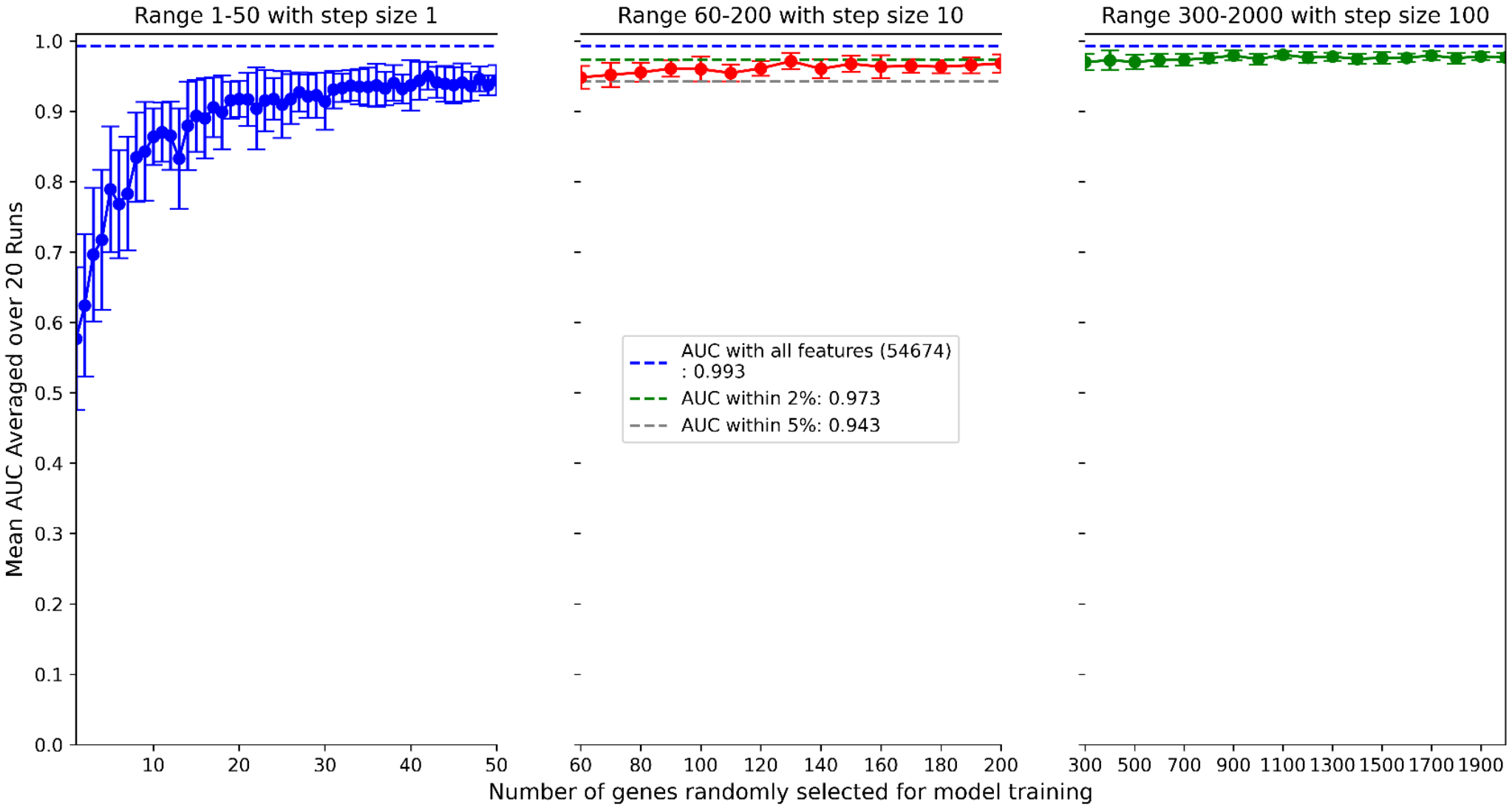

**Random Forest performance with Brain (GSE50161) dataset** (mean and standard deviation are reported over 20 runs). Models trained and tested on 80:20 split shows that a random subset of size ~50 (~0.09% of all features) is able to match within-5% Accuracy and AUC of all features.

**Fig. S12.**

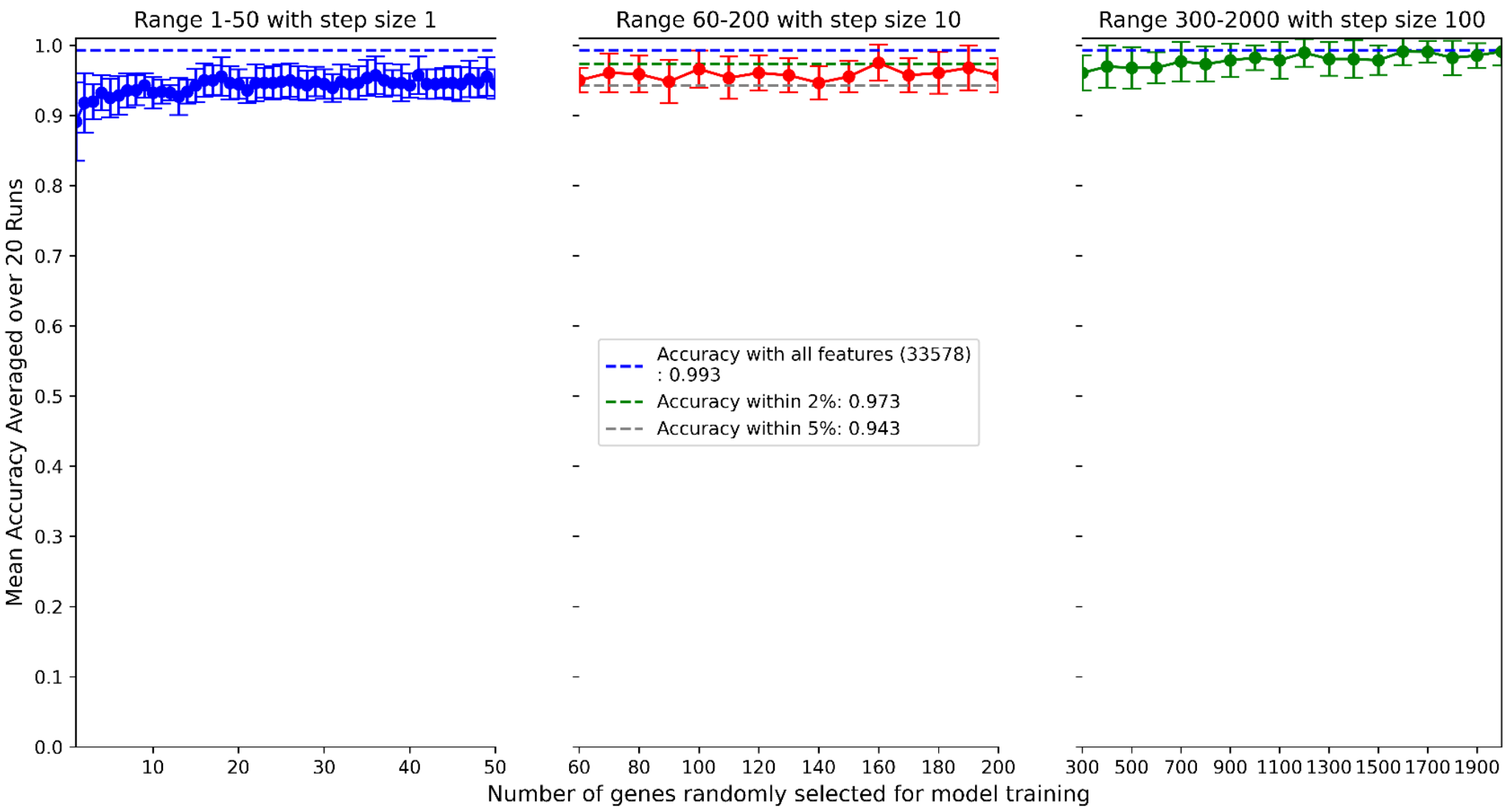

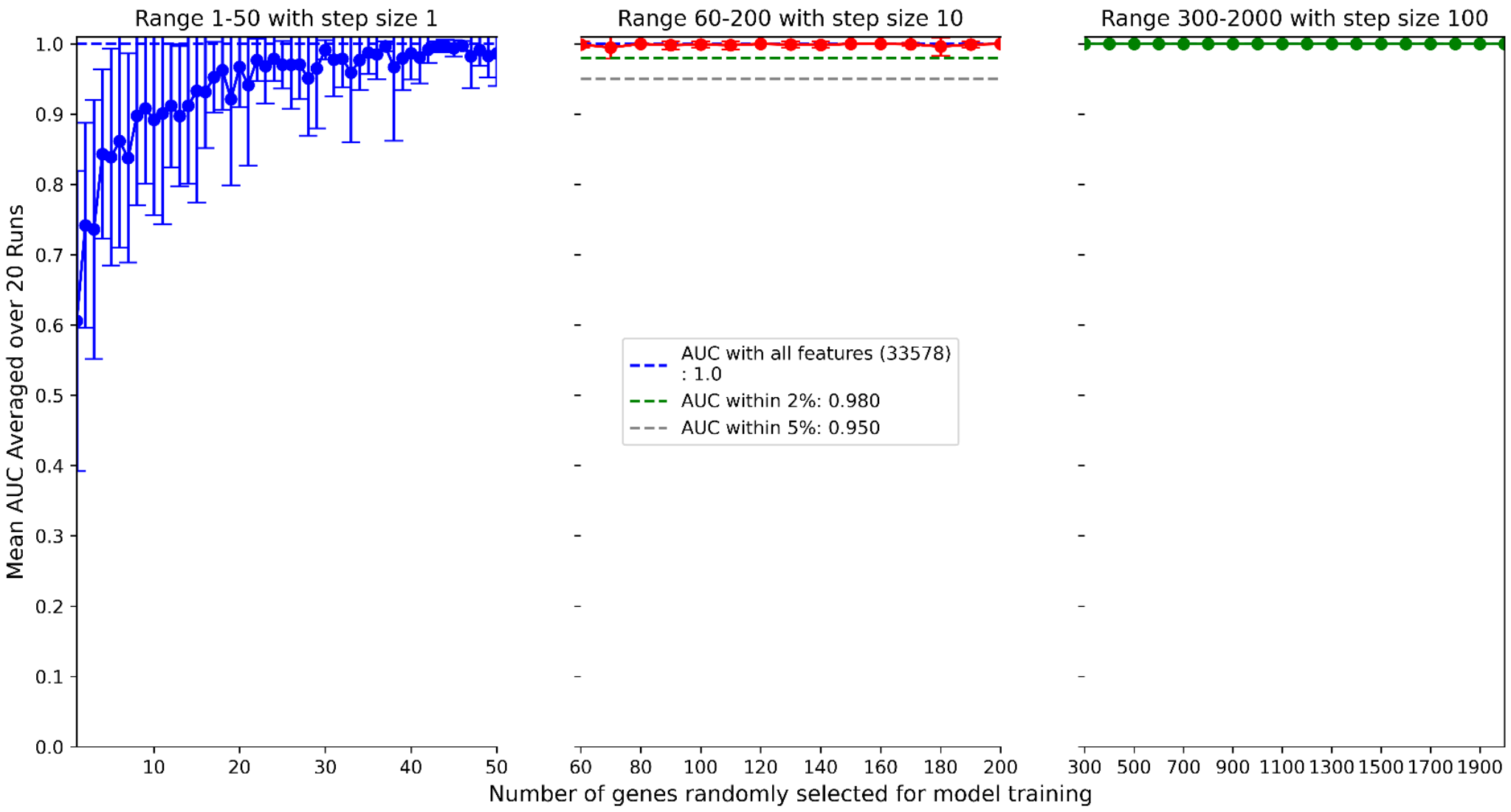

**Random Forest performance with Breast (GSE22820) dataset** (mean and standard deviation are reported over 20 runs). Models trained and tested on 80:20 split shows that a random subset of size ~50 (~0.14% of all features) is able to match within-5% Accuracy and full AUC of all features. (The unusually high accuracy with just one feature is because there is a severe class imbalance in this dataset).

## Fig. S13.

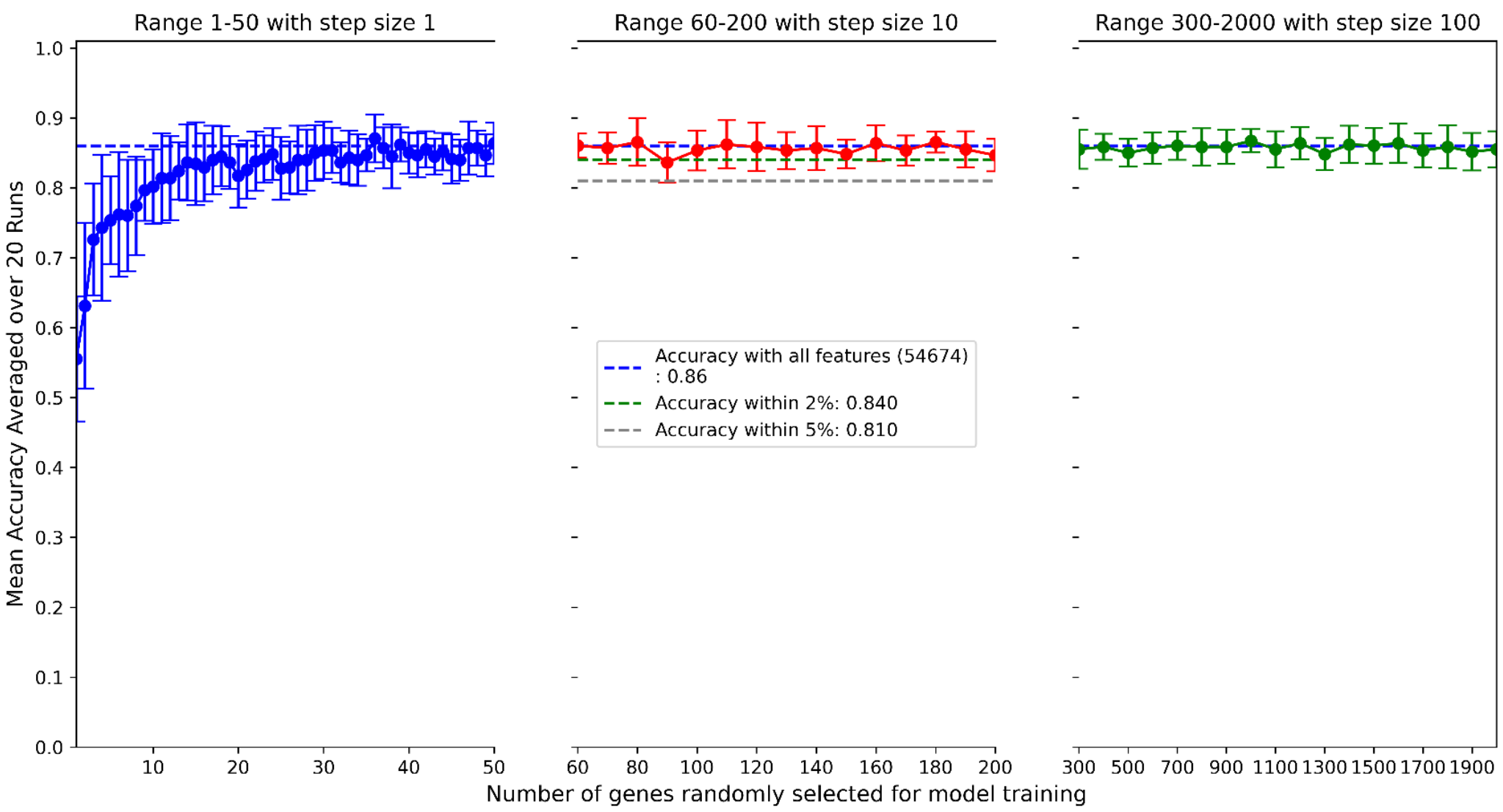

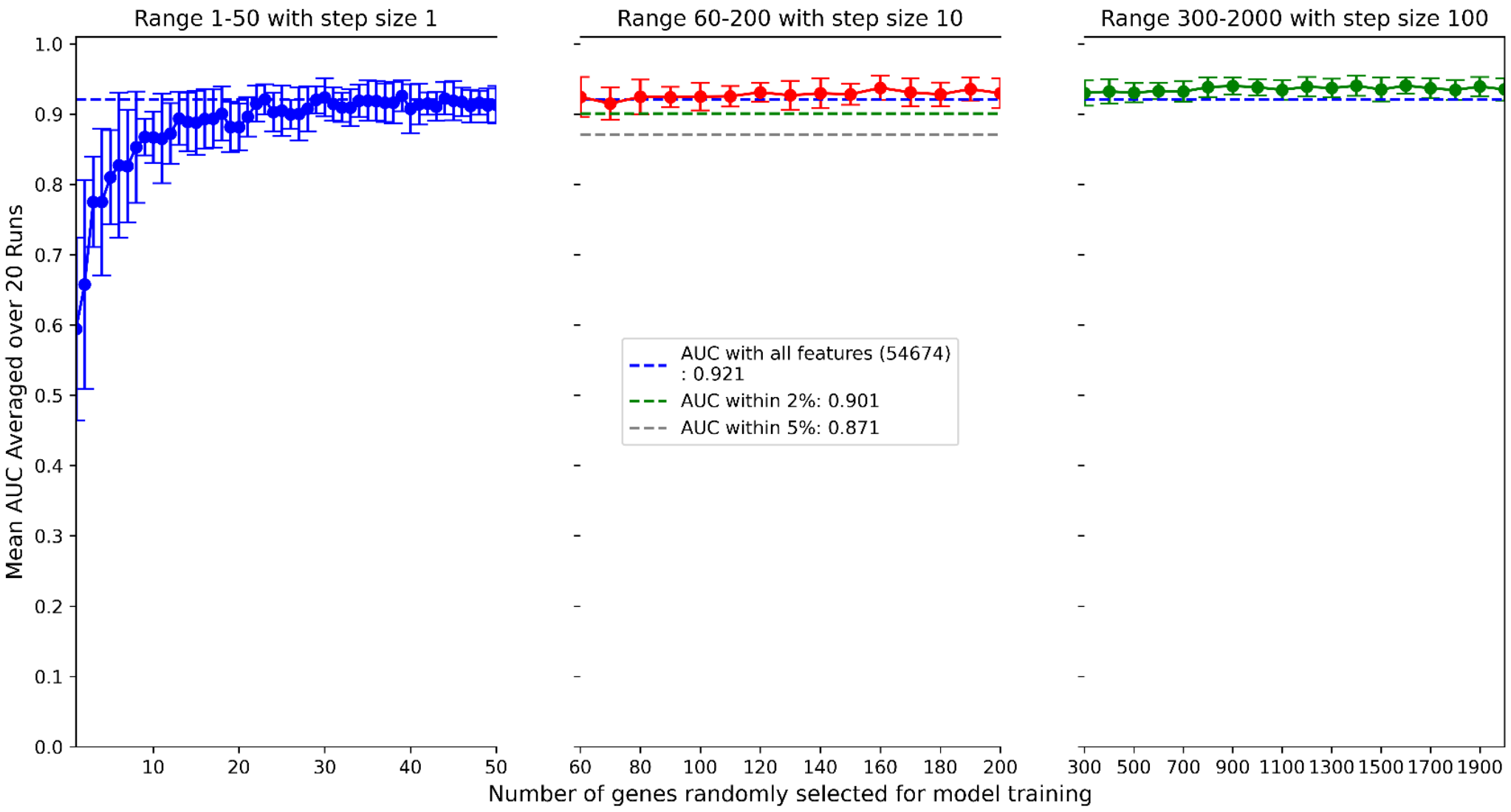

**Random Forest performance with Renal (GSE53757) dataset** (mean and standard deviation are reported over 20 runs). Models trained and tested on 80:20 split shows that a random subset of size ~30 (~0.06% of all features) is able to match full Accuracy and full AUC of all features.

## Fig. S14.

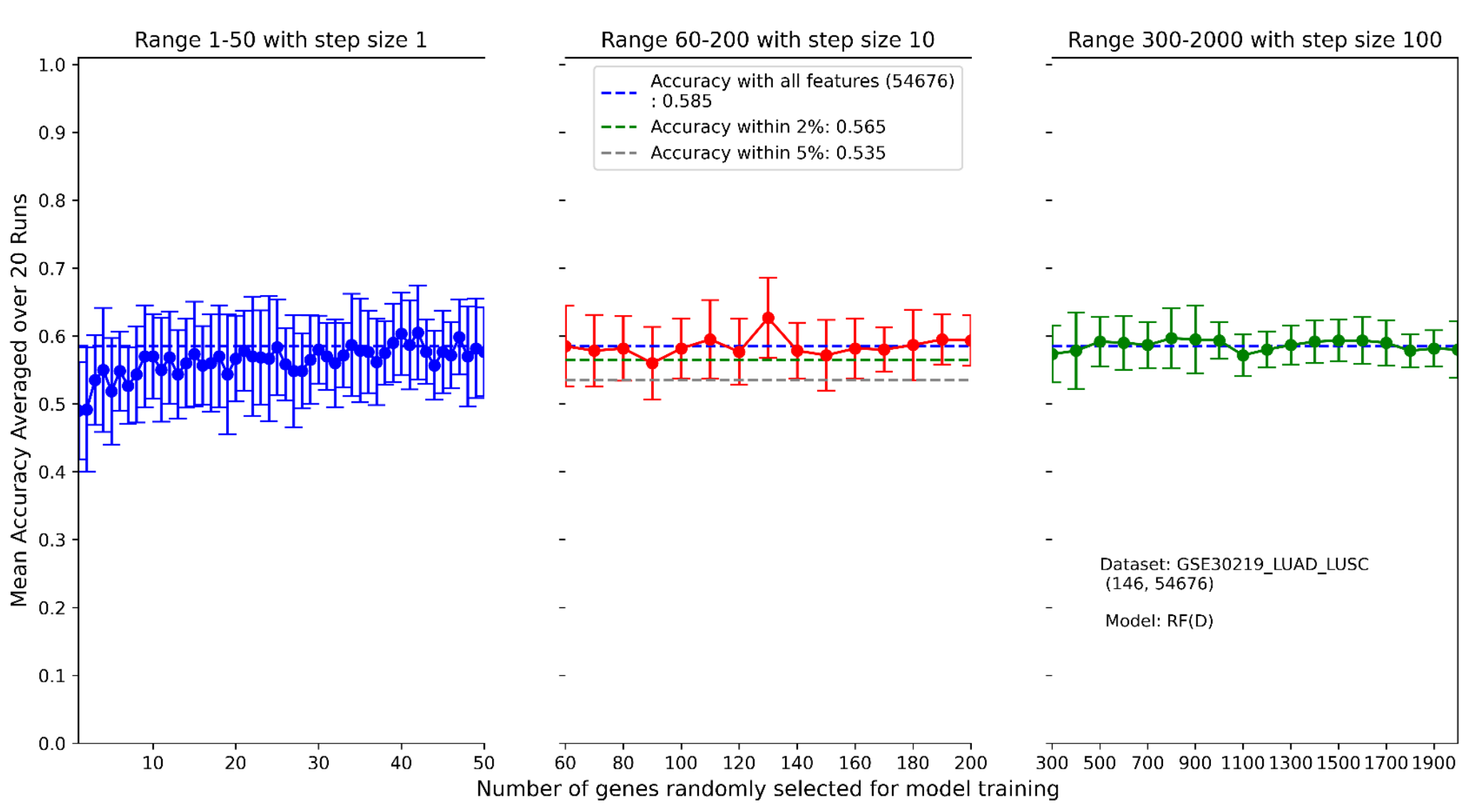

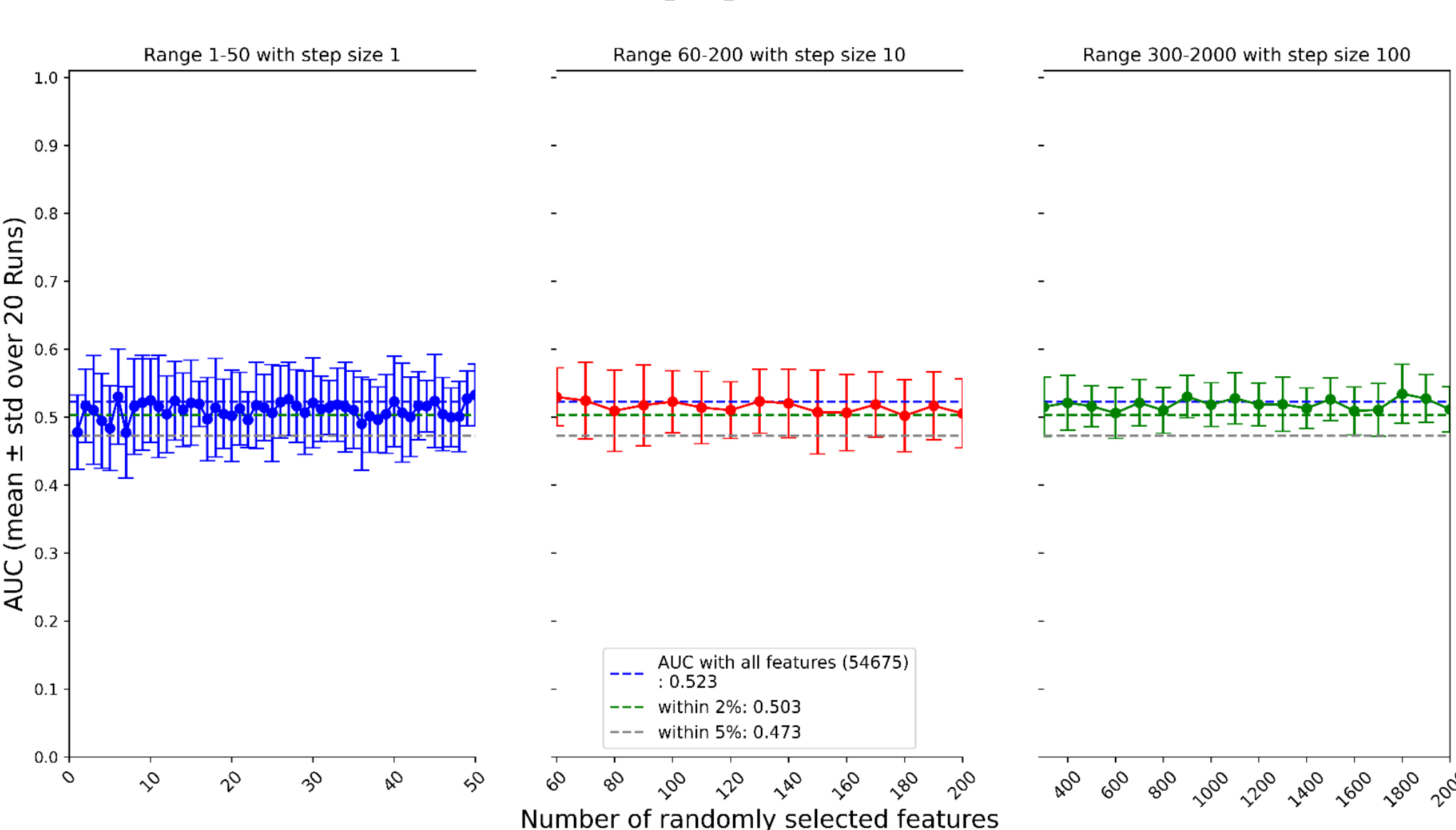

**Random Forest performance with Lung Cancer (GSE30219) dataset** (mean and standard deviation are reported over 20 runs). Models trained and tested on 80:20 split shows that a random subset is able to match full Accuracy and full AUC of all features.

**Fig. S15.**

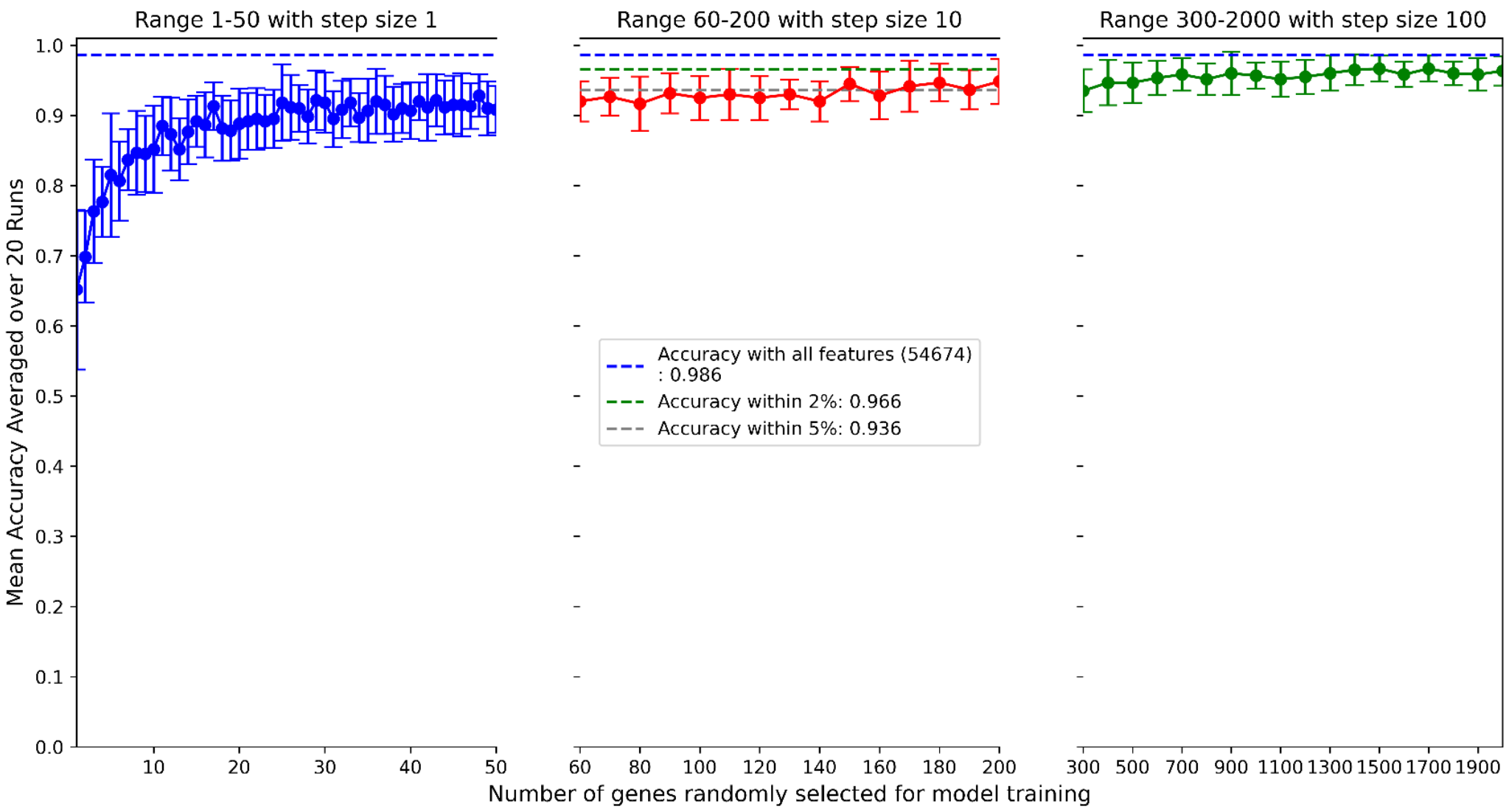

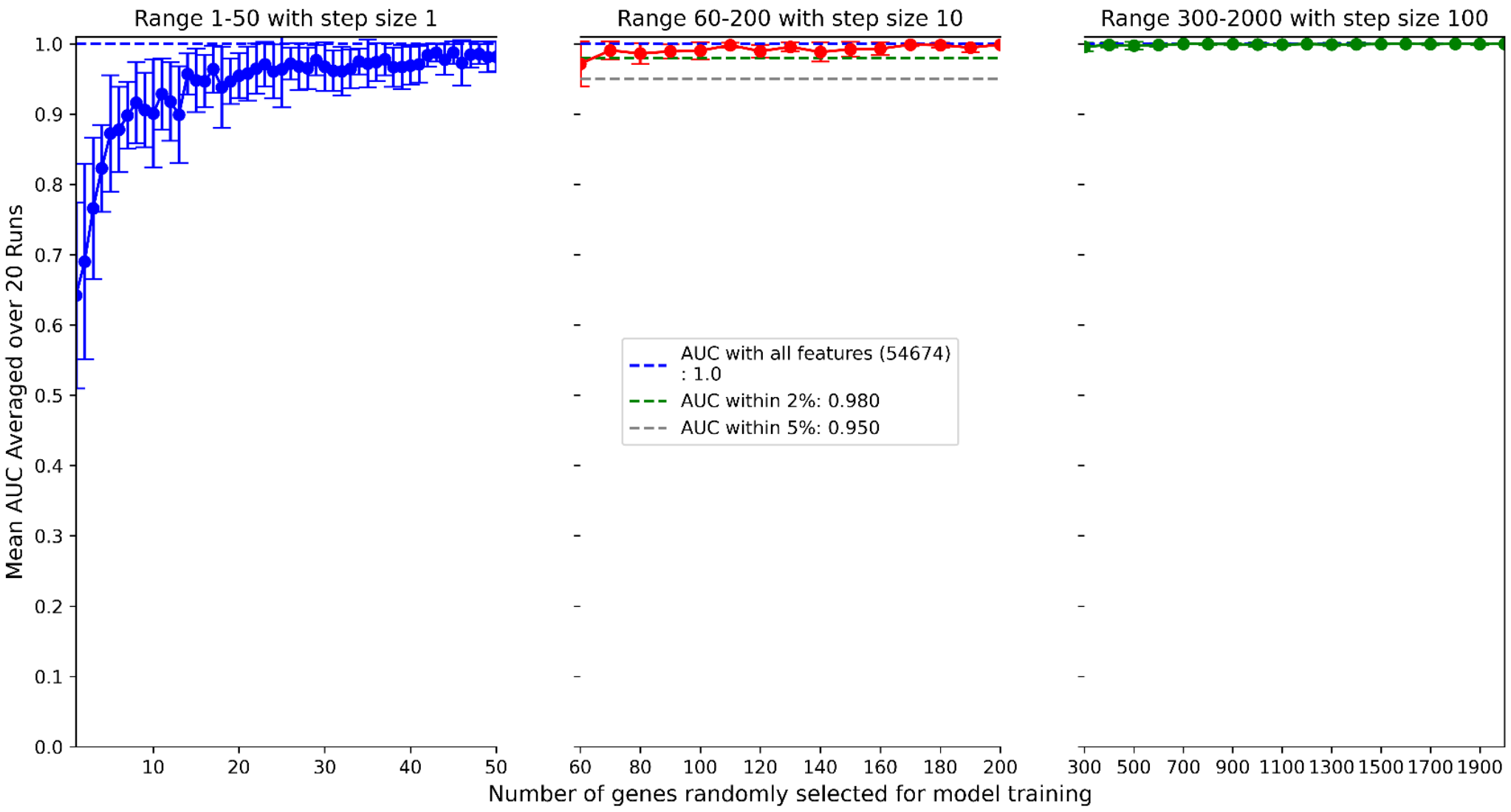

**Random Forest performance with Lung Cancer (GSE30219) dataset** (mean and standard deviation are reported over 20 runs). Models trained and tested on 80:20 split shows that a random subset is able to match within-2% Accuracy and full AUC of all features.

**Fig. S16.**

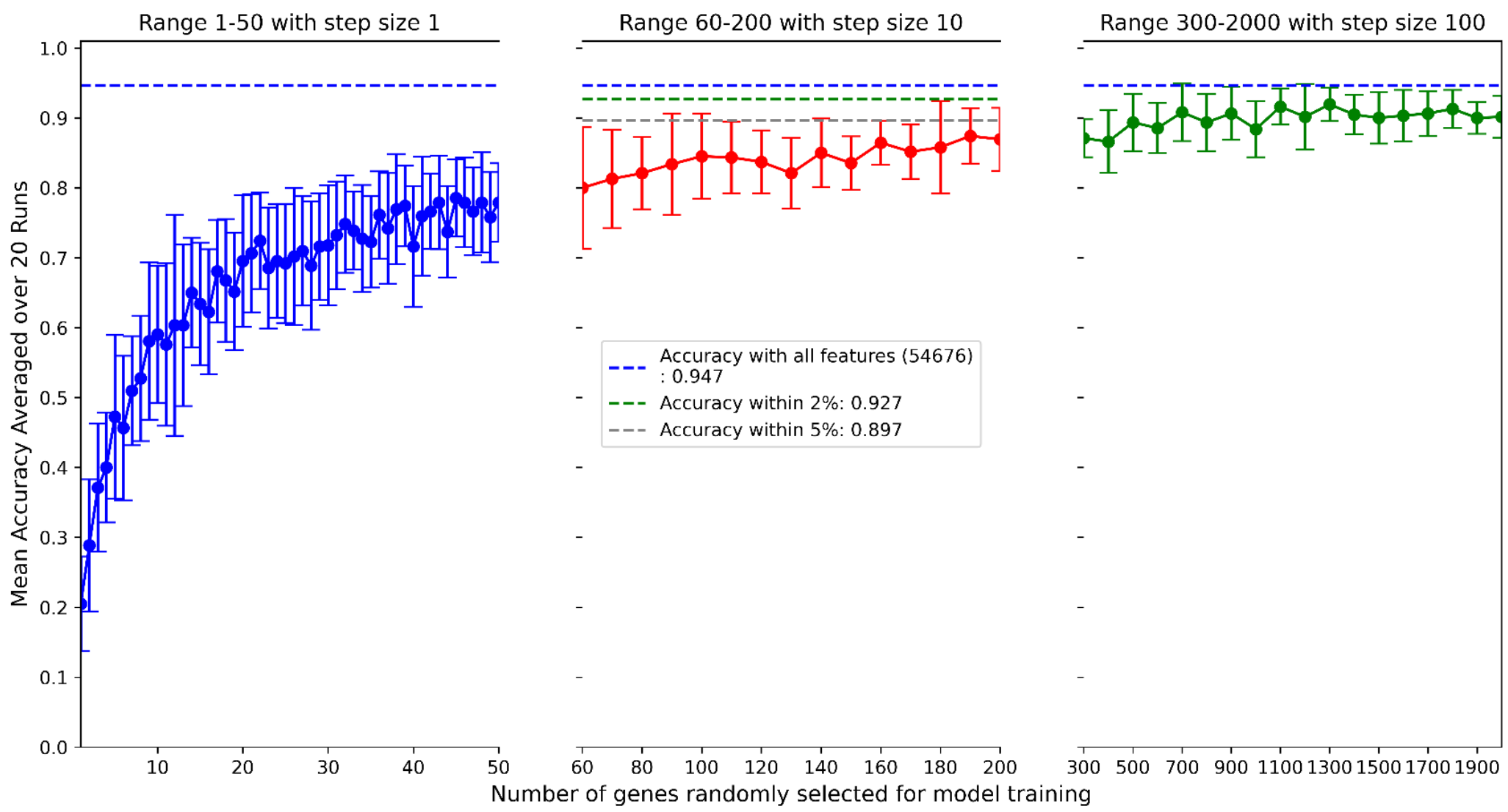

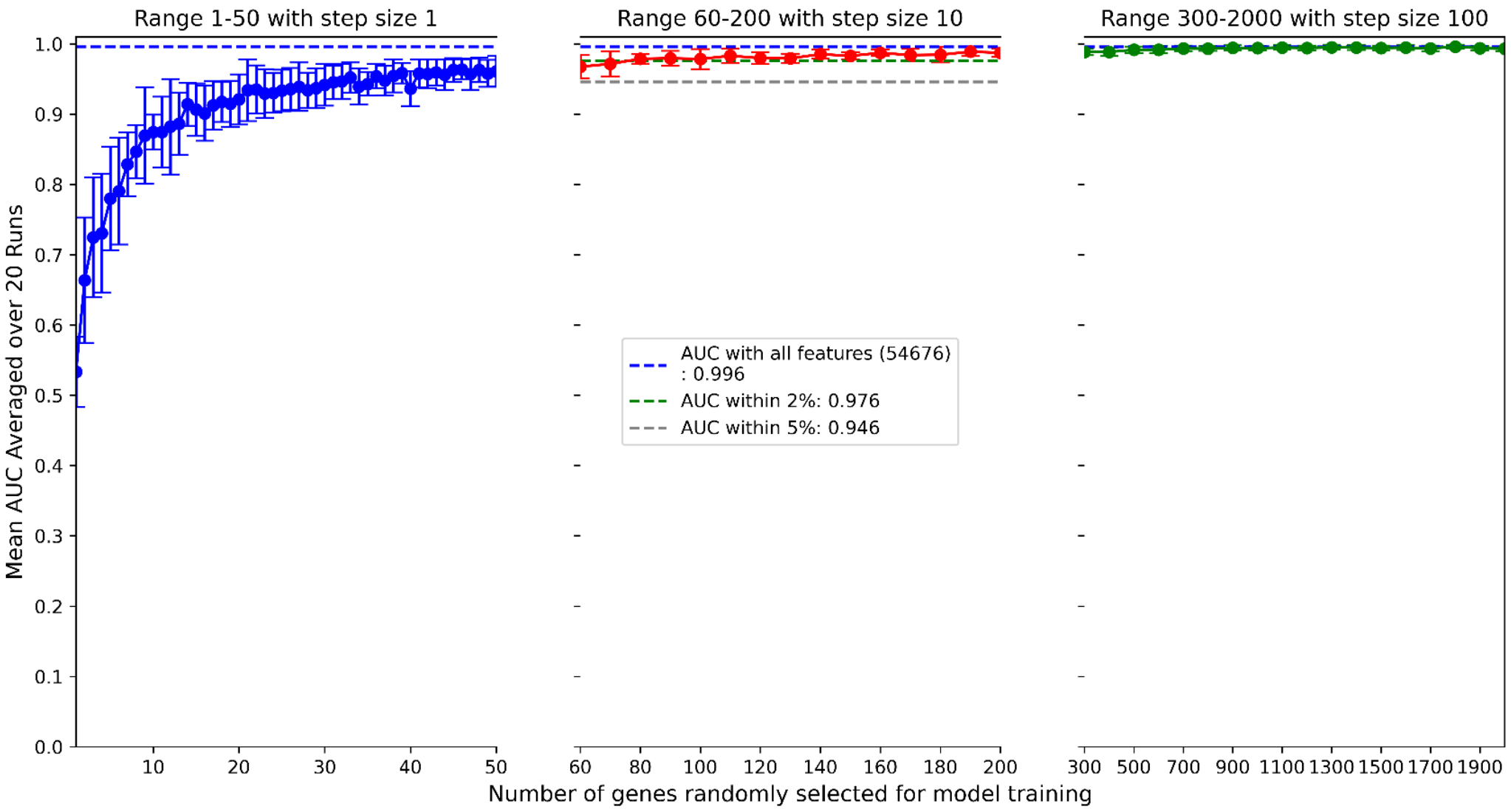

**Random Forest performance with Breast Cancer (GSE45827) dataset** (mean and standard deviation are reported over 20 runs). Models trained and tested on 80:20 split shows that a random subset is able to match within-5% Accuracy and full AUC of all features.

**Fig. S17.**

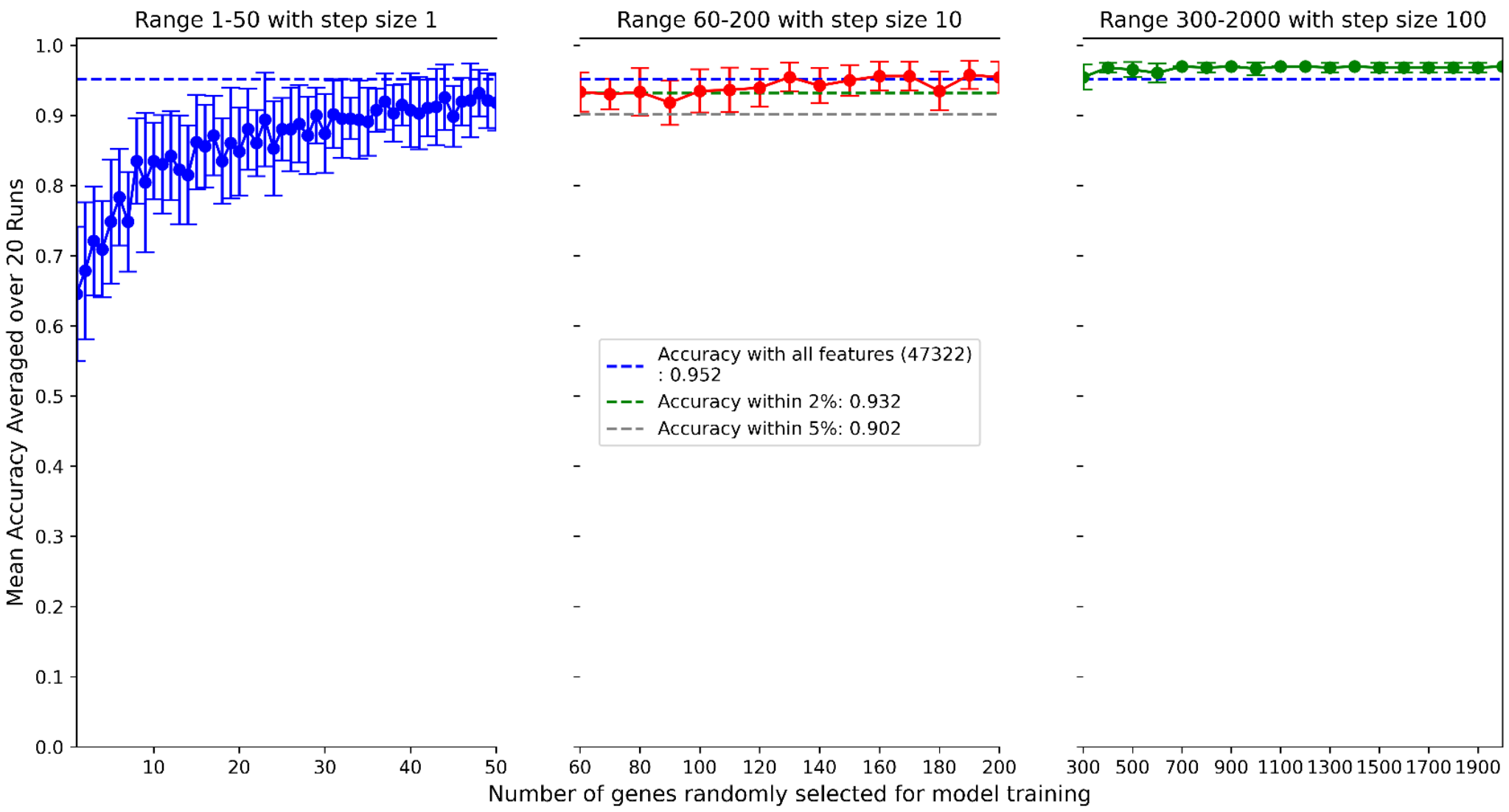

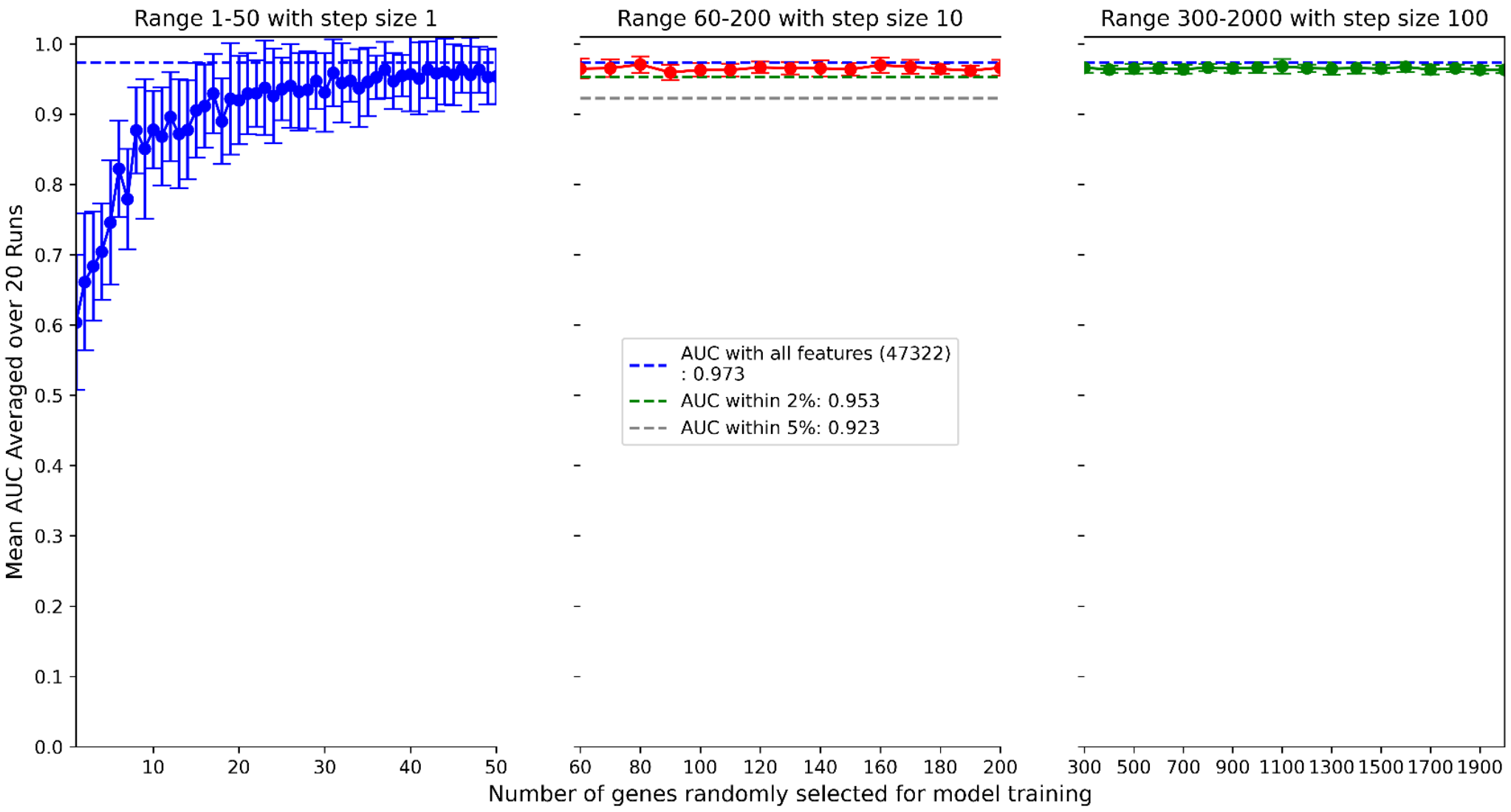

**Random Forest performance with Liver Cancer (GSE76427) dataset** (mean and standard deviation are reported over 20 runs). Models trained and tested on 80:20 split shows that a random subset is able to match full Accuracy and full AUC of all features.

**Fig. S18.**

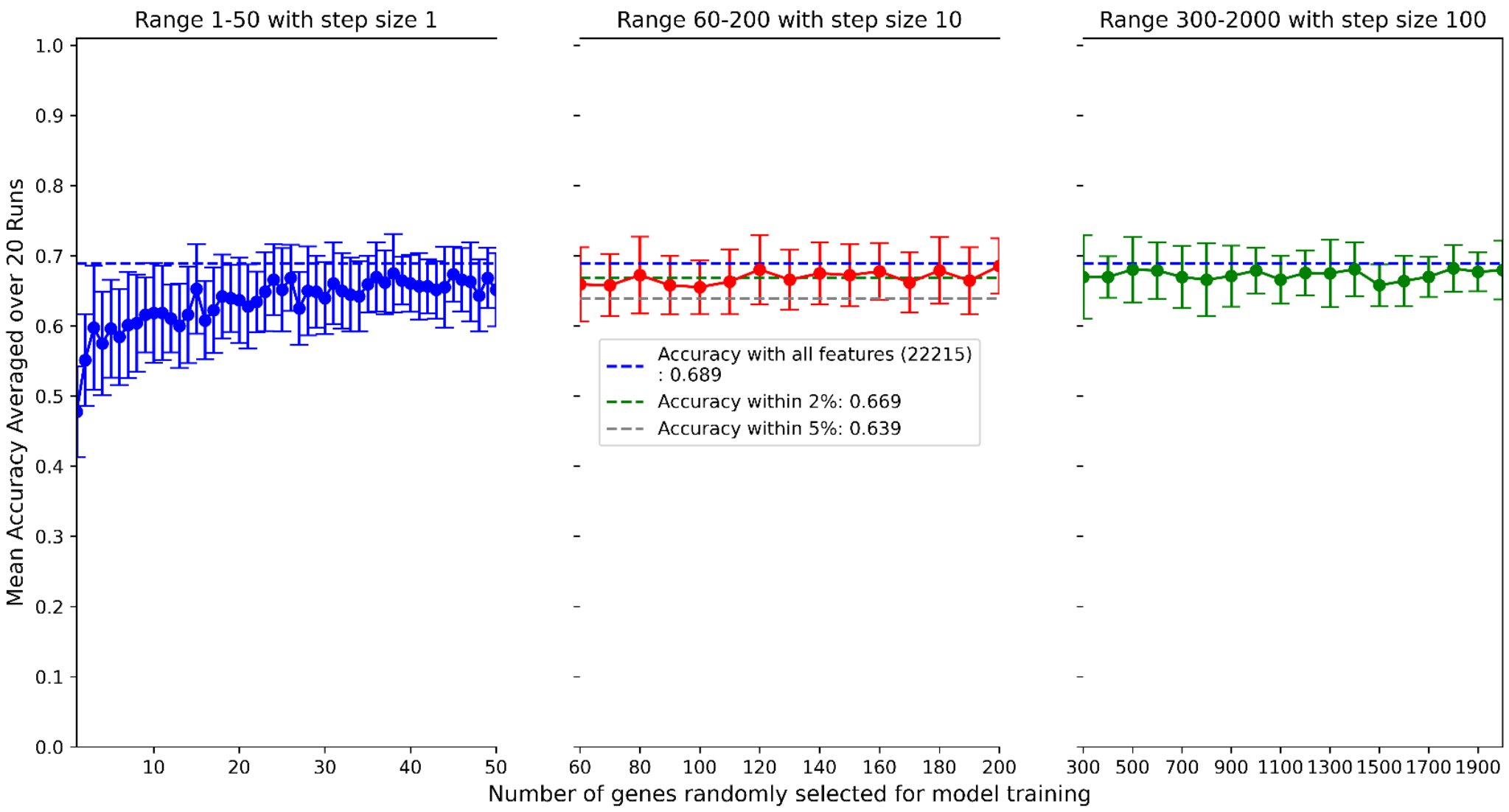

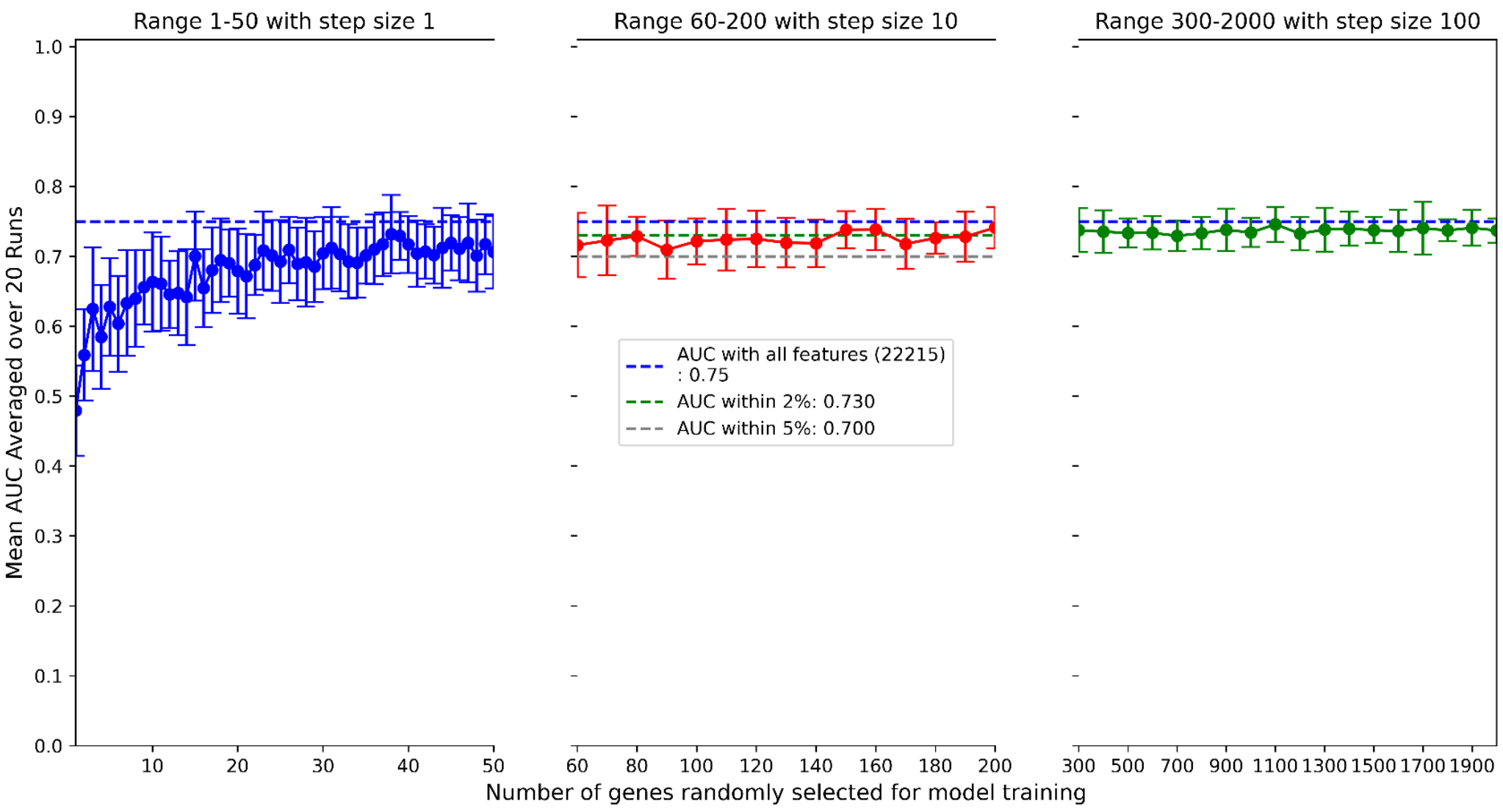

**Random Forest performance with Lung Cancer (GSE4115) dataset** (mean and standard deviation are reported over 20 runs). Models trained and tested on 80:20 split shows that a random subset is able to match within-2% Accuracy and AUC of all features.

**Fig. S19.**

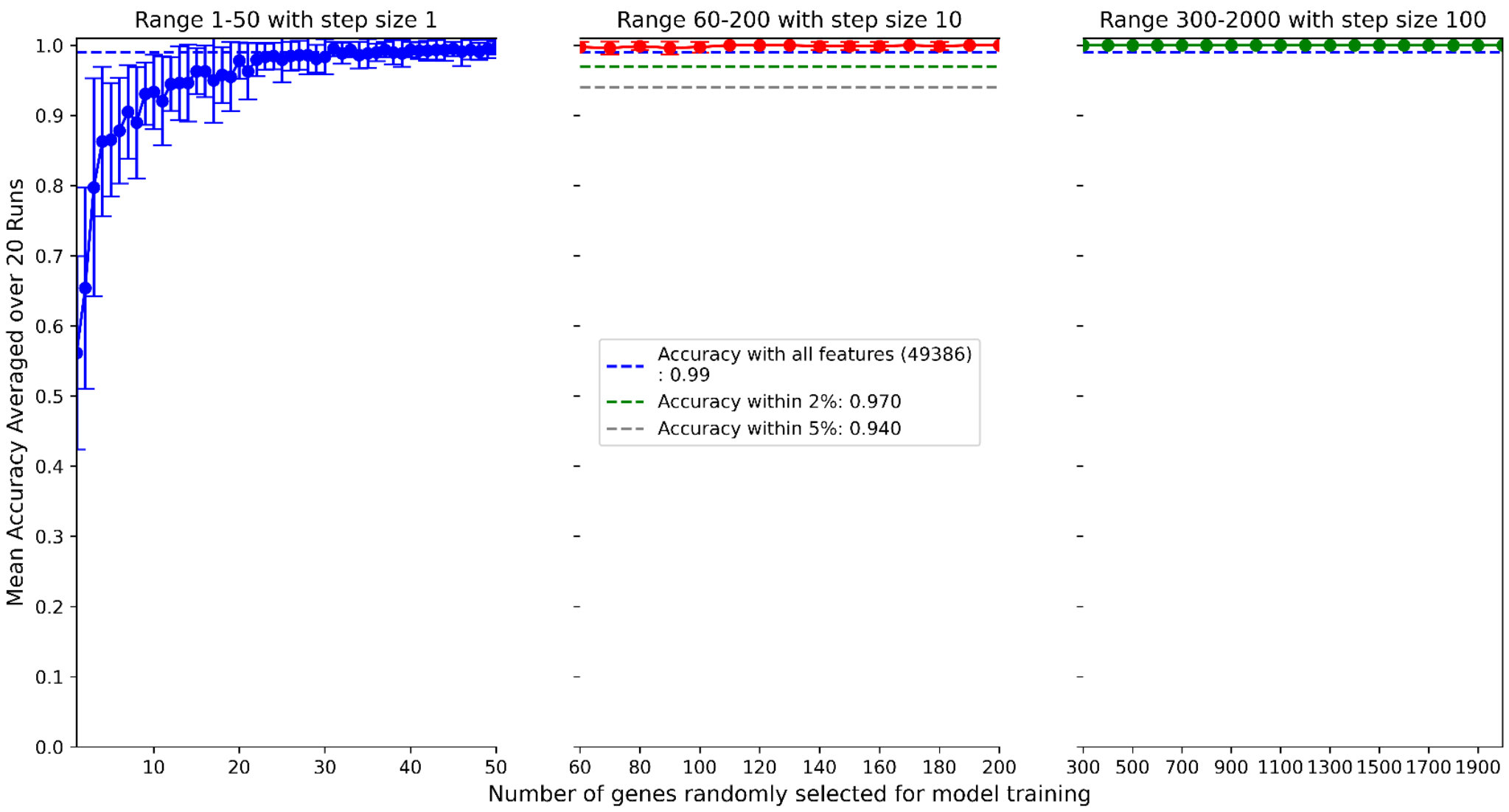

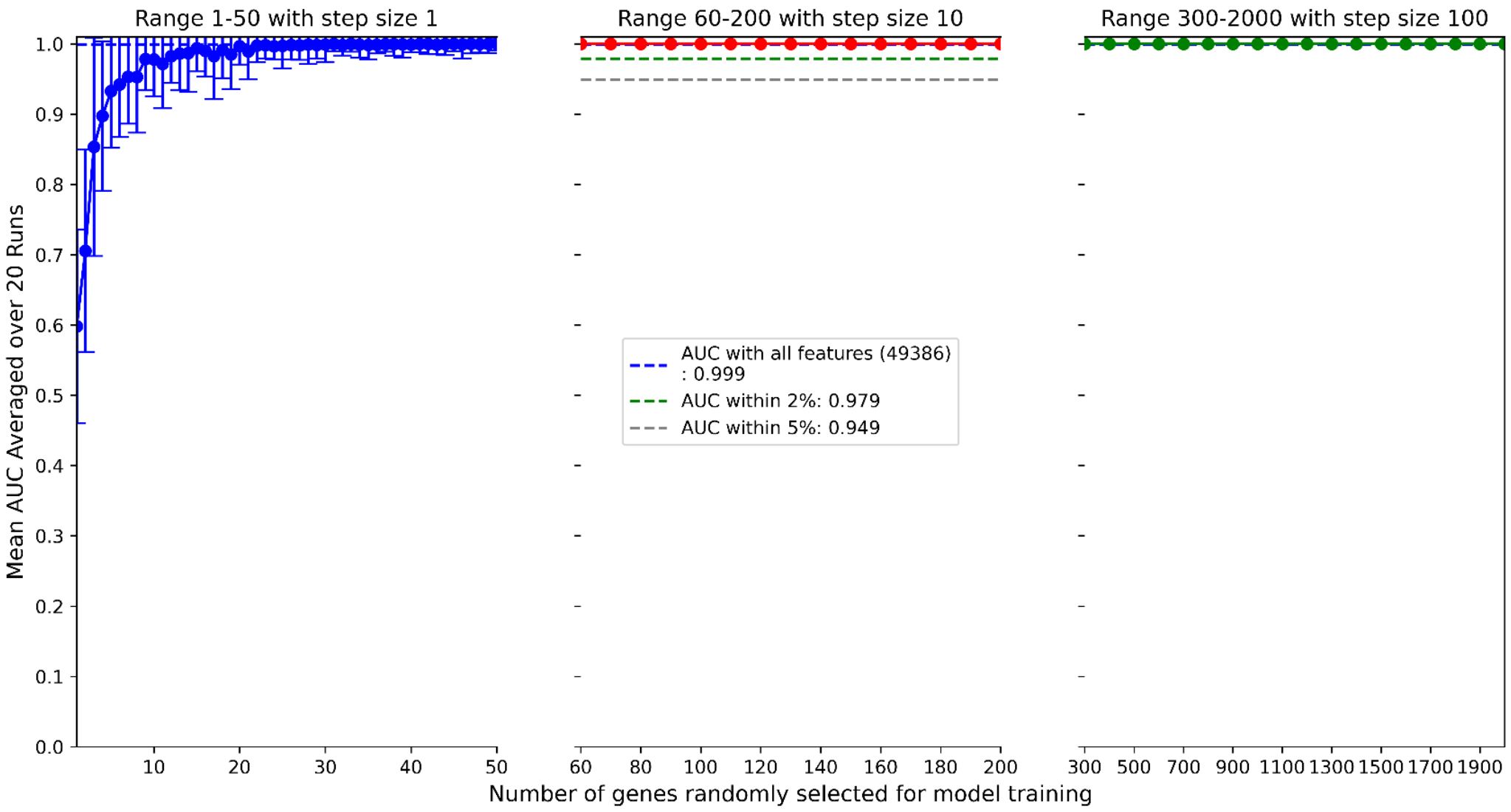

**Random Forest performance with Colorectal Cancer (GSE44076) dataset** (mean and standard deviation are reported over 20 runs). Models trained and tested on 80:20 split shows that a random subset is able to match full Accuracy and full AUC of all features.

**Fig. S20.**

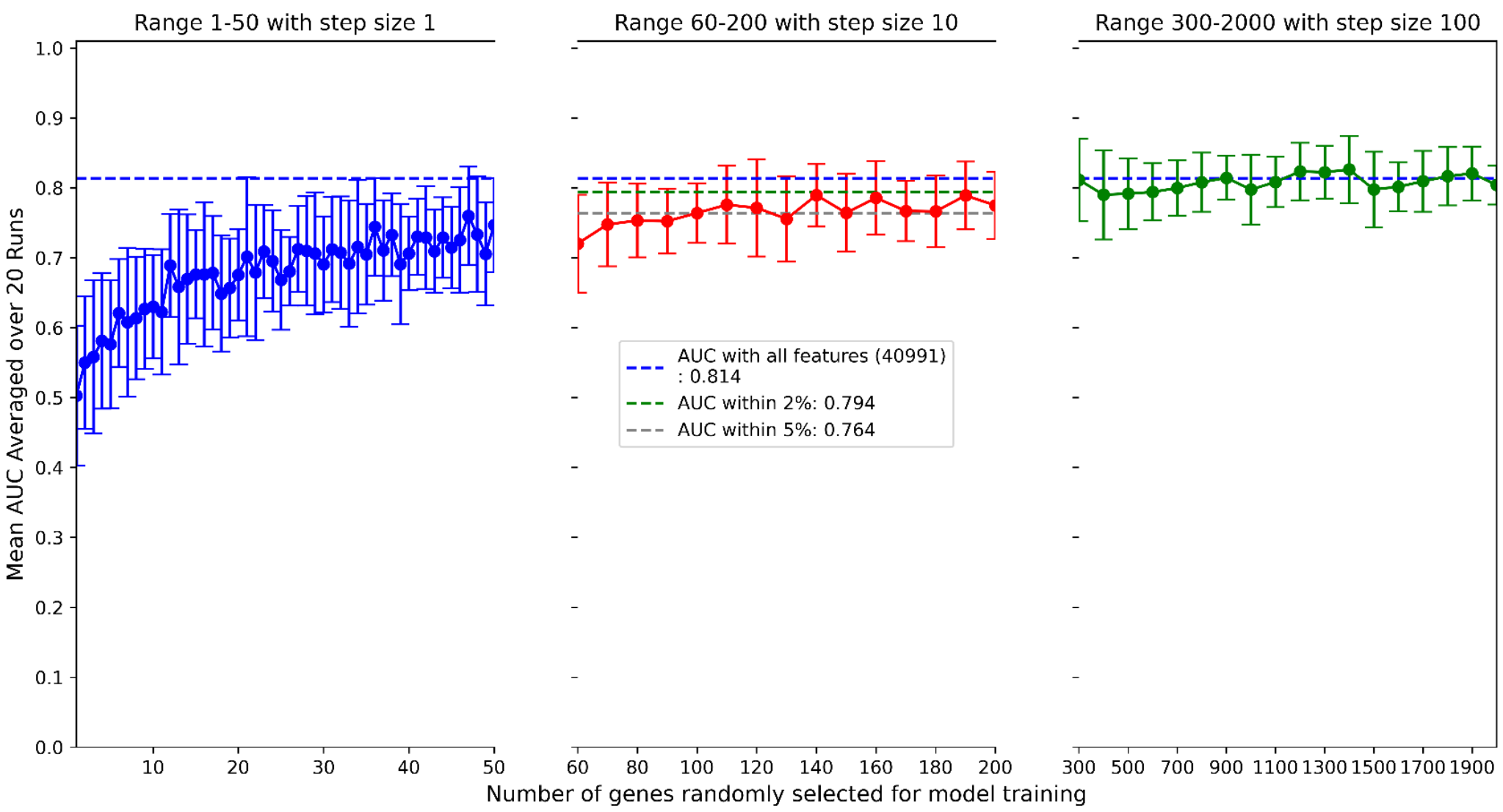

**Random Forest performance with Colon Cancer (GSE11223) dataset** (mean and standard deviation are reported over 20 runs). Models trained and tested on 80:20 split shows that a random subset is able to match full AUC of all features.

**Fig. S21.**

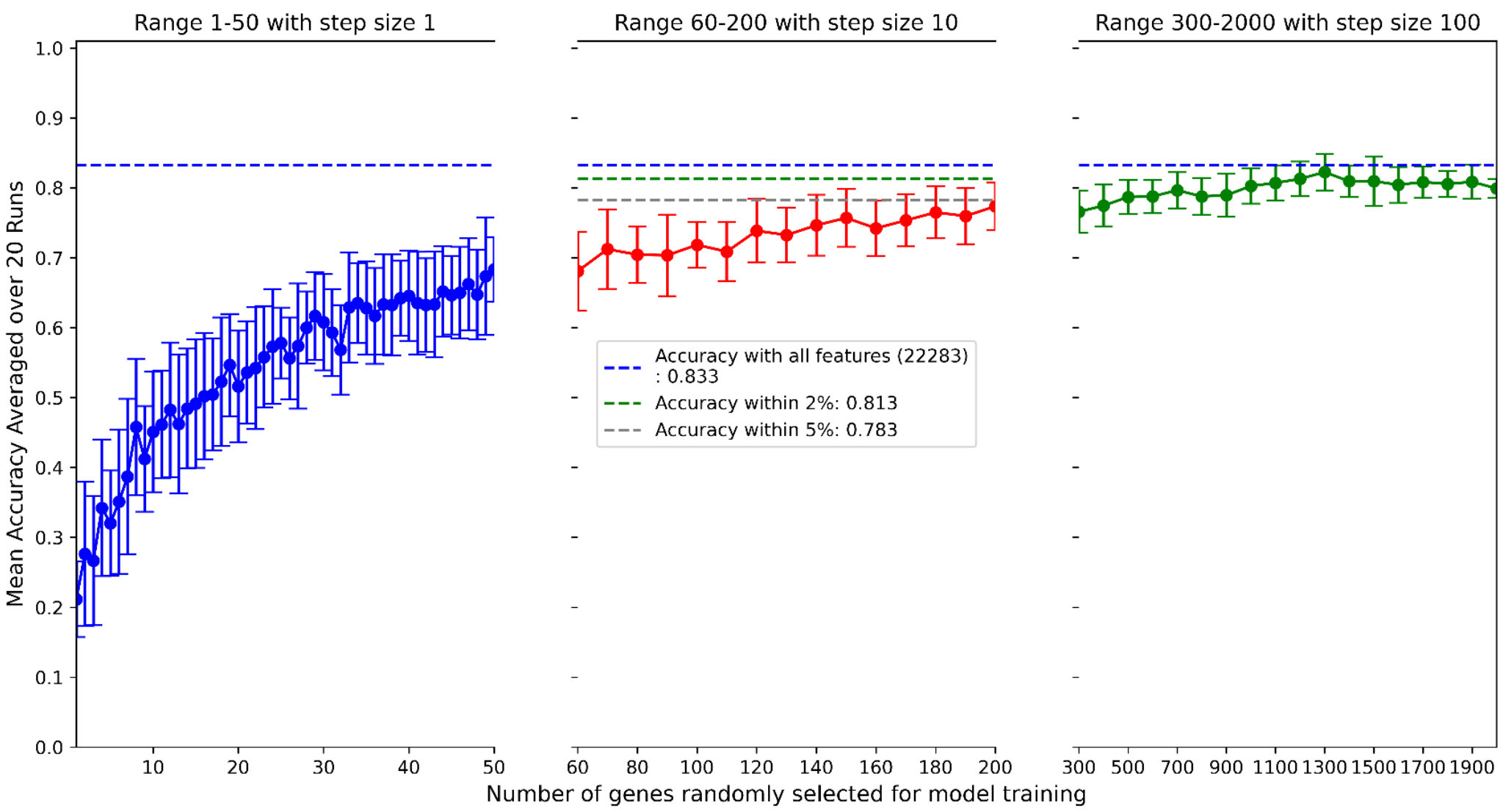

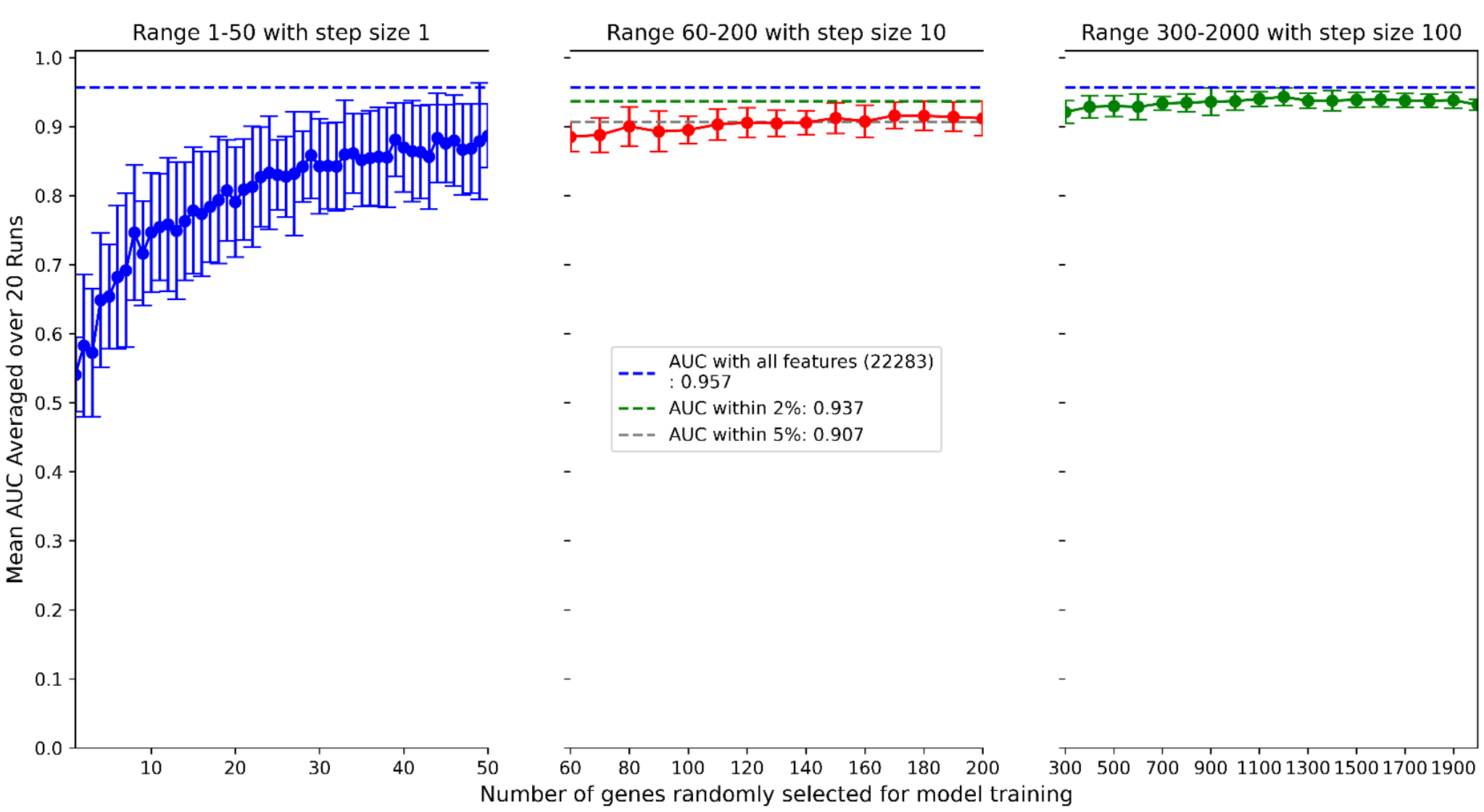

**Random Forest performance with Leumkemia (GSE28497) dataset** (mean and standard deviation are reported over 20 runs). Models trained and tested on 80:20 split shows that a random subset is able to match within-2% Accuracy and AUC of all features.

## Fig. S22.

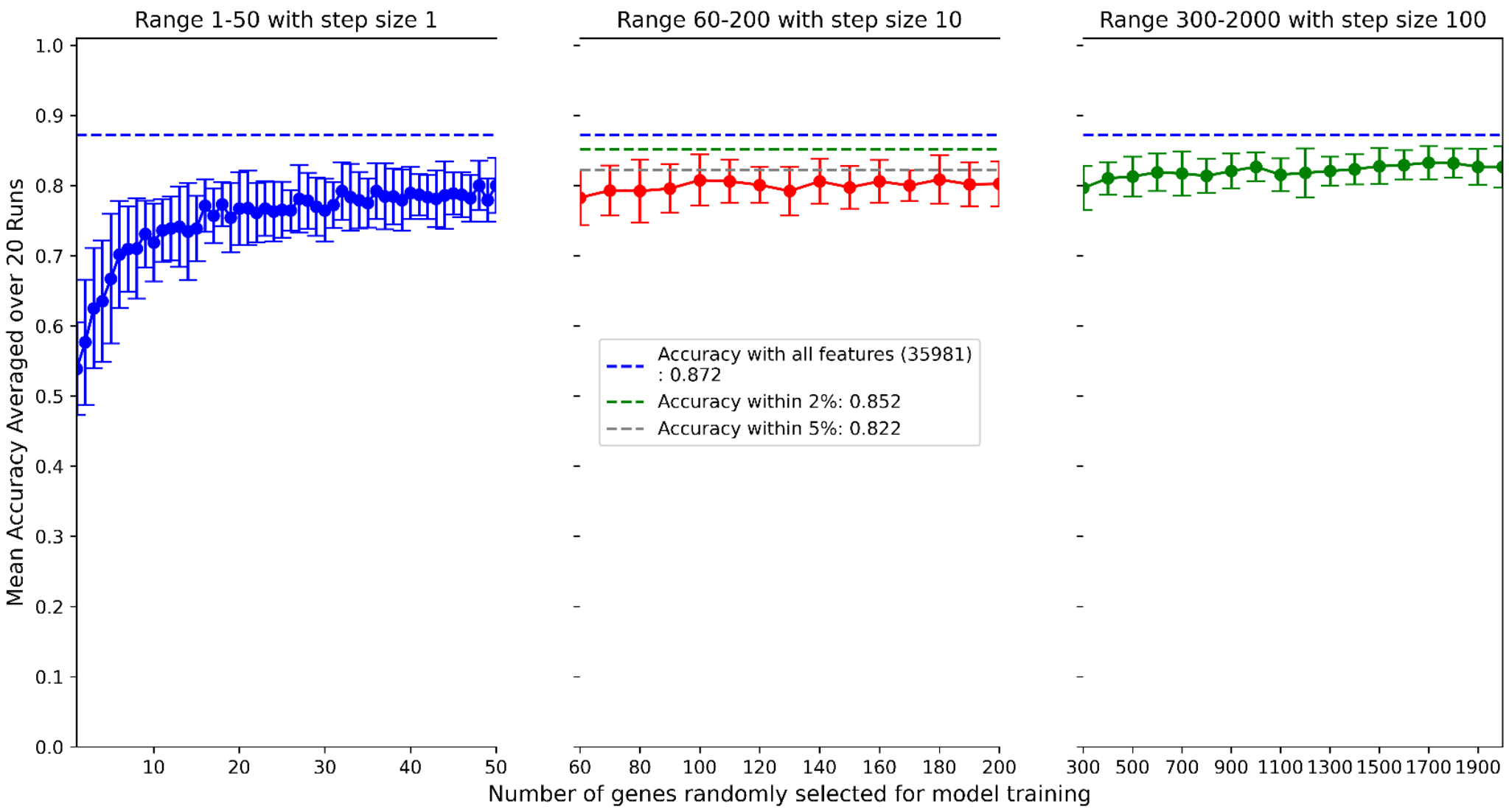

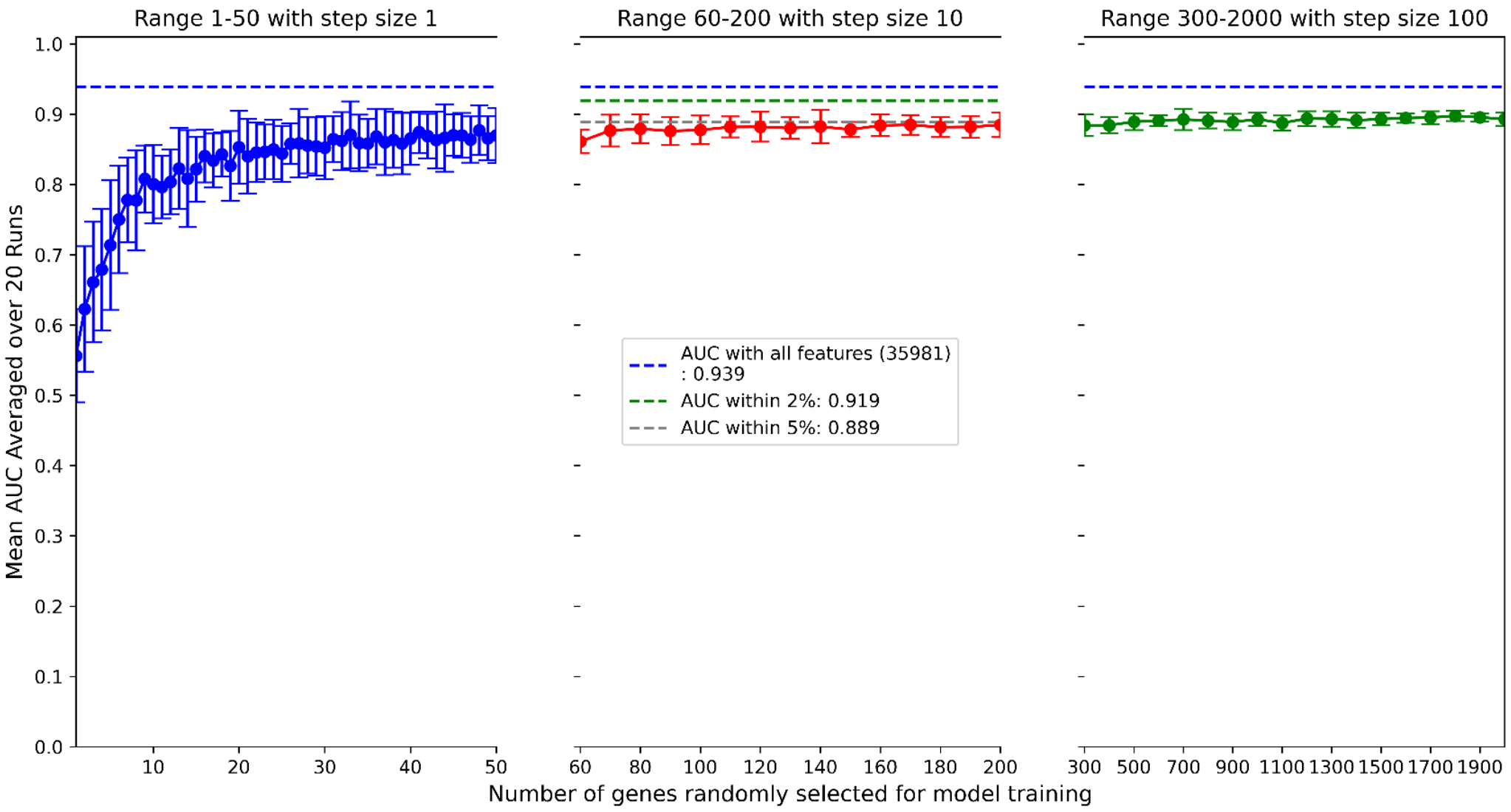

**Random Forest performance with Breast Cancer (GSE70947) dataset** (mean and standard deviation are reported over 20 runs). Models trained and tested on 80:20 split shows that a random subset is able to match within-5% Accuracy and AUC of all features.

**Fig. 23.**

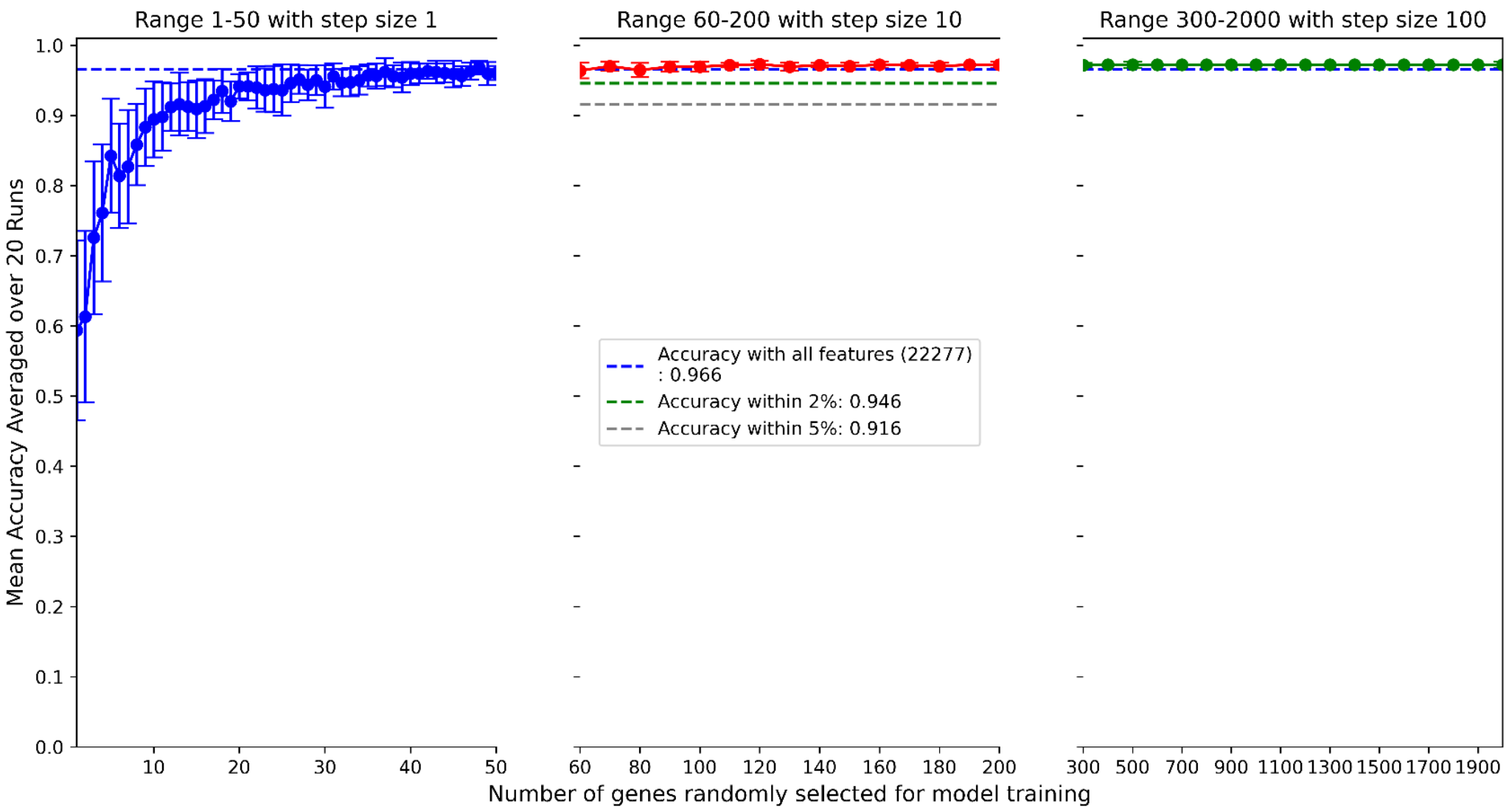

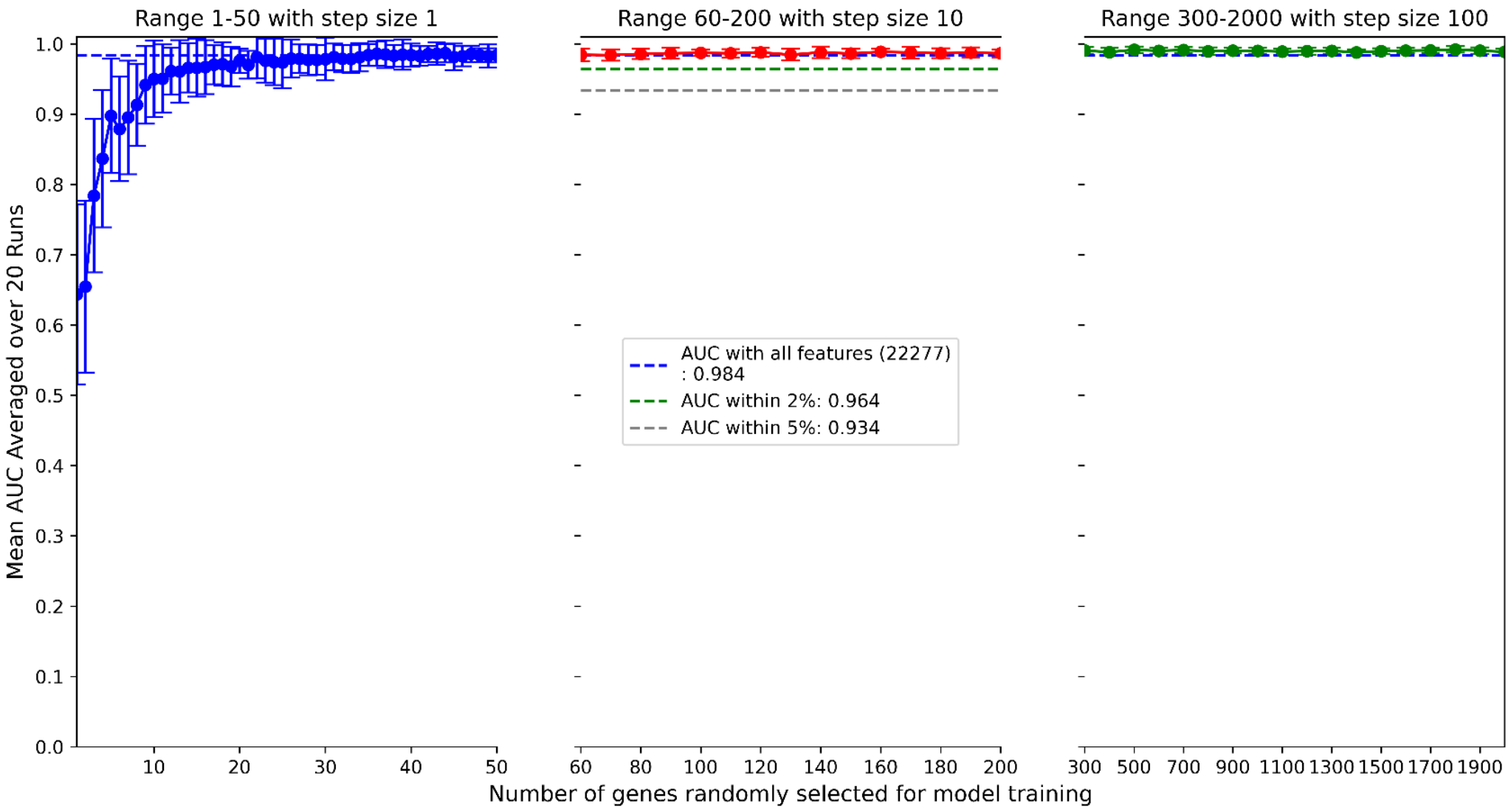

**Random Forest performance with Liver Cancer (GSE14520) dataset** (mean and standard deviation are reported over 20 runs). Models trained and tested on 80:20 split shows that a small random subset is able to match full Accuracy and AUC of all features.

**Fig. 24.**

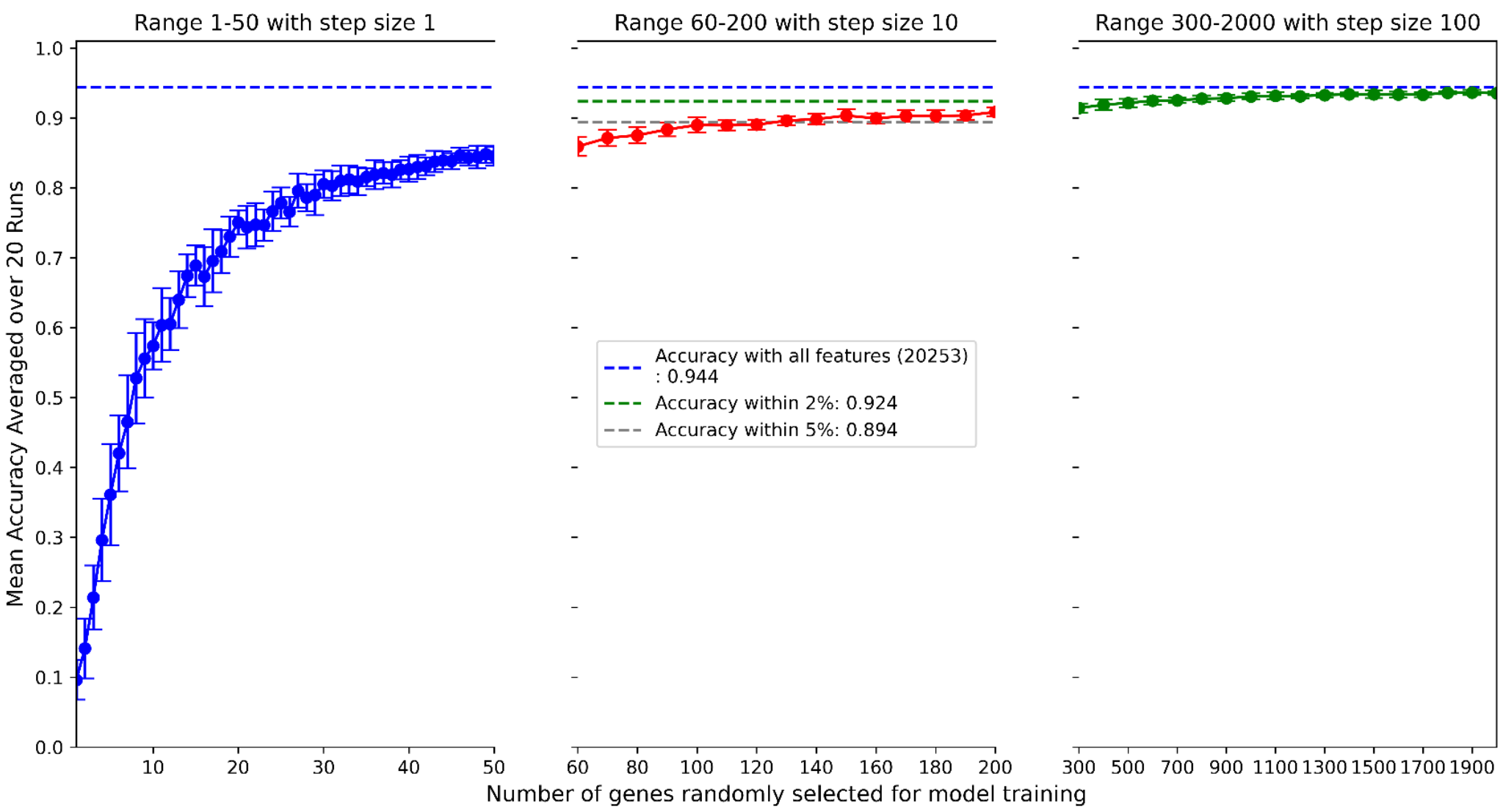

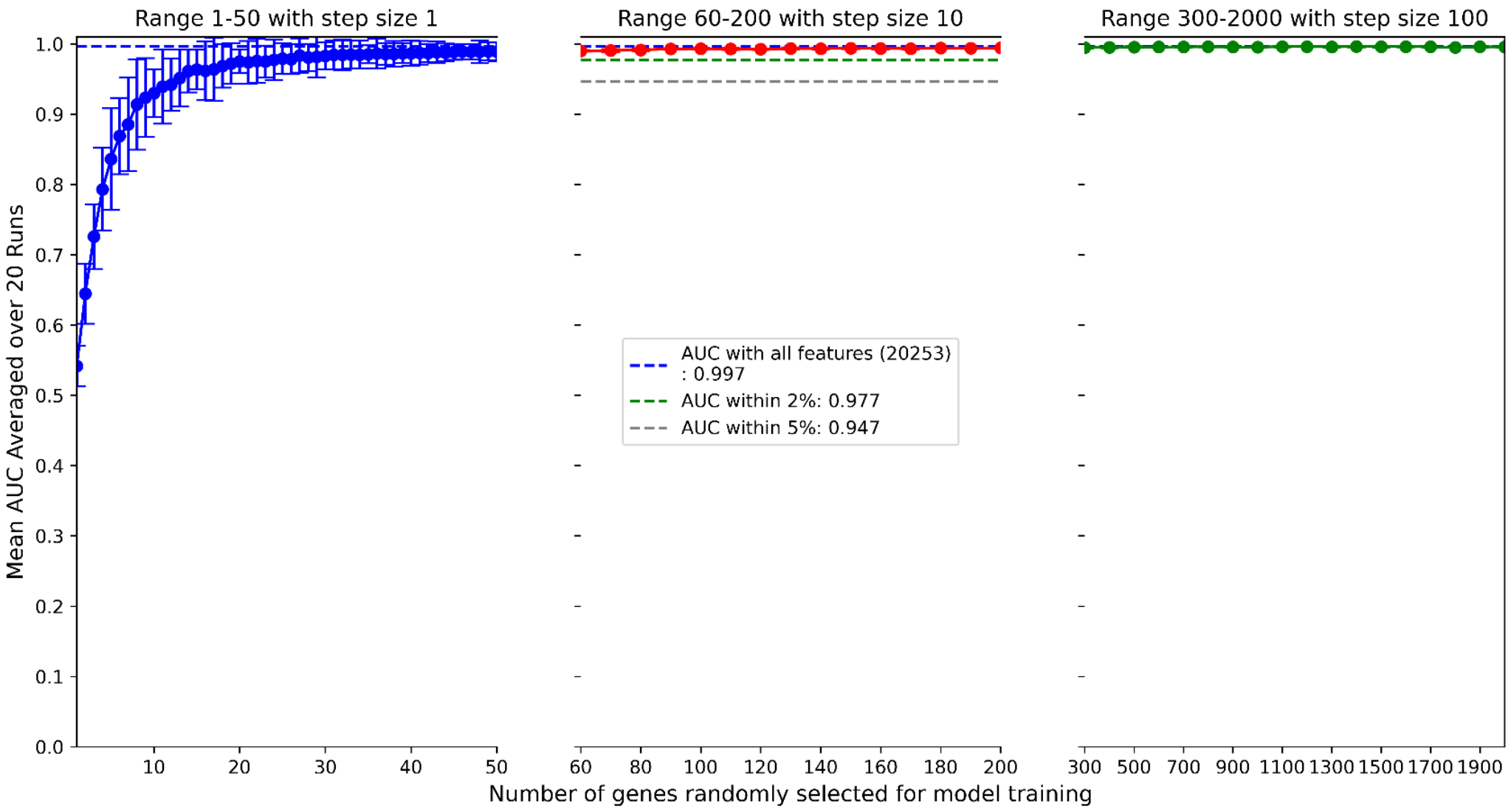

**Random Forest performance with bulk RNA-Seq TCGA dataset with 33 classes** (mean and standard deviation are reported over 20 runs). Models trained and tested on 80:20 split shows that a small random subset is able to match full Accuracy and AUC of all features.

## Fig. 25.

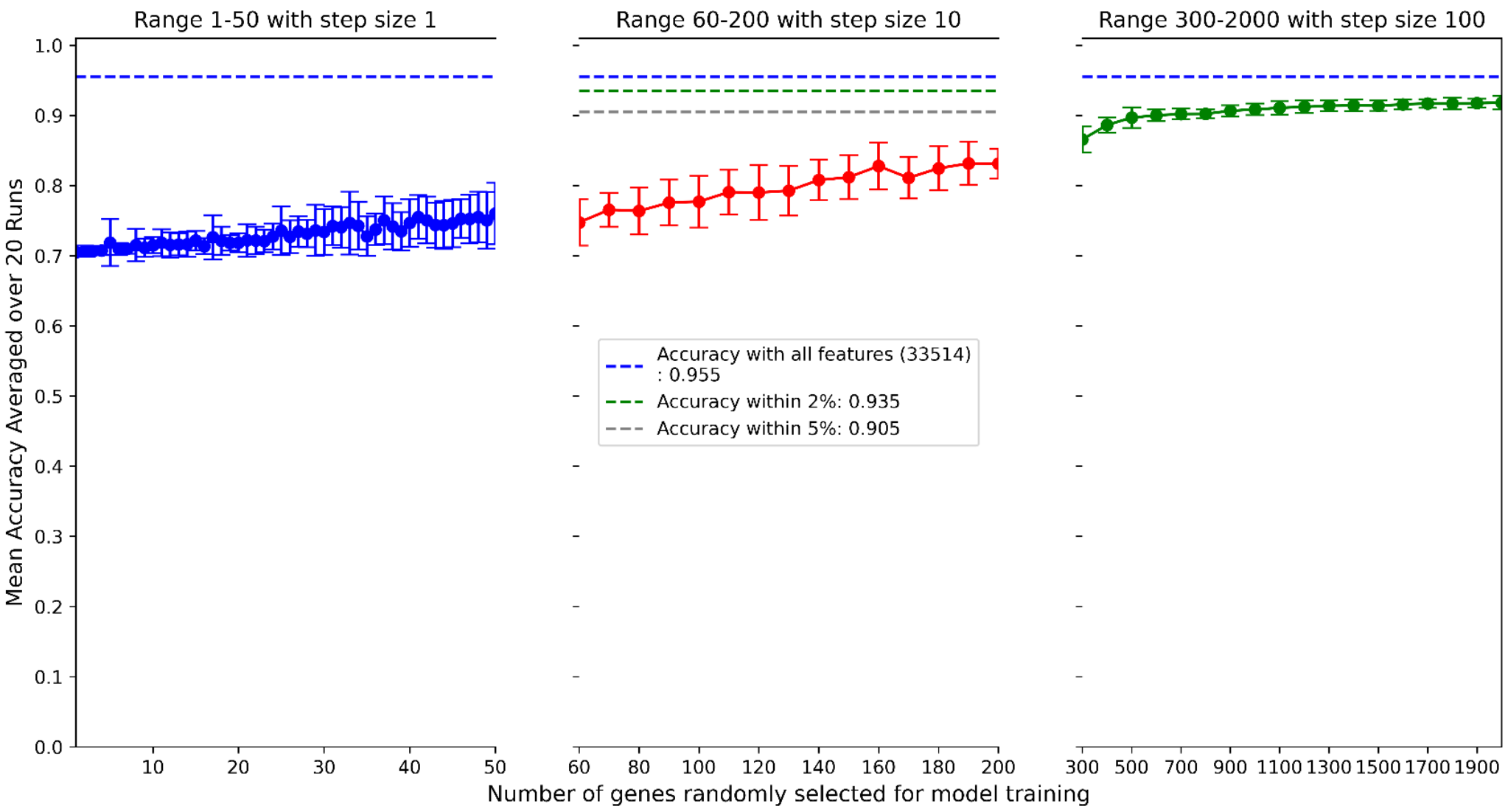

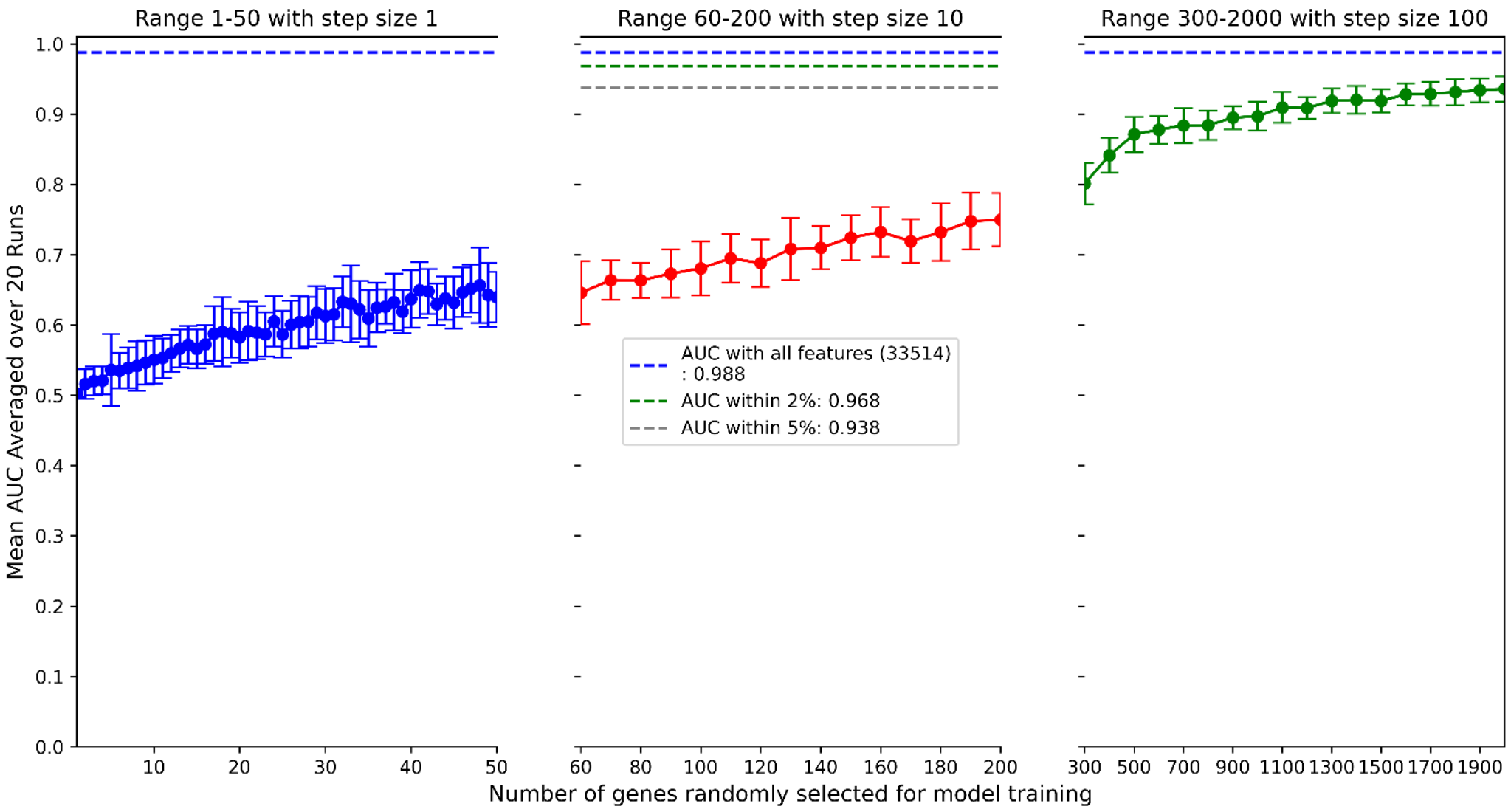

**Random Forest performance with single-cell RNA-Seq Lung dataset with 9 classes** (mean and standard deviation are reported over 20 runs). Models trained and tested on 80:20 split shows that a small random subset is able to match within-5% Accuracy and AUC of all features.

**Fig. 26.**

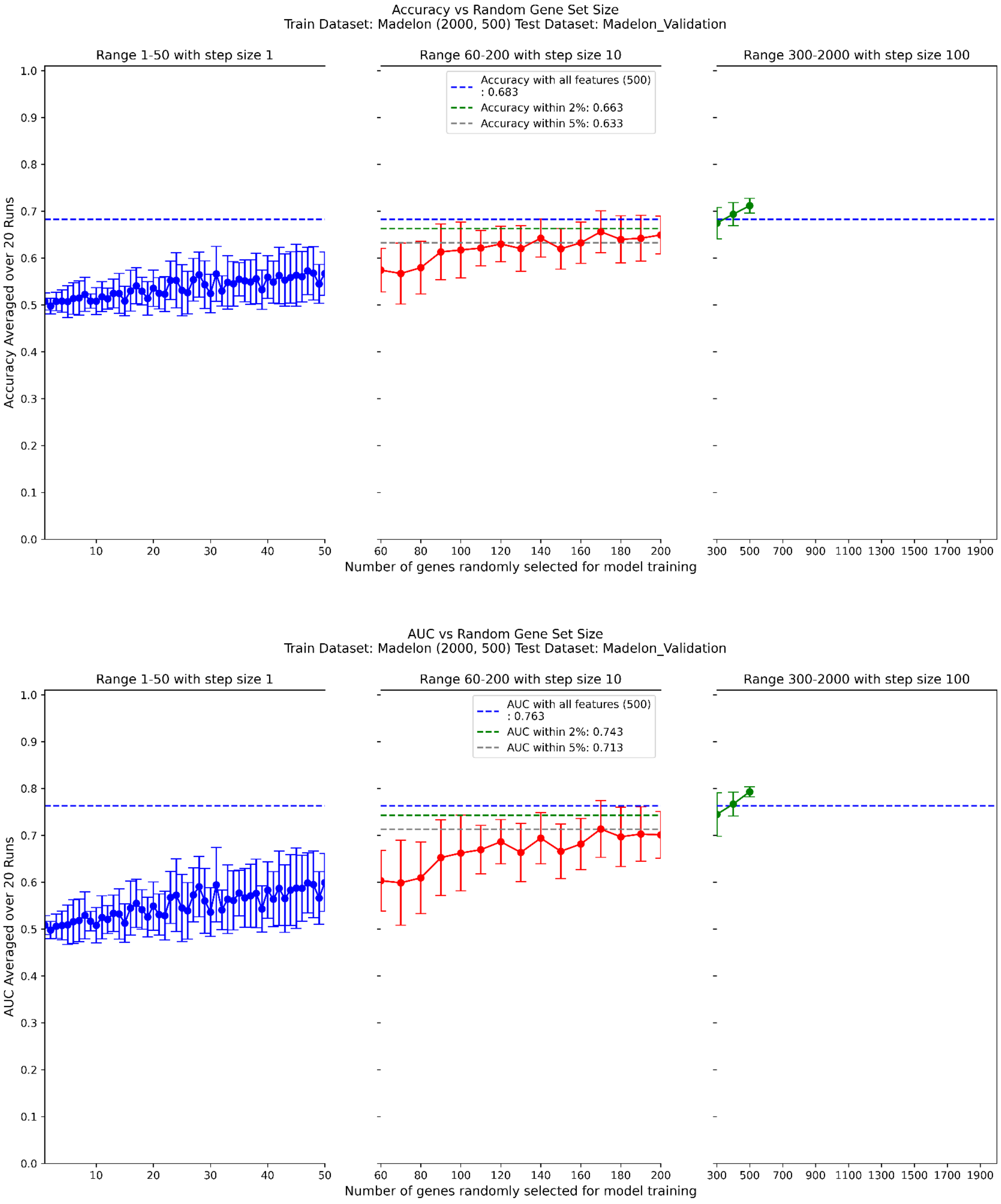

**Random Forest performance with Madelon dataset** (mean and standard deviation are reported over 20 runs). The task of MADELON is to classify random data. Models trained and tested on 80:20 split shows that a small random subset is able to match within-5% Accuracy and AUC of all features. (As there are only 500 features in this dataset, there is no result beyond 500).

**Fig. 27.**

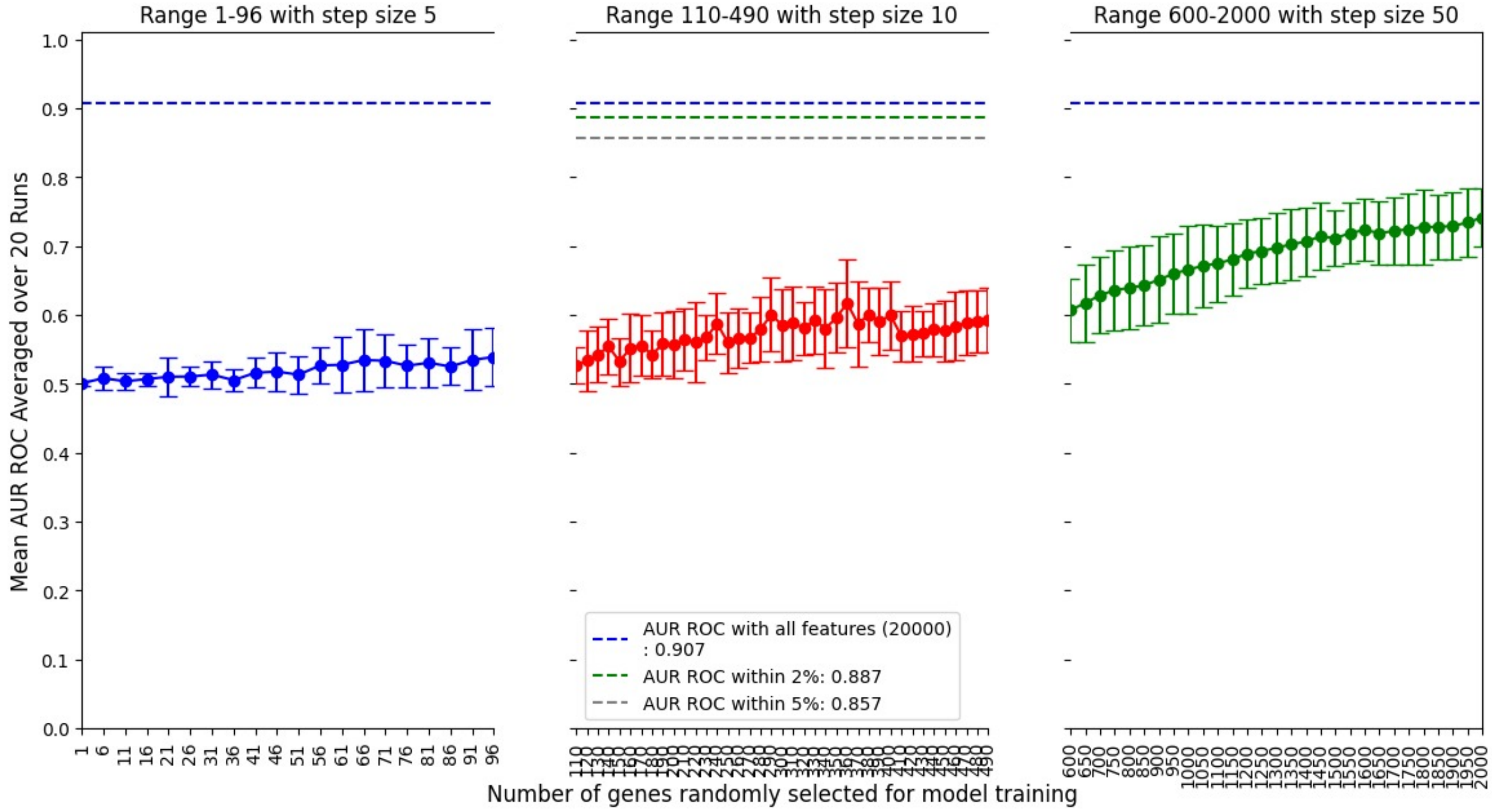

**Random Forest performance with Dexter dataset** (mean and standard deviation are reported over 20 runs). The task of DEXTER is to filter texts about "corporate acquisitions". Models trained and tested on 80:20 split shows that a random subset is NOT able to match AUC of all features.

**Fig. 28.**

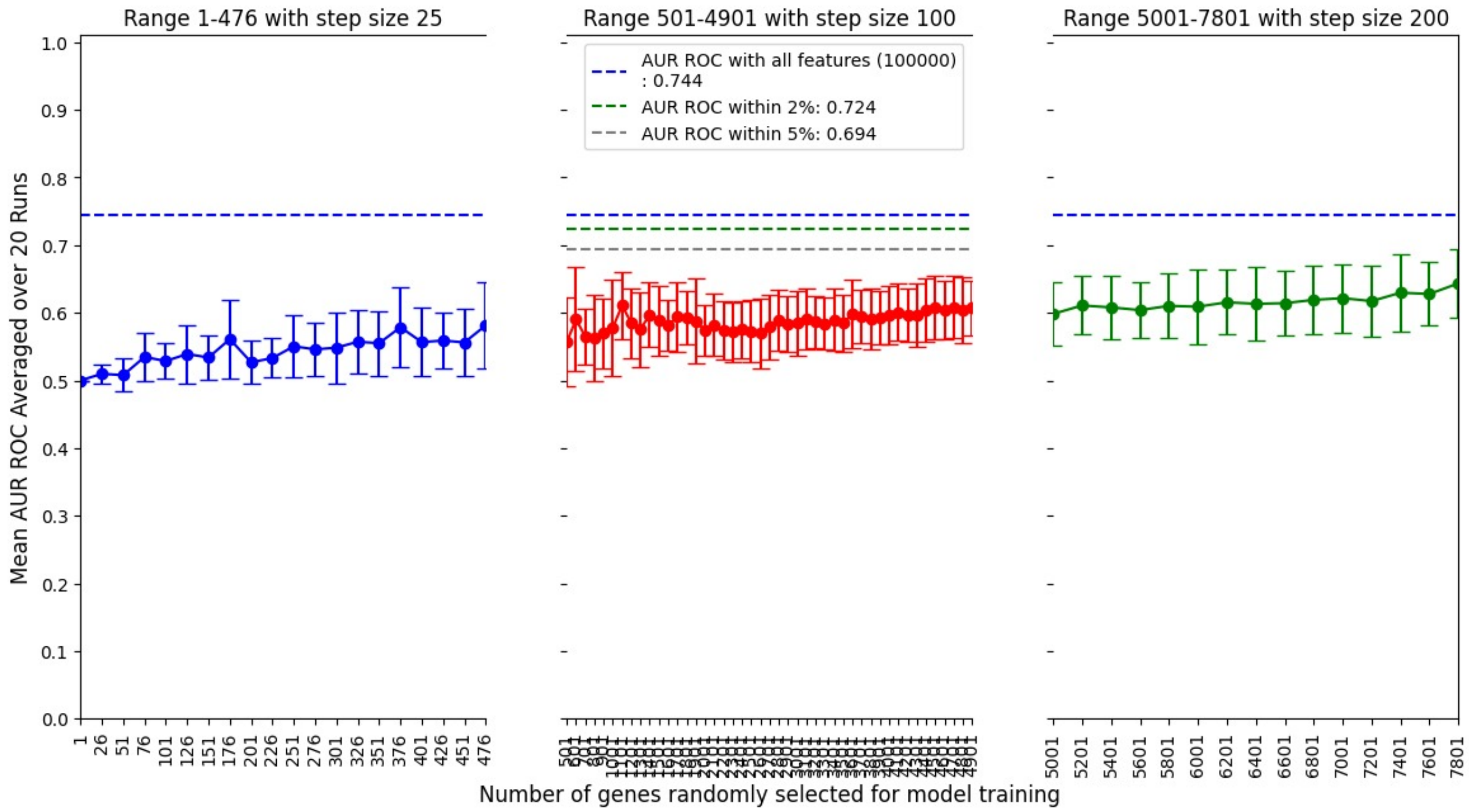

**Random Forest performance with Dorothea dataset** (mean and standard deviation are reported over 20 runs). The task of DOROTHEA is to predict which compounds bind to Thrombin. Models trained and tested on 80:20 split shows that a random subset is NOT able to match AUC of all features.